\documentclass{article}

    \usepackage[final]{neurips_data_2023}

\usepackage[utf8]{inputenc} % allow utf-8 input
\usepackage[T1]{fontenc}    % use 8-bit T1 fonts
\usepackage{hyperref}       % hyperlinks
\usepackage{url}            % simple URL typesetting
\usepackage{booktabs}       % professional-quality tables
\usepackage{amsfonts}       % blackboard math symbols
\usepackage{nicefrac}       % compact symbols for 1/2, etc.
\usepackage{microtype}      % microtypography
\usepackage{xcolor}         % colors
\usepackage{adjustbox}
\usepackage{custom}
\usepackage{cleveref}

\definecolor{dgreen}{rgb}{0.0,0.5,0.0}

\title{CORL: Research-oriented Deep Offline Reinforcement Learning Library}

\author{%
  Denis Tarasov\\
  Tinkoff\\
  \texttt{den.tarasov@tinkoff.ai} \\
  \And
  Alexander Nikulin\\
  Tinkoff\\
  \texttt{a.p.nikulin@tinkoff.ai} \\
  \And
  Dmitry Akimov\\
  Tinkoff\\
  \texttt{d.akimov@tinkoff.ai} \\
  \And
  Vladislav Kurenkov\\
  Tinkoff\\
  \texttt{v.kurenkov@tinkoff.ai} \\
  \And
  Sergey Kolesnikov\\
  Tinkoff\\
  \texttt{s.s.kolesnikov@tinkoff.ai} \\
}

\begin{document}

\maketitle

\begin{abstract}
  CORL\footnote{CORL Repository: \url{https://github.com/corl-team/CORL}} is an open-source library that provides thoroughly benchmarked single-file implementations of both deep offline and offline-to-online reinforcement learning algorithms. It emphasizes a simple developing experience with a straightforward codebase and a modern analysis tracking tool. In CORL, we isolate methods implementation into separate single files, making performance-relevant details easier to recognize. Additionally, an experiment tracking feature is available to help log metrics, hyperparameters, dependencies, and more to the cloud. Finally, we have ensured the reliability of the implementations by benchmarking commonly employed D4RL datasets providing a transparent source of results that can be reused for robust evaluation tools such as performance profiles, probability of improvement, or expected online performance.
\end{abstract}

\section{Introduction}
\label{introduction}
Deep Offline Reinforcement Learning \citep{levine2020offline} has been showing significant advancements in numerous domains such as robotics \citep{smithWalkParkLearning2022, kumar2021a}, autonomous driving \citep{Diehl2021UMBRELLAUM} and recommender systems \citep{Chen2022OffPolicyAF}. Due to such rapid development, many open-source offline RL solutions\footnote{\url{https://github.com/hanjuku-kaso/awesome-offline-rl##oss}}
emerged to help RL practitioners understand and improve well-known offline RL techniques in different fields. On the one hand, they introduce offline RL algorithms standard interfaces and user-friendly APIs, simplifying offline RL methods incorporation into \textit{existing} projects.
On the other hand, introduced abstractions may hinder the learning curve for newcomers and the ease of adoption for researchers interested in developing \textit{new} algorithms. One needs to understand the modularity design (several files on average), which (1) can be comprised of thousands of lines of code or (2) can hardly fit for a novel method\footnote{\url{https://github.com/takuseno/d3rlpy/issues/141} }.

In this technical report, we take a different perspective on an offline RL library and also incorporate emerging interest in the offline-to-online setup. We propose CORL (Clean Offline Reinforcement Learning) -- minimalistic and isolated single-file implementations of deep offline and offline-to-online RL algorithms, supported by open-sourced D4RL \citep{d4rl} benchmark results. The uncomplicated design allows practitioners to read and understand the implementations of the algorithms straightforwardly. Moreover, CORL supports optional integration with experiments tracking tools such as Weighs\&Biases~\citep{wandb},  providing practitioners with a convenient way to analyze the results and behavior of all algorithms, not merely relying on a final performance commonly reported in papers.

We hope that the CORL library will help offline RL newcomers study implemented algorithms and aid the researchers in quickly modifying existing methods without fighting through different levels of abstraction. Finally, the obtained results may serve as a reference point for D4RL benchmarks avoiding the need to re-implement and tune existing algorithms' hyperparameters.

\begin{figure}[ht]
\begin{center}
\centerline{\includegraphics[width=1.0\textwidth]{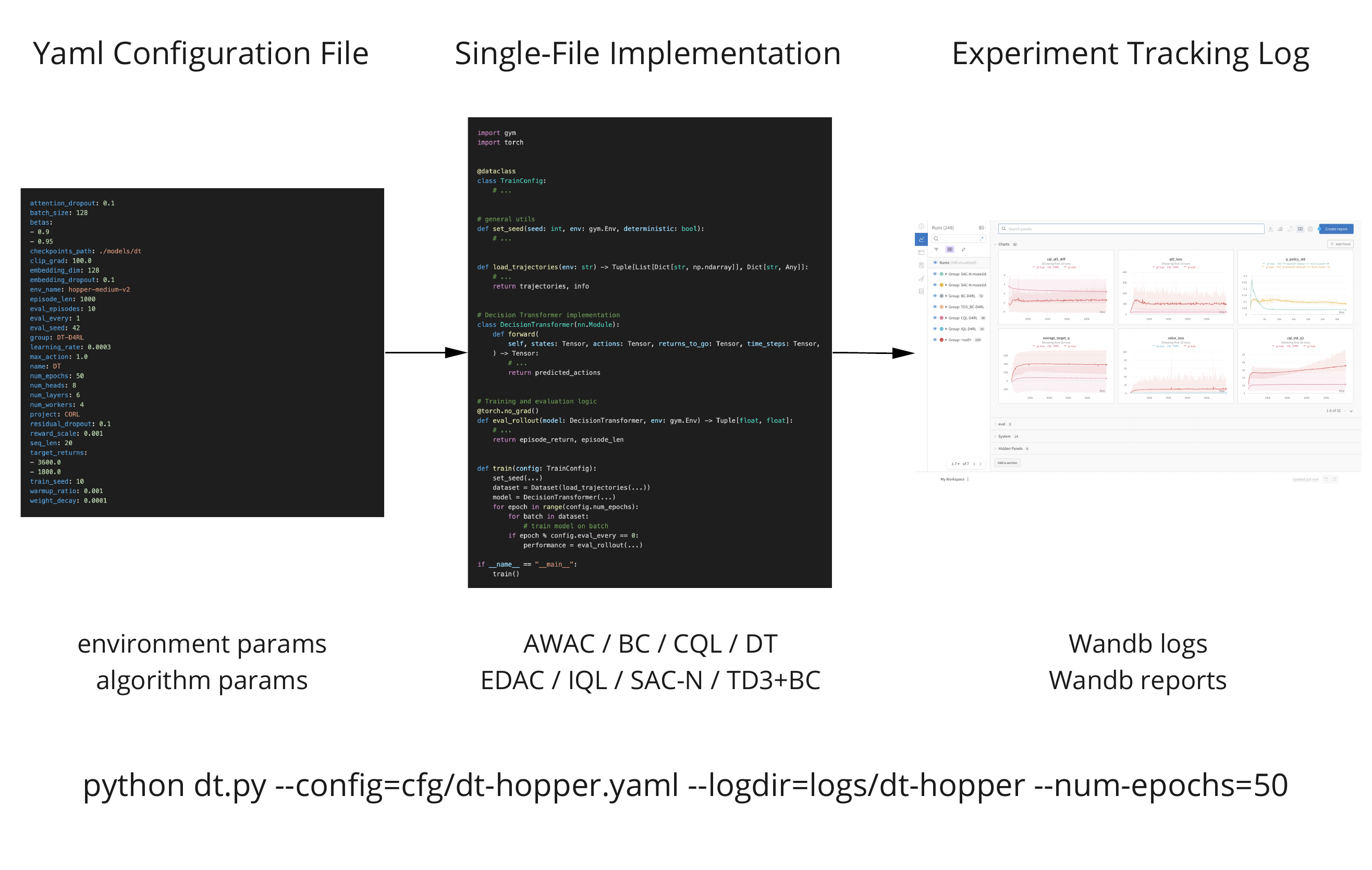}}
\caption{\centering The illustration of the CORL library design. Single-file implementation takes a yaml configuration file with both environment and algorithm parameters to run the experiment, which logs all required statistics to Weights\&Biases \citep{wandb}.}
\label{fig:corl-design}
\end{center}
\vskip -0.2in
\end{figure}

\section{Related Work}
\label{related}
Since the Atari breakthrough \citep{mnihHumanlevelControlDeep2015}, numerous open-source RL frameworks and libraries have been developed over the last years: \citep{baselines, stable-baselines, castro2018dopamine, gauci2018horizon, kenggraesser2017slmlab, garage, 10.5555/3045390.3045531, 1903.00027, JMLR:v22:20-376, liang2018rllib, JMLR:v22:20-376, erl, huang2021cleanrl, tianshou, 1909.01500}, focusing on different perspectives of the RL. For example, stable-baselines \citep{stable-baselines} provides many deep RL implementations that carefully reproduce results to back up RL practitioners with reliable baselines during methods comparison. On the other hand, Ray \citep{liang2018rllib} focuses on implementations scalability and production-friendly usage. Finally, more nuanced solutions exist, such as Dopamine \citep{castro2018dopamine}, which emphasizes different DQN variants, or ReAgent \citep{gauci2018horizon}, which applies RL to the RecSys domain.

At the same time, the offline RL branch and especially offline-to-online, which we are interested in this paper, are not yet covered as much: the only library that precisely focuses on offline RL setting is d3rlpy \citep{seno2021d3rlpy}. While CORL also focuses on offline RL methods \citep{nair2020accelerating, kumar2020conservative, kostrikov2021offline, fujimoto2021minimalist, an2021uncertainty, chen2021decision}, similar to d3rlpy, it takes a different perspective on library design and provides \textit{non-modular} independent algorithms implementations. More precisely, CORL does not introduce additional abstractions to make offline RL more general but instead gives an "easy-to-hack" starter kit for research needs. Finally, CORL also provides recent offline-to-online solutions \citep{nair2020accelerating, kumar2020conservative, kostrikov2021offline, wu2022supported, nakamoto2023cal, tarasov2023revisiting} that are gaining interest among researchers and practitioners. 

Although CORL does not represent the first non-modular RL library, which is more likely the CleanRL \citep{huang2021cleanrl} case, it has two significant differences from its predecessor. First, CORL is focused on \textit{offline} and \textit{offline-to-online} RL, while CleanRL implements \textit{online} RL algorithms. Second, CORL intends to minimize the complexity of the requirements and external dependencies. To be more concrete, CORL does not have additional requirements with abstractions such as $stable$-$baselines$ \citep{stable-baselines} or $envpool$ \citep{weng2022envpool} but instead implements everything from scratch in the codebase.    

\section{CORL Design}
\label{method}

\subsection*{Single-File Implementations}
Implementational subtleties significantly impact agent performance in deep RL \citep{10.5555/3504035.3504427, Engstrom2020ImplementationMI, fujimoto2021minimalist}. 
Unfortunately, user-friendly abstractions and general interfaces, the core idea behind modular libraries, encapsulate and often hide these important nuances from the practitioners.
For such a reason, CORL unwraps these details by adopting single-file implementations. To be more concrete, we put environment details, algorithms hyperparameters, and evaluation parameters into a single file\footnote{We follow the PEP8 style guide with a maximum line length of 89, which increases LOC a bit.}. For example, we provide
\begin{itemize}
\item $any\_percent\_bc.py$ (404 LOC\footnote{Lines Of Code}) as a baseline algorithm for offline RL methods comparison,
\item $td3\_bc.py$ (511 LOC) as a competitive minimalistic offline RL algorithm~\citep{fujimoto2021minimalist},
\item $dt.py$ (540 LOC) as an example of the recently proposed trajectory optimization approach~\citep{chen2021decision}
\end{itemize}

\autoref{fig:corl-design} depicts an overall library design. 
To avoid over-complicated offline implementations, we treat offline and offline-to-online versions of the same algorithms separately. While such design produces code duplications among realization, it has several essential benefits from the both educational and research perspective:
\begin{itemize}
\item \textbf{Smooth learning curve}. Having the entire code in one place makes understanding all its aspects more straightforward. In other words, one may find it easier to dive into 540 LOC of single-file Decision Transformer \citep{chen2021decision} implementation rather than 10+ files of the original implementation\footnote{Original Decision Transformer implementation: \url{https://github.com/kzl/decision-transformer}}.
\item \textbf{Simple prototyping}. As we are not interested in the code's general applicability, we could make it implementation-specific. Such a design also removes the need for inheritance from general primitives or their refactoring, reducing abstraction overhead to zero. At the same time, this idea gives us complete freedom during code modification. 
% \textbf{For example, ...}
\item \textbf{Faster debugging}. Without additional abstractions, implementation simplifies to a single for-loop with a global Python name scope. Furthermore, such flat architecture makes accessing and inspecting any created variable easier during training, which is crucial in the presence of modifications and debugging.
\end{itemize}

\subsection*{Configuration files}

Although it is a typical pattern to use a command line interface (CLI) for single-file experiments in the research community, CORL slightly improves it with predefined configuration files. Utilizing YAML parsing through CLI, for each experiment, we gather all environment and algorithm hyperparameters into such files so that one can use them as an initial setup. We found that such setup (1) simplifies experiments, eliminating the need to keep all algorithm-environment-specific parameters in mind, and (2) keeps it convenient with the familiar CLI approach.

\subsection*{Experiment Tracking}
Offline RL evaluation is another challenging aspect of the current offline RL state \citep{pmlr-v162-kurenkov22a}. To face this uncertainty, CORL supports integration with Weights\&Biases \citep{wandb}, a modern experiment tracking tool. With each experiment, CORL automatically saves (1) source code, (2) dependencies (requirements.txt), (3) hardware setup, (4) OS environment variables, (5) hyperparameters, (6) training, and system metrics, (7) logs (stdout, stderr). See \autoref{app:tracking} for an example.

Although, Weights\&Biases is a proprietary solution, other alternatives, such as Tensorboard \citep{tensorflow2015-whitepaper} or Aim \citep{Arakelyan_Aim_2020}, could be used within a few lines of code change. It is also important to note that with Weights\&Biases tracking, one could easily use CORL with sweeps or public reports. 

We found full metrics tracking during the training process necessary for two reasons. First, it removes the possible bias of the final or best performance commonly reported in papers. For example, one could evaluate offline RL performance as max archived score, while another uses the average scores over $N$ (last) evaluations \citep{seno2021d3rlpy}. Second, it provides an opportunity for advanced performance analysis such as EOP \citep{pmlr-v162-kurenkov22a} or RLiable \citep{agarwal2021deep}. In short, when provided with all metrics logs, one can utilize all performance statistics, not merely relying on commonly used alternatives.

\section{Benchmarking D4RL}

\subsection{Offline}
In our library, we implemented the following offline algorithms: $N\%$\footnote{$N$ is a percentage of best trajectories with the highest return used for training. We omit the percentage when it is equal to $100$.} Behavioral Cloning (BC), TD3 + BC~\citep{fujimoto2021minimalist}, CQL~\citep{kumar2020conservative}, IQL~\citep{kostrikov2021offline}, AWAC~\citep{nair2020accelerating}, ReBRAC~\citep{tarasov2023revisiting}, SAC-N, EDAC~\citep{an2021uncertainty}, and Decision Transformer (DT)~\citep{chen2021decision}. We evaluated every algorithm on the D4RL benchmark \citep{d4rl}, focusing on Gym-MuJoCo, Maze2D, AntMaze, and Adroit tasks. Each algorithm was run for one million gradient steps\footnote{\label{note1}Except SAC-$N$, EDAC, and DT due to their original hyperparameters. See \autoref{appendix:hyperparameters} for details. } and evaluated using ten episodes for Gym-MuJoCo and Adroit tasks. For Maze2d, we use 100 evaluation episodes. In our experiments, we tried to rely on the hyperparameters proposed in the original works (see \autoref{appendix:hyperparameters} for details) as much as possible.

\begin{figure}[ht]
\centering
\captionsetup{justification=centering}
     \centering
        \begin{subfigure}[b]{0.49\textwidth}
         \centering
         \includegraphics[width=1.0\textwidth]{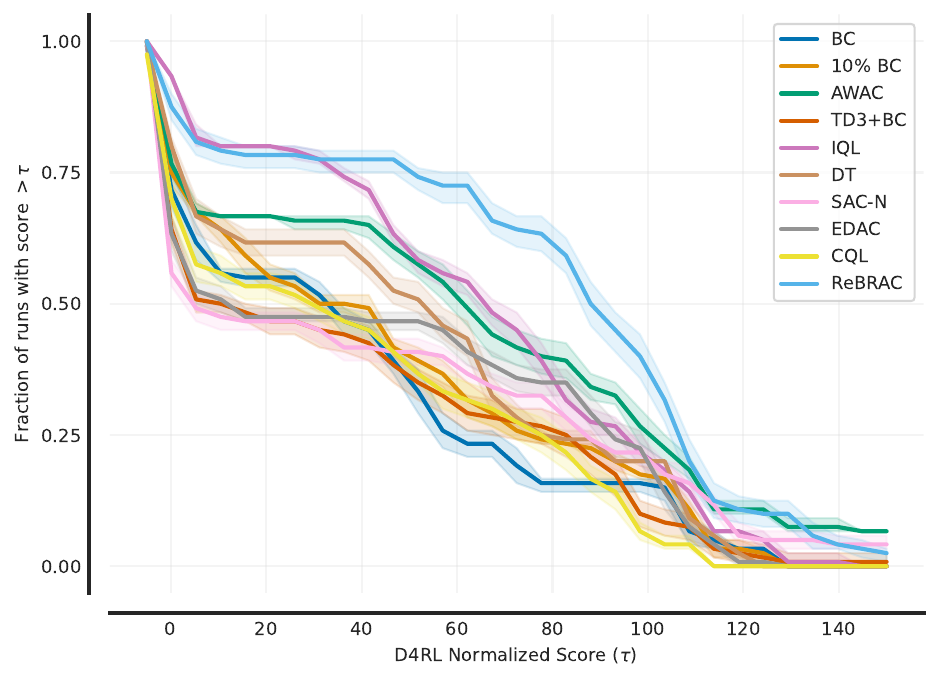}
         \caption{}
         \label{fig:perf_prof_offline}
        \end{subfigure}
        \begin{subfigure}[b]{0.49\textwidth}
         \centering
         \includegraphics[width=1.0\textwidth]{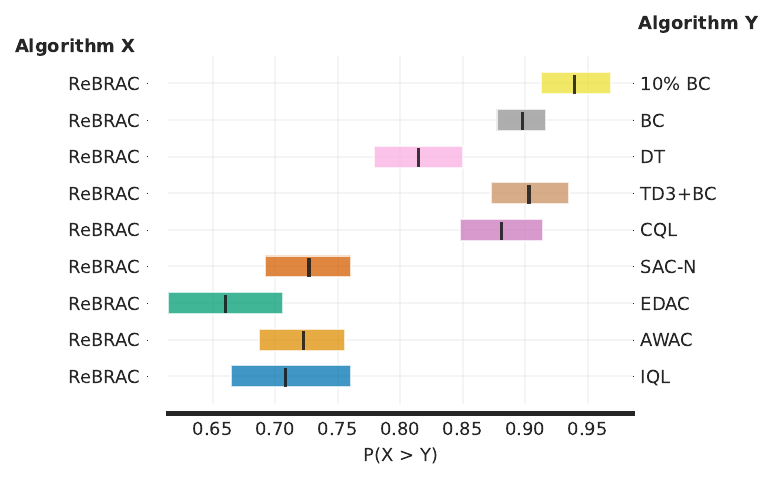}
         \caption{}
         \label{fig:prob_improve_offline}
        \end{subfigure}
        \caption{{(a) Performance profiles after offline training (b) Probability of improvement of ReBRAC to other algorithms after offline training. The curves \citep{agarwal2021deep} are for D4RL benchmark spanning Gym-MuJoCo, Maze2d, AntMaze, and Adroit datasets.}}
        \label{fig:curves_offline}
\end{figure}

The final performance is reported in \Cref{tab:gym} and the maximal performance in \Cref{tab:gym_max}. The scores are normalized to the range between 0 and 100 \citep{d4rl}. Following the recent work by \citet{seno2021d3rlpy}, we report the last and best-obtained scores to illustrate each algorithm's potential performance and overfitting properties. \autoref{fig:curves_offline} shows the performance profiles and probability of improvement of ReBRAC over other algorithms \citep{agarwal2021deep}. See \autoref{appendix:results} for complete training performance graphs.

\begin{table}[ht]%[ht]
	\centering
	\small
	\caption{Normalized performance of the last trained policy on D4RL averaged over 4 random seeds.}
% 	\vspace{0.2em}
	\label{tab:gym}
	\begin{adjustbox}{max width=\columnwidth}
		\begin{tabular}{l|rrrrrrrrrr}
			\toprule
			\multirow{2}{*}{\textbf{Task Name}} & \multirow{2}{*}{\textbf{BC}} & \multirow{2}{*}{\textbf{BC-10\%}} & \multirow{2}{*}{\textbf{TD3+BC}} & \multirow{2}{*}{\textbf{AWAC}} & \multirow{2}{*}{\textbf{CQL}} & \multirow{2}{*}{\textbf{IQL}}  & \multirow{2}{*}{\textbf{ReBRAC}} & \multirow{2}{*}{\textbf{SAC-$N$}} & \multirow{2}{*}{\textbf{EDAC}} & \multirow{2}{*}{\textbf{DT}} \\
			& & & & & & & \\
			\midrule
           halfcheetah-medium-v2 & 42.40 $\pm$ 0.19 & 42.46 $\pm$ 0.70 & 48.10 $\pm$ 0.18 &  50.02 $\pm$ 0.27 & 47.04 $\pm$ 0.22 & 48.31 $\pm$ 0.22 & 64.04 $\pm$ 0.68 & \textbf{68.20} $\pm$ 1.28 & 67.70 $\pm$ 1.04 & 42.20 $\pm$ 0.26 \\
halfcheetah-medium-replay-v2 & 35.66 $\pm$ 2.33 & 23.59 $\pm$ 6.95 & 44.84 $\pm$ 0.59 & 45.13 $\pm$ 0.88 & 45.04 $\pm$ 0.27 & 44.46 $\pm$ 0.22 & 51.18 $\pm$ 0.31 & 60.70 $\pm$ 1.01 & \textbf{62.06} $\pm$ 1.10 & 38.91 $\pm$ 0.50 \\
halfcheetah-medium-expert-v2 & 55.95 $\pm$ 7.35 & 90.10 $\pm$ 2.45 & 90.78 $\pm$ 6.04 & 95.00 $\pm$ 0.61 & 95.63 $\pm$ 0.42 & 94.74 $\pm$ 0.52 & 103.80 $\pm$ 2.95 & 98.96 $\pm$ 9.31 & \textbf{104.76} $\pm$ 0.64 & 91.55 $\pm$ 0.95 \\
\midrule
hopper-medium-v2 & 53.51 $\pm$ 1.76 & 55.48 $\pm$ 7.30 & 60.37 $\pm$ 3.49 & 63.02 $\pm$ 4.56 & 59.08 $\pm$ 3.77 & 67.53 $\pm$ 3.78 & \textbf{102.29} $\pm$ 0.17 & 40.82 $\pm$ 9.91 & 101.70 $\pm$ 0.28 & 65.10 $\pm$ 1.61 \\
hopper-medium-replay-v2 & 29.81 $\pm$ 2.07 & 70.42 $\pm$ 8.66 & 64.42 $\pm$ 21.52 &  98.88 $\pm$ 2.07 & 95.11 $\pm$ 5.27 & 97.43 $\pm$ 6.39 & 94.98 $\pm$ 6.53 & \textbf{100.33} $\pm$ 0.78 & 99.66 $\pm$ 0.81 & 81.77 $\pm$ 6.87 \\
hopper-medium-expert-v2 & 52.30 $\pm$ 4.01 & 111.16 $\pm$ 1.03 & 101.17 $\pm$ 9.07 & 101.90 $\pm$ 6.22 & 99.26 $\pm$ 10.91 & 107.42 $\pm$ 7.80 & 109.45 $\pm$ 2.34 & 101.31 $\pm$ 11.63 & 105.19 $\pm$ 10.08 & \textbf{110.44} $\pm$ 0.33 \\
\midrule
walker2d-medium-v2 & 63.23 $\pm$ 16.24 & 67.34 $\pm$ 5.17 & 82.71 $\pm$ 4.78 & 68.52 $\pm$ 27.19 & 80.75 $\pm$ 3.28 & 80.91 $\pm$ 3.17 & 85.82 $\pm$ 0.77 & 87.47 $\pm$ 0.66 & \textbf{93.36} $\pm$ 1.38 & 67.63 $\pm$ 2.54 \\
walker2d-medium-replay-v2 & 21.80 $\pm$ 10.15 & 54.35 $\pm$ 6.34 & 85.62 $\pm$ 4.01 &  80.62 $\pm$ 3.58 & 73.09 $\pm$ 13.22 & 82.15 $\pm$ 3.03 & 84.25 $\pm$ 2.25 & 78.99 $\pm$ 0.50 & \textbf{87.10} $\pm$ 2.78 & 59.86 $\pm$ 2.73 \\
walker2d-medium-expert-v2 & 98.96 $\pm$ 15.98 & 108.70 $\pm$ 0.25 & 110.03 $\pm$ 0.36 & 111.44 $\pm$ 1.62 & 109.56 $\pm$ 0.39 & 111.72 $\pm$ 0.86 & 111.86 $\pm$ 0.43 & \textbf{114.93} $\pm$ 0.41 & 114.75 $\pm$ 0.74 & 107.11 $\pm$ 0.96 \\
\midrule
\textbf{Gym-MuJoCo avg} & 50.40 & 69.29 & 76.45 & 79.39 & 78.28 & 81.63 & 89.74 & 83.52 & \textbf{92.92} & 73.84 \\
\midrule
maze2d-umaze-v1 & 0.36 $\pm$ 8.69 & 12.18 $\pm$ 4.29 & 29.41 $\pm$ 12.31 & 65.65 $\pm$ 5.34  & -8.90 $\pm$ 6.11 & 42.11 $\pm$ 0.58 & 106.87 $\pm$ 22.16 & \textbf{130.59} $\pm$ 16.52 & 95.26 $\pm$ 6.39 & 18.08 $\pm$ 25.42 \\
maze2d-medium-v1 & 0.79 $\pm$ 3.25 & 14.25 $\pm$ 2.33 & 59.45 $\pm$ 36.25 & 84.63 $\pm$ 35.54 & 86.11 $\pm$ 9.68 & 34.85 $\pm$ 2.72 & \textbf{105.11} $\pm$ 31.67 & 88.61 $\pm$ 18.72 & 57.04 $\pm$ 3.45 & 31.71 $\pm$ 26.33 \\
maze2d-large-v1 & 2.26 $\pm$ 4.39 & 11.32 $\pm$ 5.10 & 97.10 $\pm$ 25.41 & \textbf{215.50} $\pm$ 3.11 & 23.75 $\pm$ 36.70 & 61.72 $\pm$ 3.50 & 78.33 $\pm$ 61.77 & 204.76 $\pm$ 1.19 & 95.60 $\pm$ 22.92 & 35.66 $\pm$ 28.20 \\
\midrule
\textbf{Maze2d avg} & 1.13 & 12.58 & 61.99 & 121.92 & 33.65 & 46.23 & 96.77 & \textbf{141.32} & 82.64 & 28.48 \\
\midrule
antmaze-umaze-v2 & 55.25 $\pm$ 4.15 & 65.75 $\pm$ 5.26 & 70.75 $\pm$ 39.18 &  56.75 $\pm$ 9.09 & 92.75 $\pm$ 1.92 & 77.00 $\pm$ 5.52 & \textbf{97.75} $\pm$ 1.48 & 0.00 $\pm$ 0.00 & 0.00 $\pm$ 0.00 & 57.00 $\pm$ 9.82 \\
antmaze-umaze-diverse-v2 & 47.25 $\pm$ 4.09 & 44.00 $\pm$ 1.00 & 44.75 $\pm$ 11.61 & 54.75 $\pm$ 8.01 & 37.25 $\pm$ 3.70 & 54.25 $\pm$ 5.54 & \textbf{83.50} $\pm$ 7.02 & 0.00 $\pm$ 0.00 & 0.00 $\pm$ 0.00 & 51.75 $\pm$ 0.43 \\
antmaze-medium-play-v2 & 0.00 $\pm$ 0.00 & 2.00 $\pm$ 0.71 & 0.25 $\pm$ 0.43 &  0.00 $\pm$ 0.00  & 65.75 $\pm$ 11.61 & 65.75 $\pm$ 11.71 & \textbf{89.50} $\pm$ 3.35 & 0.00 $\pm$ 0.00 & 0.00 $\pm$ 0.00 & 0.00 $\pm$ 0.00 \\
antmaze-medium-diverse-v2 & 0.75 $\pm$ 0.83 & 5.75 $\pm$ 9.39 & 0.25 $\pm$ 0.43 &  0.00 $\pm$ 0.00 & 67.25 $\pm$ 3.56 & 73.75 $\pm$ 5.45 & \textbf{83.50} $\pm$ 8.20 & 0.00 $\pm$ 0.00 & 0.00 $\pm$ 0.00 & 0.00 $\pm$ 0.00 \\
antmaze-large-play-v2 & 0.00 $\pm$ 0.00 & 0.00 $\pm$ 0.00 & 0.00 $\pm$ 0.00 & 0.00 $\pm$ 0.00 & 20.75 $\pm$ 7.26 & 42.00 $\pm$ 4.53 & \textbf{52.25} $\pm$ 29.01 & 0.00 $\pm$ 0.00 & 0.00 $\pm$ 0.00 & 0.00 $\pm$ 0.00 \\
antmaze-large-diverse-v2 & 0.00 $\pm$ 0.00 & 0.75 $\pm$ 0.83 & 0.00 $\pm$ 0.00 &  0.00 $\pm$ 0.00 & 20.50 $\pm$ 13.24 & 30.25 $\pm$ 3.63 & \textbf{64.00} $\pm$ 5.43 & 0.00 $\pm$ 0.00 & 0.00 $\pm$ 0.00 & 0.00 $\pm$ 0.00 \\
\midrule
\textbf{AntMaze avg} & 17.21 & 19.71 & 19.33 & 18.58 & 50.71 & 57.17 & \textbf{78.42} & 0.00 & 0.00 & 18.12 \\
\midrule
pen-human-v1 & 71.03 $\pm$ 6.26 & 26.99 $\pm$ 9.60 & -3.88 $\pm$ 0.21 & 76.65 $\pm$ 11.71 & 13.71 $\pm$ 16.98 & 78.49 $\pm$ 8.21 & \textbf{103.16} $\pm$ 8.49 & 6.86 $\pm$ 5.93 & 5.07 $\pm$ 6.16 & 67.68 $\pm$ 5.48 \\
pen-cloned-v1 & 51.92 $\pm$ 15.15 & 46.67 $\pm$ 14.25 & 5.13 $\pm$ 5.28 &  85.72 $\pm$ 16.92 & 1.04 $\pm$ 6.62 & 83.42 $\pm$ 8.19 & \textbf{102.79} $\pm$ 7.84 & 31.35 $\pm$ 2.14 & 12.02 $\pm$ 1.75 & 64.43 $\pm$ 1.43 \\
pen-expert-v1 & 109.65 $\pm$ 7.28 & 114.96 $\pm$ 2.96 & 122.53 $\pm$ 21.27 &  \textbf{159.91} $\pm$ 1.87 & -1.41 $\pm$ 2.34 & 128.05 $\pm$ 9.21 & 152.16 $\pm$ 6.33 & 87.11 $\pm$ 48.95 & -1.55 $\pm$ 0.81 & 116.38 $\pm$ 1.27 \\
\midrule
door-human-v1 & 2.34 $\pm$ 4.00 & -0.13 $\pm$ 0.07 & -0.33 $\pm$ 0.01 &  2.39 $\pm$ 2.26 & \textbf{5.53} $\pm$ 1.31 & 3.26 $\pm$ 1.83 & -0.10 $\pm$ 0.01 & -0.38 $\pm$ 0.00 & -0.12 $\pm$ 0.13 & 4.44 $\pm$ 0.87 \\
door-cloned-v1 & -0.09 $\pm$ 0.03 & 0.29 $\pm$ 0.59 & -0.34 $\pm$ 0.01 &  -0.01 $\pm$ 0.01 & -0.33 $\pm$ 0.01 & 3.07 $\pm$ 1.75 & 0.06 $\pm$ 0.05 & -0.33 $\pm$ 0.00 & 2.66 $\pm$ 2.31 & \textbf{7.64} $\pm$ 3.26 \\
door-expert-v1 & 105.35 $\pm$ 0.09 & 104.04 $\pm$ 1.46 & -0.33 $\pm$ 0.01 &  104.57 $\pm$ 0.31 & -0.32 $\pm$ 0.02 & \textbf{106.65} $\pm$ 0.25 & 106.37 $\pm$ 0.29 & -0.33 $\pm$ 0.00 & 106.29 $\pm$ 1.73 & 104.87 $\pm$ 0.39 \\
\midrule
hammer-human-v1 & 3.03 $\pm$ 3.39 & -0.19 $\pm$ 0.02 & 1.02 $\pm$ 0.24 &  1.01 $\pm$ 0.51 & 0.14 $\pm$ 0.11 & \textbf{1.79} $\pm$ 0.80 & 0.24 $\pm$ 0.24 & 0.24 $\pm$ 0.00 & 0.28 $\pm$ 0.18 & 1.28 $\pm$ 0.15 \\
hammer-cloned-v1 & 0.55 $\pm$ 0.16 & 0.12 $\pm$ 0.08 & 0.25 $\pm$ 0.01 & 1.27 $\pm$ 2.11 & 0.30 $\pm$ 0.01 & 1.50 $\pm$ 0.69 & \textbf{5.00} $\pm$ 3.75 & 0.14 $\pm$ 0.09 & 0.19 $\pm$ 0.07 & 1.82 $\pm$ 0.55 \\
hammer-expert-v1 & 126.78 $\pm$ 0.64 & 121.75 $\pm$ 7.67 & 3.11 $\pm$ 0.03 &  127.08 $\pm$ 0.13 & 0.26 $\pm$ 0.01 & 128.68 $\pm$ 0.33 & \textbf{133.62} $\pm$ 0.27 & 25.13 $\pm$ 43.25 & 28.52 $\pm$ 49.00 & 117.45 $\pm$ 6.65 \\
\midrule
relocate-human-v1 & 0.04 $\pm$ 0.03 & -0.14 $\pm$ 0.08 & -0.29 $\pm$ 0.01 &  0.45 $\pm$ 0.53 & 0.06 $\pm$ 0.03 & 0.12 $\pm$ 0.04 & \textbf{0.16} $\pm$ 0.30 & -0.31 $\pm$ 0.01 & -0.17 $\pm$ 0.17 & 0.05 $\pm$ 0.01 \\
relocate-cloned-v1 & -0.06 $\pm$ 0.01 & -0.00 $\pm$ 0.02 & -0.30 $\pm$ 0.01 & -0.01 $\pm$ 0.03 & -0.29 $\pm$ 0.01 & 0.04 $\pm$ 0.01 & \textbf{1.66} $\pm$ 2.59 & -0.01 $\pm$ 0.10 & 0.17 $\pm$ 0.35 & 0.16 $\pm$ 0.09 \\
relocate-expert-v1 & 107.58 $\pm$ 1.20 & 97.90 $\pm$ 5.21 & -1.73 $\pm$ 0.96 & \textbf{109.52} $\pm$ 0.47 & -0.30 $\pm$ 0.02 & 106.11 $\pm$ 4.02 & 107.52 $\pm$ 2.28 & -0.36 $\pm$ 0.00 & 71.94 $\pm$ 18.37 & 104.28 $\pm$ 0.42 \\
\midrule
\textbf{Adroit avg} & 48.18 & 42.69 & 10.40 & 55.71 & 1.53 & 53.43 & \textbf{59.39} & 12.43 & 18.78 & 49.21 \\
\midrule
\textbf{Total avg} & 37.95 & 43.06 & 37.16 & 62.01 & 37.61 & 61.92 & \textbf{76.04} & 44.16 & 43.65 & 48.31 \\
		\bottomrule
		\end{tabular}
	\end{adjustbox}
\end{table}
\begin{table}[!ht]%[ht]

	\centering
	\small
	\caption{Normalized performance of the best trained policy on D4RL averaged over 4 random seeds.
% 	Normalized scores collected with the best trained policy on D4RL averaged over 4 seeds.
	}
% 	\vspace{0.2em}
	\label{tab:gym_max}
	\begin{adjustbox}{max width=\columnwidth}
		\begin{tabular}{l|rrrrrrrrrr}
			\toprule
			\multirow{2}{*}{\textbf{Task Name}} & \multirow{2}{*}{\textbf{BC}} & \multirow{2}{*}{\textbf{BC-10\%}} & \multirow{2}{*}{\textbf{TD3+BC}} & \multirow{2}{*}{\textbf{AWAC}} & \multirow{2}{*}{\textbf{CQL}} & \multirow{2}{*}{\textbf{IQL}}  & \multirow{2}{*}{\textbf{ReBRAC}} & \multirow{2}{*}{\textbf{SAC-$N$}} & \multirow{2}{*}{\textbf{EDAC}} & \multirow{2}{*}{\textbf{DT}} \\
			& & & & & & & \\
			\midrule
            halfcheetah-medium-v2 & 43.60 $\pm$ 0.14 & 43.90 $\pm$ 0.13 & 48.93 $\pm$ 0.11 & 50.81 $\pm$ 0.15 & 47.62 $\pm$ 0.03 & 48.84 $\pm$ 0.07 & 65.62 $\pm$ 0.46 & \textbf{72.21} $\pm$ 0.31 & 69.72 $\pm$ 0.92 & 42.73 $\pm$ 0.10 \\
halfcheetah-medium-replay-v2 & 40.52 $\pm$ 0.19 & 42.27 $\pm$ 0.46 & 45.84 $\pm$ 0.26 &  46.47 $\pm$ 0.26 & 46.43 $\pm$ 0.19 & 45.35 $\pm$ 0.08 & 52.22 $\pm$ 0.31 & \textbf{67.29} $\pm$ 0.34 & 66.55 $\pm$ 1.05 & 40.31 $\pm$ 0.28 \\
halfcheetah-medium-expert-v2 & 79.69 $\pm$ 3.10 & 94.11 $\pm$ 0.22 & 96.59 $\pm$ 0.87 &  96.83 $\pm$ 0.23 & 97.04 $\pm$ 0.17 & 95.38 $\pm$ 0.17 & 108.89 $\pm$ 1.20 & \textbf{111.73} $\pm$ 0.47 & 110.62 $\pm$ 1.04 & 93.40 $\pm$ 0.21 \\
\midrule
hopper-medium-v2 & 69.04 $\pm$ 2.90 & 73.84 $\pm$ 0.37 & 70.44 $\pm$ 1.18 & 95.42 $\pm$ 3.67 & 70.80 $\pm$ 1.98 & 80.46 $\pm$ 3.09 & 103.19 $\pm$ 0.16 & 101.79 $\pm$ 0.20 & \textbf{103.26} $\pm$ 0.14 & 69.42 $\pm$ 3.64 \\
hopper-medium-replay-v2 & 68.88 $\pm$ 10.33 & 90.57 $\pm$ 2.07 & 98.12 $\pm$ 1.16 & 101.47 $\pm$ 0.23  & 101.63 $\pm$ 0.55 & 102.69 $\pm$ 0.96 & 102.57 $\pm$ 0.45 & \textbf{103.83} $\pm$ 0.53 & 103.28 $\pm$ 0.49 & 88.74 $\pm$ 3.02 \\
hopper-medium-expert-v2 & 90.63 $\pm$ 10.98 & 113.13 $\pm$ 0.16 & 113.22 $\pm$ 0.43 &  \textbf{113.26} $\pm$ 0.49 & 112.84 $\pm$ 0.66 & \textbf{113.18} $\pm$ 0.38 & 113.16 $\pm$ 0.43 & 111.24 $\pm$ 0.15 & 111.80 $\pm$ 0.11 & 111.18 $\pm$ 0.21 \\
\midrule
walker2d-medium-v2 & 80.64 $\pm$ 0.91 & 82.05 $\pm$ 0.93 & 86.91 $\pm$ 0.28 & 85.86 $\pm$ 3.76 & 84.77 $\pm$ 0.20 & 87.58 $\pm$ 0.48 & 87.79 $\pm$ 0.19 & 90.17 $\pm$ 0.54 & \textbf{95.78} $\pm$ 1.07 & 74.70 $\pm$ 0.56 \\
walker2d-medium-replay-v2 & 48.41 $\pm$ 7.61 & 76.09 $\pm$ 0.40 & \textbf{91.17} $\pm$ 0.72 & 86.70 $\pm$ 0.94 & 89.39 $\pm$ 0.88 & 89.94 $\pm$ 0.93 & 91.11 $\pm$ 0.63 & 85.18 $\pm$ 1.63 & 89.69 $\pm$ 1.39 & 68.22 $\pm$ 1.20 \\
walker2d-medium-expert-v2 & 109.95 $\pm$ 0.62 & 109.90 $\pm$ 0.09 & 112.21 $\pm$ 0.06 &  113.40 $\pm$ 2.22 & 111.63 $\pm$ 0.38 & 113.06 $\pm$ 0.53 & 112.49 $\pm$ 0.18 & \textbf{116.93} $\pm$ 0.42 & 116.52 $\pm$ 0.75 & 108.71 $\pm$ 0.34 \\
\midrule
\textbf{Gym-MuJoCo avg} & 70.15 & 80.65 & 84.83 & 87.80 & 84.68 & 86.28 & 93.00 & 95.60 & \textbf{96.36} & 77.49 \\
\midrule
maze2d-umaze-v1 & 16.09 $\pm$ 0.87 & 22.49 $\pm$ 1.52 & 99.33 $\pm$ 16.16 &  136.96 $\pm$ 10.89 & 92.05 $\pm$ 13.66 & 50.92 $\pm$ 4.23 & \textbf{162.28} $\pm$ 1.79 & 153.12 $\pm$ 6.49 & 149.88 $\pm$ 1.97 & 63.83 $\pm$ 17.35 \\
maze2d-medium-v1 & 19.16 $\pm$ 1.24 & 27.64 $\pm$ 1.87 & 150.93 $\pm$ 3.89 & 152.73 $\pm$ 20.78  & 128.66 $\pm$ 5.44 & 122.69 $\pm$ 30.00 & 150.12 $\pm$ 4.48 & 93.80 $\pm$ 14.66 & \textbf{154.41} $\pm$ 1.58 & 68.14 $\pm$ 12.25 \\
maze2d-large-v1 & 20.75 $\pm$ 6.66 & 41.83 $\pm$ 3.64 & 197.64 $\pm$ 5.26 &  \textbf{227.31} $\pm$ 1.47 & 157.51 $\pm$ 7.32 & 162.25 $\pm$ 44.18 & 197.55 $\pm$ 5.82 & 207.51 $\pm$ 0.96 & 182.52 $\pm$ 2.68 & 50.25 $\pm$ 19.34 \\
\midrule
\textbf{Maze2d avg} & 18.67 & 30.65 & 149.30 & \textbf{172.33} & 126.07 & 111.95 & {169.98} & 151.48 & 162.27 & 60.74 \\
\midrule
antmaze-umaze-v2 & 68.50 $\pm$ 2.29 & 77.50 $\pm$ 1.50 & 98.50 $\pm$ 0.87 & 70.75 $\pm$ 8.84 & 94.75 $\pm$ 0.83 & 84.00 $\pm$ 4.06 & \textbf{100.00} $\pm$ 0.00 & 0.00 $\pm$ 0.00 & 42.50 $\pm$ 28.61 & 64.50 $\pm$ 2.06 \\
antmaze-umaze-diverse-v2 & 64.75 $\pm$ 4.32 & 63.50 $\pm$ 2.18 & 71.25 $\pm$ 5.76 & 81.50 $\pm$ 4.27 & 53.75 $\pm$ 2.05 & 79.50 $\pm$ 3.35 & \textbf{96.75} $\pm$ 2.28 & 0.00 $\pm$ 0.00 & 0.00 $\pm$ 0.00 & 60.50 $\pm$ 2.29 \\
antmaze-medium-play-v2 & 4.50 $\pm$ 1.12 & 6.25 $\pm$ 2.38 & 3.75 $\pm$ 1.30 &  25.00 $\pm$ 10.70 & 80.50 $\pm$ 3.35 & 78.50 $\pm$ 3.84 & \textbf{93.50} $\pm$ 2.60 & 0.00 $\pm$ 0.00 & 0.00 $\pm$ 0.00 & 0.75 $\pm$ 0.43 \\
antmaze-medium-diverse-v2 & 4.75 $\pm$ 1.09 & 16.50 $\pm$ 5.59 & 5.50 $\pm$ 1.50 & 10.75 $\pm$ 5.31 & 71.00 $\pm$ 4.53 & 83.50 $\pm$ 1.80 & \textbf{91.75} $\pm$ 2.05 & 0.00 $\pm$ 0.00 & 0.00 $\pm$ 0.00 & 0.50 $\pm$ 0.50 \\
antmaze-large-play-v2 & 0.50 $\pm$ 0.50 & 13.50 $\pm$ 9.76 & 1.25 $\pm$ 0.43 &  0.50 $\pm$ 0.50 & 34.75 $\pm$ 5.85 & 53.50 $\pm$ 2.50 & \textbf{68.75} $\pm$ 13.90 & 0.00 $\pm$ 0.00 & 0.00 $\pm$ 0.00 & 0.00 $\pm$ 0.00 \\
antmaze-large-diverse-v2 & 0.75 $\pm$ 0.43 & 6.25 $\pm$ 1.79 & 0.25 $\pm$ 0.43 & 0.00 $\pm$ 0.00 & 36.25 $\pm$ 3.34 & 53.00 $\pm$ 3.00 & \textbf{69.50} $\pm$ 7.26 & 0.00 $\pm$ 0.00 & 0.00 $\pm$ 0.00 & 0.00 $\pm$ 0.00 \\
\midrule
\textbf{AntMaze avg} & 23.96 & 30.58 & 30.08 & 31.42 & 61.83 & 72.00 & \textbf{86.71} & 0.00 & 7.08 & 21.04 \\
\midrule
pen-human-v1 & 99.69 $\pm$ 7.45 & 59.89 $\pm$ 8.03 & 9.95 $\pm$ 8.19 & 119.03 $\pm$ 6.55 & 58.91 $\pm$ 1.81 & 106.15 $\pm$ 10.28 & \textbf{127.28} $\pm$ 3.22 & 56.48 $\pm$ 7.17 & 35.84 $\pm$ 10.57 & 77.83 $\pm$ 2.30 \\
pen-cloned-v1 & 99.14 $\pm$ 12.27 & 83.62 $\pm$ 11.75 & 52.66 $\pm$ 6.33 &  125.78 $\pm$ 3.28 & 14.74 $\pm$ 2.31 & 114.05 $\pm$ 4.78 & \textbf{128.64} $\pm$ 7.15 & 52.69 $\pm$ 5.30 & 26.90 $\pm$ 7.85 & 71.17 $\pm$ 2.70 \\
pen-expert-v1 & 128.77 $\pm$ 5.88 & 134.36 $\pm$ 3.16 & 142.83 $\pm$ 7.72 & \textbf{162.53} $\pm$ 0.30 & 14.86 $\pm$ 4.07 & 140.01 $\pm$ 6.36 & 157.62 $\pm$ 0.26 & 116.43 $\pm$ 40.26 & 36.04 $\pm$ 4.60 & 119.49 $\pm$ 2.31 \\
\midrule
door-human-v1 & 9.41 $\pm$ 4.55 & 7.00 $\pm$ 6.77 & -0.11 $\pm$ 0.06 &  \textbf{17.70} $\pm$ 2.55 & 13.28 $\pm$ 2.77 & 13.52 $\pm$ 1.22 & 0.27 $\pm$ 0.43 & -0.10 $\pm$ 0.06 & 2.51 $\pm$ 2.26 & 7.36 $\pm$ 1.24 \\
door-cloned-v1 & 3.40 $\pm$ 0.95 & 10.37 $\pm$ 4.09 & -0.20 $\pm$ 0.11 & 10.53 $\pm$ 2.82 & -0.08 $\pm$ 0.13 & 9.02 $\pm$ 1.47 & 7.73 $\pm$ 6.80 & -0.21 $\pm$ 0.10 & \textbf{20.36} $\pm$ 1.11 & 11.18 $\pm$ 0.96 \\
door-expert-v1 & 105.84 $\pm$ 0.23 & 105.92 $\pm$ 0.24 & 4.49 $\pm$ 7.39 & 106.60 $\pm$ 0.27 & 59.47 $\pm$ 25.04 & 107.29 $\pm$ 0.37 & 106.78 $\pm$ 0.04 & 0.05 $\pm$ 0.02 & \textbf{109.22} $\pm$ 0.24 & 105.49 $\pm$ 0.09 \\
\midrule
hammer-human-v1 & 12.61 $\pm$ 4.87 & 6.23 $\pm$ 4.79 & 2.38 $\pm$ 0.14 &\textbf{ 16.95} $\pm$ 3.61 & 0.30 $\pm$ 0.05 & 6.86 $\pm$ 2.38 & 1.18 $\pm$ 0.15 & 0.25 $\pm$ 0.00 & 3.49 $\pm$ 2.17 & 1.68 $\pm$ 0.11 \\
hammer-cloned-v1 & 8.90 $\pm$ 4.04 & 8.72 $\pm$ 3.28 & 0.96 $\pm$ 0.30 &  10.74 $\pm$ 5.54 & 0.32 $\pm$ 0.03 & 11.63 $\pm$ 1.70 & \textbf{48.16} $\pm$ 6.20 & 12.67 $\pm$ 15.02 & 0.27 $\pm$ 0.01 & 2.74 $\pm$ 0.22 \\
hammer-expert-v1 & 127.89 $\pm$ 0.57 & 128.15 $\pm$ 0.66 & 33.31 $\pm$ 47.65 & 129.08 $\pm$ 0.26 & 0.93 $\pm$ 1.12 & 129.76 $\pm$ 0.37 & \textbf{134.74} $\pm$ 0.30 & 91.74 $\pm$ 47.77 & 69.44 $\pm$ 47.00 & 127.39 $\pm$ 0.10 \\
\midrule
relocate-human-v1 & 0.59 $\pm$ 0.27 & 0.16 $\pm$ 0.14 & -0.29 $\pm$ 0.01 & 1.77 $\pm$ 0.84 & 1.03 $\pm$ 0.20 & 1.22 $\pm$ 0.28 & \textbf{3.70} $\pm$ 2.34 & -0.18 $\pm$ 0.14 & 0.05 $\pm$ 0.02 & 0.08 $\pm$ 0.02 \\
relocate-cloned-v1 & 0.45 $\pm$ 0.31 & 0.74 $\pm$ 0.45 & -0.02 $\pm$ 0.04 & 0.39 $\pm$ 0.13 & -0.07 $\pm$ 0.02 & 1.78 $\pm$ 0.70 & \textbf{9.25} $\pm$ 2.56 & 0.10 $\pm$ 0.04 & 4.11 $\pm$ 1.39 & 0.34 $\pm$ 0.09 \\
relocate-expert-v1 & 110.31 $\pm$ 0.36 & 109.77 $\pm$ 0.60 & 0.23 $\pm$ 0.27 & \textbf{111.21} $\pm$ 0.32 & 0.03 $\pm$ 0.10 & 110.12 $\pm$ 0.82 & 111.14 $\pm$ 0.23 & -0.07 $\pm$ 0.08 & 98.32 $\pm$ 3.75 & 106.49 $\pm$ 0.30 \\
\midrule
\textbf{Adroit avg} & 58.92 & 54.58 & 20.51 & 67.69  & 13.65 & 62.62 & \textbf{69.71} & 27.49 & 33.88 & 52.60 \\
\midrule
\textbf{Total avg} & 51.27 & 55.21 & 54.60 &  76.93 & 55.84 & 76.53 & \textbf{90.12} & 54.82 & 60.10 & 54.57 \\
		\bottomrule
		\end{tabular}
	\end{adjustbox}
\end{table}

Based on these results, we make several valuable observations. First, ReBRAC, IQL and AWAC are the most competitive baselines in offline setup on average. Note that AWAC is often omitted in recent works.

\hypothesis{Observation 1}{ReBRAC, IQL and AWAC are the strongest offline baselines on average.}

Second, EDAC outperforms all other algorithms on Gym-MuJoCo by a significant margin, and to our prior knowledge, there are still no algorithms that perform much better on these tasks. SAC-N shows the best performance on Maze2d tasks. However, simultaneously, SAC-N and EDAC cannot solve AntMaze tasks and perform poorly in the Adroit domain. 

\hypothesis{Observation 2}{SAC-N and EDAC are the strongest baselines for Gym-MuJoCo and Maze2d, but they perform poorly on both AntMaze and Adroit domains.}

Third, during our experiments, we observed that the hyperparameters proposed for CQL in \citet{kumar2020conservative} do not perform as well as claimed on most tasks. CQL is extremely sensitive to the choice of hyperparameters, and we had to tune them a lot to make it work on each domain (see \autoref{table:cql_hyp}). For example, AntMaze requires five hidden layers for the critic networks, while other tasks' performance suffers with this number of layers. The issue of sensitivity\footnote{See also \url{https://github.com/aviralkumar2907/CQL/issues/9}, \url{https://github.com/tinkoff-ai/CORL/issues/14} and \url{https://github.com/young-geng/CQL/issues/5}}  was already mentioned in prior works as well \citep{an2021uncertainty, ghasemipour2022so}.

\hypothesis{Observation 3}{CQL is extremely sensitive to the choice of hyperparameters and implementation details.}
\vspace{0.05in}

Fourth, we also observe that the hyperparameters do not always work the same way when transferring between Deep Learning frameworks \footnote{\url{https://github.com/tinkoff-ai/CORL/issues/33}}. Our implementations of IQL and CQL use PyTorch, but the parameters from reference JAX implementations sometimes strongly underperform (e.g., IQL on Hopper tasks and CQL on Adroit).

\hypothesis{Observation 4}{Hyperparameters are not always transferable between Deep Learning frameworks.}

\subsection{Offline-to-Online}
We also implement the following algorithms in offline-to-online setup: CQL~\citep{kumar2020conservative}, IQL~\citep{kostrikov2021offline}, AWAC~\citep{nair2020accelerating}, SPOT~\citep{wu2022supported} Cal-QL~\citep{nakamoto2023cal}, ReBRAC~\citep{tarasov2023revisiting}. Inspired by \citet{nakamoto2023cal}, we evaluate algorithms on AntMaze and Adroit Cloned datasets\footnote{Note, \citet{nakamoto2023cal} used modified Cloned datasets while we employ original data from D4RL because these datasets are more common to for benchmarking.}. Each algorithm is trained offline over 1 million steps and tuned using online transitions over another 1 million steps. The AntMaze tasks are evaluated using 100 episodes, while the Adroit tasks are tested with ten episodes.

\begin{table}[!ht]%[ht]

	\centering
	\small
	\caption{Normalized performance of algorithms after offline pretraining and online finetuning on D4RL averaged over 4 random seeds.
% 	Normalized scores collected with the best trained policy on D4RL averaged over 4 seeds.
	}
% 	\vspace{0.2em}
	\label{tab:online}
	\begin{adjustbox}{max width=\columnwidth}
		\begin{tabular}{l|cccccc}
			\toprule
			\multirow{2}{*}{\textbf{Task Name}} & \multirow{2}{*}{\textbf{AWAC}} & \multirow{2}{*}{\textbf{CQL}} & \multirow{2}{*}{\textbf{IQL}} & \multirow{2}{*}{\textbf{SPOT}} & \multirow{2}{*}{\textbf{Cal-QL}} & \multirow{2}{*}{\textbf{ReBRAC}}\\
			& & & & & \\
			\midrule
antmaze-umaze-v2 & 52.75 $\pm$ 8.67 $\to$  98.75 $\pm$ 1.09 & 94.00 $\pm$ 1.58 $\to$  99.50 $\pm$ 0.87 & 77.00 $\pm$ 0.71 $\to$  96.50 $\pm$ 1.12 & 91.00 $\pm$ 2.55 $\to$  99.50 $\pm$ 0.50 & 76.75 $\pm$ 7.53 $\to$  \textbf{99.75} $\pm$ 0.43  & 98.00 $\pm$ 1.82 $\to$ 74.75 $\pm$ 49.17 \\
antmaze-umaze-diverse-v2 & 56.00 $\pm$ 2.74 $\to$  0.00 $\pm$ 0.00 & 9.50 $\pm$ 9.91 $\to$  \textbf{99.00} $\pm$ 1.22 & 59.50 $\pm$ 9.55 $\to$  63.75 $\pm$ 25.02 & 36.25 $\pm$ 2.17 $\to$  95.00 $\pm$ 3.67 & 32.00 $\pm$ 27.79 $\to$  98.50 $\pm$ 1.12  & 73.75 $\pm$ 15.32 $\to$ 98.0 $\pm$ 3.36 \\
antmaze-medium-play-v2 & 0.00 $\pm$ 0.00 $\to$  0.00 $\pm$ 0.00 & 59.00 $\pm$ 11.18 $\to$  97.75 $\pm$ 1.30 & 71.75 $\pm$ 2.95 $\to$  89.75 $\pm$ 1.09 & 67.25 $\pm$ 10.47 $\to$  97.25 $\pm$ 1.30 & 71.75 $\pm$ 3.27 $\to$  \textbf{98.75} $\pm$ 1.64  & 87.5 $\pm$ 4.35 $\to$ 98.0 $\pm$ 1.82\\
antmaze-medium-diverse-v2 & 0.00 $\pm$ 0.00 $\to$  0.00 $\pm$ 0.00 & 63.50 $\pm$ 6.84 $\to$  97.25 $\pm$ 1.92 & 64.25 $\pm$ 1.92 $\to$  92.25 $\pm$ 2.86 & 73.75 $\pm$ 7.29 $\to$  94.50 $\pm$ 1.66 & 62.00 $\pm$ 4.30 $\to$  \textbf{98.25} $\pm$ 1.48  & 85.25 $\pm$ 2.5 $\to$  98.75 $\pm$ 0.5 \\
antmaze-large-play-v2 & 0.00 $\pm$ 0.00 $\to$  0.00 $\pm$ 0.00 & 28.75 $\pm$ 7.76 $\to$  88.25 $\pm$ 2.28 & 38.50 $\pm$ 8.73 $\to$  64.50 $\pm$ 17.04 & 31.50 $\pm$ 12.58 $\to$  87.00 $\pm$ 3.24 & 31.75 $\pm$ 8.87 $\to$  \textbf{97.25} $\pm$ 1.79  & 68.5 $\pm$ 7.1 $\to$ 31.5 $\pm$ 38.75 \\
antmaze-large-diverse-v2 & 0.00 $\pm$ 0.00 $\to$  0.00 $\pm$ 0.00 & 35.50 $\pm$ 3.64 $\to$  \textbf{91.75} $\pm$ 3.96 & 26.75 $\pm$ 3.77 $\to$  64.25 $\pm$ 4.15 & 17.50 $\pm$ 7.26 $\to$  81.00 $\pm$ 14.14 & 44.00 $\pm$ 8.69 $\to$  91.50 $\pm$ 3.91  & 67.0 $\pm$ 12.24 $\to$ 72.25 $\pm$ 48.18 \\
\midrule
\textbf{AntMaze avg} & 18.12 $\to$  16.46 (\textcolor{red}{-1.66}) & 48.38 $\to$  95.58 (\textcolor{dgreen}{+47.20}) & 56.29 $\to$  78.50 (\textcolor{dgreen}{+22.21}) & 52.88 $\to$  92.38 (\textcolor{dgreen}{+39.50}) &  53.04 $\to$ \textbf{97.33} (\textcolor{dgreen}{+24.29}) & 79.99 $\to$ 78.87(\textcolor{red}{-1.11}) \\
\midrule
pen-cloned-v1 & 88.66 $\pm$ 15.10 $\to$  86.82 $\pm$ 11.12 & -2.76 $\pm$ 0.08 $\to$  -1.28 $\pm$ 2.16 & 84.19 $\pm$ 3.96 $\to$  \textbf{102.02} $\pm$ 20.75 & 6.19 $\pm$ 5.21 $\to$  43.63 $\pm$ 20.09 & -2.66 $\pm$ 0.04 $\to$  -2.68 $\pm$ 0.12  & 74.04 $\pm$ 13.82 $\to$ 138.15 $\pm$ 3.71 \\
door-cloned-v1 & 0.93 $\pm$ 1.66 $\to$  0.01 $\pm$ 0.00 & -0.33 $\pm$ 0.01 $\to$  -0.33 $\pm$ 0.01 & 1.19 $\pm$ 0.93 $\to$  \textbf{20.34} $\pm$ 9.32 & -0.21 $\pm$ 0.14 $\to$  0.02 $\pm$ 0.31 & -0.33 $\pm$ 0.01 $\to$  -0.33 $\pm$ 0.01  & 0.06 $\pm$  0.04 $\to$ 102.38 $\pm$ 9.54 \\
hammer-cloned-v1 & 1.80 $\pm$ 3.01 $\to$  0.24 $\pm$ 0.04 & 0.56 $\pm$ 0.55 $\to$  2.85 $\pm$ 4.81 & 1.35 $\pm$ 0.32 $\to$  \textbf{57.27} $\pm$ 28.49 & 3.97 $\pm$ 6.39 $\to$  3.73 $\pm$ 4.99 & 0.25 $\pm$ 0.04 $\to$  0.17 $\pm$ 0.17  & 6.53 $\pm$ 3.86 $\to$ 124.65 $\pm$ 8.51 \\
relocate-cloned-v1 & -0.04 $\pm$ 0.04 $\to$  -0.04 $\pm$ 0.01 & -0.33 $\pm$ 0.01 $\to$  -0.33 $\pm$ 0.01 & 0.04 $\pm$ 0.04 $\to$  \textbf{0.32} $\pm$ 0.38 & -0.24 $\pm$ 0.01 $\to$  -0.15 $\pm$ 0.05 & -0.31 $\pm$ 0.05 $\to$  -0.31 $\pm$ 0.04  & 0.69 $\pm$ 0.71 $\to$ 6.96 $\pm$ 5.3 \\
\midrule
\textbf{Adroit Avg} & 22.84 $\to$  21.76 (\textcolor{red}{-1.08}) & -0.72 $\to$  0.22 (\textcolor{dgreen}{+0.94}) & 21.69 $\to$  \textbf{44.99} (\textcolor{dgreen}{+23.3}) & 2.43 $\to$  11.81 (\textcolor{dgreen}{+9.38}) & -0.76 $\to$    -0.79 (\textcolor{red}{-0.03})  & 20.33 $\to$ 93.03 (\textcolor{dgreen}{+72.7}) \\
\midrule
\textbf{Total avg} & 20.01 $\to$  18.58 (\textcolor{red}{-1.43}) & 28.74 $\to$  57.44 (\textcolor{dgreen}{+28.7}) & 42.45 $\to$  \textbf{65.10} (\textcolor{dgreen}{+22.65}) & 32.70 $\to$  60.15 (\textcolor{dgreen}{+27.45}) & 31.52 $\to$  58.08 (\textcolor{dgreen}{+26.56})  & 56.12 $\to$ 84.53 (\textcolor{dgreen}{+28.41}) \\
         	\bottomrule
		\end{tabular}
	\end{adjustbox}
\end{table}

\begin{table}[!ht]%[ht]

	\centering
	\small
	\caption{Cumulative regret of online finetuning calculated as $1 - \text{\textit{average success rate}}$ averaged over 4 random seeds.
% 	Normalized scores collected with the best trained policy on D4RL averaged over 4 seeds.
	}
% 	\vspace{0.2em}
	\label{tab:regret}
	\begin{adjustbox}{max width=\columnwidth}
		\begin{tabular}{l|rrrrrr}
			\toprule
			\multirow{2}{*}{\textbf{Task Name}} & \multirow{2}{*}{\textbf{AWAC}} & \multirow{2}{*}{\textbf{CQL}} & \multirow{2}{*}{\textbf{IQL}} & \multirow{2}{*}{\textbf{SPOT}} & \multirow{2}{*}{\textbf{Cal-QL}} & \multirow{2}{*}{\textbf{ReBRAC}}\\
			& & & & & \\
			\midrule
   antmaze-umaze-v2 & 0.04 $\pm$ 0.01 & 0.02 $\pm$ 0.00 & 0.07 $\pm$ 0.00 & 0.02 $\pm$ 0.00 & \textbf{0.01} $\pm$ 0.00 & 0.10 $\pm$ 0.20\\
antmaze-umaze-diverse-v2 & 0.88 $\pm$ 0.01 & 0.09 $\pm$ 0.01 & 0.43 $\pm$ 0.11 & 0.22 $\pm$ 0.07 & \textbf{0.05} $\pm$ 0.01 & 0.04 $\pm$ 0.02 \\
antmaze-medium-play-v2 & 1.00 $\pm$ 0.00 & 0.08 $\pm$ 0.01 & 0.09 $\pm$ 0.01 & 0.06 $\pm$ 0.00 & \textbf{0.04} $\pm$ 0.01 & 0.02 $\pm$ 0.00\\
antmaze-medium-diverse-v2 & 1.00 $\pm$ 0.00 & 0.08 $\pm$ 0.00 & 0.10 $\pm$ 0.01 & 0.05 $\pm$ 0.01 & \textbf{0.04} $\pm$ 0.01 & 0.03 $\pm$ 0.00 \\
antmaze-large-play-v2 & 1.00 $\pm$ 0.00 & 0.21 $\pm$ 0.02 & 0.34 $\pm$ 0.05 & 0.29 $\pm$ 0.07 & \textbf{0.13} $\pm$ 0.02 & 0.14 $\pm$ 0.05 \\
antmaze-large-diverse-v2 & 1.00 $\pm$ 0.00 & 0.21 $\pm$ 0.03 & 0.41 $\pm$ 0.03 & 0.23 $\pm$ 0.08 & \textbf{0.13} $\pm$ 0.02 &  0.29 $\pm$ 0.45 \\
\midrule
\textbf{AntMaze avg} & 0.82 & 0.11 & 0.24 & 0.15 & \textbf{0.07} & 0.10 \\
\midrule
pen-cloned-v1 & 0.46 $\pm$ 0.02 & 0.97 $\pm$ 0.00 & \textbf{0.37} $\pm$ 0.01 & 0.58 $\pm$ 0.02 & 0.98 $\pm$ 0.01 & 0.08  $\pm$ 0.01 \\
door-cloned-v1 & 1.00 $\pm$ 0.00 & 1.00 $\pm$ 0.00 & \textbf{0.83} $\pm$ 0.03 & 0.99 $\pm$ 0.01 & 1.00 $\pm$ 0.00 & 0.18 $\pm$ 0.06 \\
hammer-cloned-v1 & 1.00 $\pm$ 0.00 & 1.00 $\pm$ 0.00 & \textbf{0.65} $\pm$ 0.10 & 0.98 $\pm$ 0.01 & 1.00 $\pm$ 0.00 & 0.12 $\pm$  0.03 \\
relocate-cloned-v1 & 1.00 $\pm$ 0.00 & 1.00 $\pm$ 0.00 & 1.00 $\pm$ 0.00 & 1.00 $\pm$ 0.00 & 1.00 $\pm$ 0.00 & 0.9 $\pm$ 0.06 \\
\midrule
\textbf{Adroit avg} &  0.86 & 0.99 & \textbf{0.71} & 0.89 & 0.99 & 0.32\\
\midrule
\textbf{Total avg}  & 0.84 & 0.47 & \textbf{0.43} & 0.44 & 0.44 & 0.19\\
         	\bottomrule
		\end{tabular}
	\end{adjustbox}
\end{table}

\begin{figure}[ht]
\centering
\captionsetup{justification=centering}
     \centering
        \begin{subfigure}[b]{0.49\textwidth}
         \centering
         \includegraphics[width=1.0\textwidth]{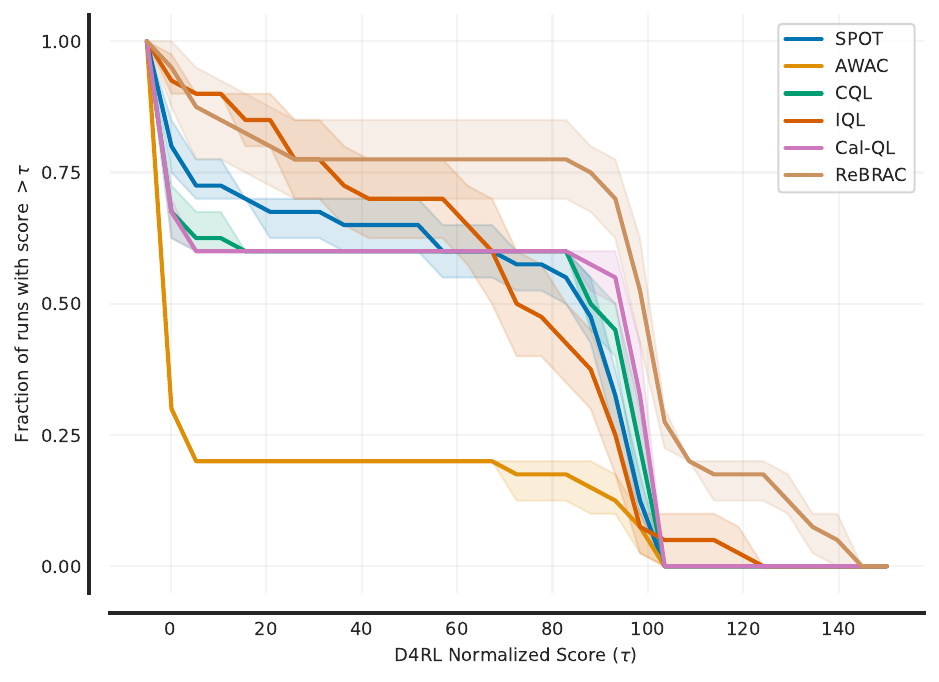}
         \caption{}
         \label{fig:perf_prof_online}
        \end{subfigure}
        \begin{subfigure}[b]{0.49\textwidth}
         \centering
         \includegraphics[width=1.0\textwidth]{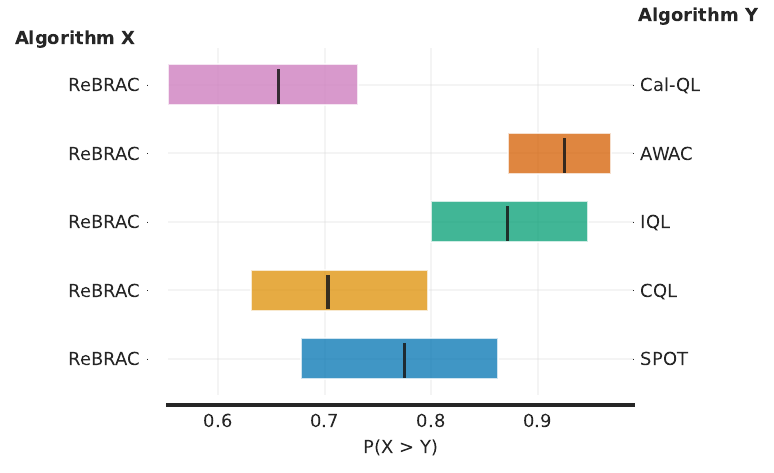}
         \caption{}
         \label{fig:prob_improve_online}
        \end{subfigure}
        \caption{{(a) Performance profiles after online tuning (b) Probability of improvement of ReBRAC to other algorithms after online tuning. The curves \citep{agarwal2021deep} are for D4RL benchmark spanning AntMaze and Adroit cloned datasets.}}
        \label{fig:curves_online}
\end{figure}

The scores, normalized after the offline stage and online tuning, are reported in \autoref{tab:online}. We also provide finetuning cumulative regret proposed by \citet{nakamoto2023cal} in \autoref{tab:regret}. Cumulative regret is calculated as $(1 - \text{\textit{average success rate})}$\footnote{As specified by the authors: \url{https://github.com/nakamotoo/Cal-QL/issues/1}}. It is bounded between 0 and 1, indicating the range of possible values. Lower values of cumulative regret indicate better algorithm efficiency. The performance profiles and probability of improvement of ReBRAC over other algorithms after online finetuning are presented in \autoref{fig:curves_online}.

AWAC, initially proposed for finetuning purposes, appeared to be the worst of the considered algorithms, where the score is improved only on the most straightforward antmaze-umaze-v2 dataset. At the same time, on other datasets, performances either stay the same or even drop.

\hypothesis{Observation 5}{AWAC does not benefit from online tuning on the considered tasks.}
\vspace{0.05in}

Cal-QL was proposed as a modification of CQL, which is expected to work better in offline-to-online setting. However, in our experiments, after finetuning CQL obtained scores which are not very different from Cal-QL. At the same time, we could not make both algorithms solve Adroit tasks\footnote{The issues are Observations 3 and 4. Additional hyperparameters search is needed.}. 

\hypothesis{Observation 6}{There is no big difference between CQL and Cal-QL. On AntMaze, these algorithms perform the best but work poorly on Adroit.}

% \newpage
IQL starts with good offline scores on AntMaze, but it is less efficient in finetuning than other algorithms except for AWAC. At the same time, IQL and ReBRAC are the only algorithms that notably improve its scores after tuning on Adroit tasks, making them the most competitive offline-to-online baselines considering the average score.

\hypothesis{Observation 7}{Considering offline and offline-to-online results, IQL and ReBRAC appear to be the strongest baselines on average.}

\section{Conclusion}
This paper introduced CORL, a single-file implementation library for offline and offline-to-online reinforcement learning algorithms with configuration files and advanced metrics tracking support. In total, we provided implementations of ten offline and six offline-to-online algorithms. All implemented approaches were benchmarked on D4RL datasets, closely matching (sometimes overperforming) the reference results, if available. 
Focusing on implementation clarity and reproducibility, we hope that CORL will help RL practitioners in their research and applications. 

This study's benchmarking results and observations are intended to serve as references for future offline reinforcement learning research and its practical applications. By sharing comprehensive logs, researchers can readily access and utilize our results without having to re-run any of our experiments, ensuring that the results are replicable.

% \clearpage
\bibliography{_bib}

\begin{thebibliography}{42}
\providecommand{\natexlab}[1]{#1}
\providecommand{\url}[1]{\texttt{#1}}
\expandafter\ifx\csname urlstyle\endcsname\relax
  \providecommand{\doi}[1]{doi: #1}\else
  \providecommand{\doi}{doi: \begingroup \urlstyle{rm}\Url}\fi

\bibitem[Abadi et~al.(2015)Abadi, Agarwal, Barham, Brevdo, Chen, Citro, Corrado, Davis, Dean, Devin, Ghemawat, Goodfellow, Harp, Irving, Isard, Jia, Jozefowicz, Kaiser, Kudlur, Levenberg, Man\'{e}, Monga, Moore, Murray, Olah, Schuster, Shlens, Steiner, Sutskever, Talwar, Tucker, Vanhoucke, Vasudevan, Vi\'{e}gas, Vinyals, Warden, Wattenberg, Wicke, Yu, and Zheng]{tensorflow2015-whitepaper}
Mart\'{i}n Abadi, Ashish Agarwal, Paul Barham, Eugene Brevdo, Zhifeng Chen, Craig Citro, Greg~S. Corrado, Andy Davis, Jeffrey Dean, Matthieu Devin, Sanjay Ghemawat, Ian Goodfellow, Andrew Harp, Geoffrey Irving, Michael Isard, Yangqing Jia, Rafal Jozefowicz, Lukasz Kaiser, Manjunath Kudlur, Josh Levenberg, Dandelion Man\'{e}, Rajat Monga, Sherry Moore, Derek Murray, Chris Olah, Mike Schuster, Jonathon Shlens, Benoit Steiner, Ilya Sutskever, Kunal Talwar, Paul Tucker, Vincent Vanhoucke, Vijay Vasudevan, Fernanda Vi\'{e}gas, Oriol Vinyals, Pete Warden, Martin Wattenberg, Martin Wicke, Yuan Yu, and Xiaoqiang Zheng.
\newblock {TensorFlow}: Large-scale machine learning on heterogeneous systems, 2015.
\newblock URL \url{https://www.tensorflow.org/}.
\newblock Software available from tensorflow.org.

\bibitem[Agarwal et~al.(2021)Agarwal, Schwarzer, Castro, Courville, and Bellemare]{agarwal2021deep}
Rishabh Agarwal, Max Schwarzer, Pablo~Samuel Castro, Aaron Courville, and Marc~G Bellemare.
\newblock Deep reinforcement learning at the edge of the statistical precipice.
\newblock \emph{Advances in Neural Information Processing Systems}, 2021.

\bibitem[An et~al.(2021)An, Moon, Kim, and Song]{an2021uncertainty}
Gaon An, Seungyong Moon, Jang-Hyun Kim, and Hyun~Oh Song.
\newblock Uncertainty-based offline reinforcement learning with diversified q-ensemble.
\newblock \emph{Advances in neural information processing systems}, 34:\penalty0 7436--7447, 2021.

\bibitem[Arakelyan et~al.(2020)Arakelyan, Soghomonyan, and {The Aim team}]{Arakelyan_Aim_2020}
Gor Arakelyan, Gevorg Soghomonyan, and {The Aim team}.
\newblock {Aim}, 6 2020.
\newblock URL \url{https://github.com/aimhubio/aim}.

\bibitem[Biewald(2020)]{wandb}
Lukas Biewald.
\newblock Experiment tracking with weights and biases, 2020.
\newblock URL \url{https://www.wandb.com/}.
\newblock Software available from wandb.com.

\bibitem[Castro et~al.(2018)Castro, Moitra, Gelada, Kumar, and Bellemare]{castro2018dopamine}
Pablo~Samuel Castro, Subhodeep Moitra, Carles Gelada, Saurabh Kumar, and Marc~G Bellemare.
\newblock Dopamine: A research framework for deep reinforcement learning.
\newblock \emph{arXiv preprint arXiv:1812.06110}, 2018.

\bibitem[Chen et~al.(2021)Chen, Lu, Rajeswaran, Lee, Grover, Laskin, Abbeel, Srinivas, and Mordatch]{chen2021decision}
Lili Chen, Kevin Lu, Aravind Rajeswaran, Kimin Lee, Aditya Grover, Misha Laskin, Pieter Abbeel, Aravind Srinivas, and Igor Mordatch.
\newblock Decision transformer: Reinforcement learning via sequence modeling.
\newblock \emph{Advances in neural information processing systems}, 34:\penalty0 15084--15097, 2021.

\bibitem[Chen et~al.(2022)Chen, Xu, Gatto, Jain, Kumar, and Chi]{Chen2022OffPolicyAF}
Minmin Chen, Can Xu, Vince Gatto, Devanshu Jain, Aviral Kumar, and Ed~H. Chi.
\newblock Off-policy actor-critic for recommender systems.
\newblock \emph{Proceedings of the 16th ACM Conference on Recommender Systems}, 2022.

\bibitem[Dhariwal et~al.(2017)Dhariwal, Hesse, Klimov, Nichol, Plappert, Radford, Schulman, Sidor, Wu, and Zhokhov]{baselines}
Prafulla Dhariwal, Christopher Hesse, Oleg Klimov, Alex Nichol, Matthias Plappert, Alec Radford, John Schulman, Szymon Sidor, Yuhuai Wu, and Peter Zhokhov.
\newblock Openai baselines.
\newblock \url{https://github.com/openai/baselines}, 2017.

\bibitem[Diehl et~al.(2021)Diehl, Sievernich, Kr{\"u}ger, Hoffmann, and Bertram]{Diehl2021UMBRELLAUM}
Christopher~P. Diehl, Timo Sievernich, Martin Kr{\"u}ger, Frank Hoffmann, and Torsten Bertram.
\newblock Umbrella: Uncertainty-aware model-based offline reinforcement learning leveraging planning.
\newblock \emph{ArXiv}, abs/2111.11097, 2021.

\bibitem[Duan et~al.(2016)Duan, Chen, Houthooft, Schulman, and Abbeel]{10.5555/3045390.3045531}
Yan Duan, Xi~Chen, Rein Houthooft, John Schulman, and Pieter Abbeel.
\newblock Benchmarking deep reinforcement learning for continuous control.
\newblock In \emph{Proceedings of the 33rd International Conference on International Conference on Machine Learning - Volume 48}, ICML'16, pp.\  1329–1338. JMLR.org, 2016.

\bibitem[Engstrom et~al.(2020)Engstrom, Ilyas, Santurkar, Tsipras, Janoos, Rudolph, and Madry]{Engstrom2020ImplementationMI}
Logan Engstrom, Andrew Ilyas, Shibani Santurkar, Dimitris Tsipras, Firdaus Janoos, L.~Rudolph, and Aleksander Madry.
\newblock Implementation matters in deep rl: A case study on ppo and trpo.
\newblock In \emph{ICLR}, 2020.

\bibitem[Fu et~al.(2020)Fu, Kumar, Nachum, Tucker, and Levine]{d4rl}
Justin Fu, Aviral Kumar, Ofir Nachum, George Tucker, and Sergey Levine.
\newblock D4rl: Datasets for deep data-driven reinforcement learning.
\newblock \emph{arXiv preprint arXiv:2004.07219}, 2020.

\bibitem[Fujimoto \& Gu(2021)Fujimoto and Gu]{fujimoto2021minimalist}
Scott Fujimoto and Shixiang~Shane Gu.
\newblock A minimalist approach to offline reinforcement learning.
\newblock \emph{Advances in neural information processing systems}, 34:\penalty0 20132--20145, 2021.

\bibitem[Fujita et~al.(2021)Fujita, Nagarajan, Kataoka, and Ishikawa]{JMLR:v22:20-376}
Yasuhiro Fujita, Prabhat Nagarajan, Toshiki Kataoka, and Takahiro Ishikawa.
\newblock Chainerrl: A deep reinforcement learning library.
\newblock \emph{Journal of Machine Learning Research}, 22\penalty0 (77):\penalty0 1--14, 2021.
\newblock URL \url{http://jmlr.org/papers/v22/20-376.html}.

\bibitem[garage contributors(2019)]{garage}
The garage contributors.
\newblock Garage: A toolkit for reproducible reinforcement learning research.
\newblock \url{https://github.com/rlworkgroup/garage}, 2019.

\bibitem[Gauci et~al.(2018)Gauci, Conti, Liang, Virochsiri, Chen, He, Kaden, Narayanan, and Ye]{gauci2018horizon}
Jason Gauci, Edoardo Conti, Yitao Liang, Kittipat Virochsiri, Zhengxing Chen, Yuchen He, Zachary Kaden, Vivek Narayanan, and Xiaohui Ye.
\newblock Horizon: Facebook's open source applied reinforcement learning platform.
\newblock \emph{arXiv preprint arXiv:1811.00260}, 2018.

\bibitem[Ghasemipour et~al.(2022)Ghasemipour, Gu, and Nachum]{ghasemipour2022so}
Kamyar Ghasemipour, Shixiang~Shane Gu, and Ofir Nachum.
\newblock Why so pessimistic? estimating uncertainties for offline rl through ensembles, and why their independence matters.
\newblock \emph{Advances in Neural Information Processing Systems}, 35:\penalty0 18267--18281, 2022.

\bibitem[Henderson et~al.(2018)Henderson, Islam, Bachman, Pineau, Precup, and Meger]{10.5555/3504035.3504427}
Peter Henderson, Riashat Islam, Philip Bachman, Joelle Pineau, Doina Precup, and David Meger.
\newblock Deep reinforcement learning that matters.
\newblock In \emph{Proceedings of the Thirty-Second AAAI Conference on Artificial Intelligence and Thirtieth Innovative Applications of Artificial Intelligence Conference and Eighth AAAI Symposium on Educational Advances in Artificial Intelligence}, AAAI'18/IAAI'18/EAAI'18. AAAI Press, 2018.
\newblock ISBN 978-1-57735-800-8.

\bibitem[Hill et~al.(2018)Hill, Raffin, Ernestus, Gleave, Kanervisto, Traore, Dhariwal, Hesse, Klimov, Nichol, Plappert, Radford, Schulman, Sidor, and Wu]{stable-baselines}
Ashley Hill, Antonin Raffin, Maximilian Ernestus, Adam Gleave, Anssi Kanervisto, Rene Traore, Prafulla Dhariwal, Christopher Hesse, Oleg Klimov, Alex Nichol, Matthias Plappert, Alec Radford, John Schulman, Szymon Sidor, and Yuhuai Wu.
\newblock Stable baselines.
\newblock \url{https://github.com/hill-a/stable-baselines}, 2018.

\bibitem[Huang et~al.(2021)Huang, Dossa, Ye, and Braga]{huang2021cleanrl}
Shengyi Huang, Rousslan Fernand~Julien Dossa, Chang Ye, and Jeff Braga.
\newblock Cleanrl: High-quality single-file implementations of deep reinforcement learning algorithms.
\newblock \emph{arXiv preprint arXiv:2111.08819}, 2021.

\bibitem[Keng \& Graesser(2017)Keng and Graesser]{kenggraesser2017slmlab}
Wah~Loon Keng and Laura Graesser.
\newblock Slm lab.
\newblock \url{https://github.com/kengz/SLM-Lab}, 2017.

\bibitem[Kingma \& Ba(2014)Kingma and Ba]{adam}
Diederik~P Kingma and Jimmy Ba.
\newblock Adam: A method for stochastic optimization.
\newblock \emph{arXiv preprint arXiv:1412.6980}, 2014.

\bibitem[Kolesnikov \& Hrinchuk(2019)Kolesnikov and Hrinchuk]{1903.00027}
Sergey Kolesnikov and Oleksii Hrinchuk.
\newblock Catalyst.rl: A distributed framework for reproducible rl research, 2019.
\newblock URL \url{https://arxiv.org/abs/1903.00027}.

\bibitem[Kostrikov et~al.(2021)Kostrikov, Nair, and Levine]{kostrikov2021offline}
Ilya Kostrikov, Ashvin Nair, and Sergey Levine.
\newblock Offline reinforcement learning with implicit q-learning.
\newblock \emph{arXiv preprint arXiv:2110.06169}, 2021.

\bibitem[Kumar et~al.(2020)Kumar, Zhou, Tucker, and Levine]{kumar2020conservative}
Aviral Kumar, Aurick Zhou, George Tucker, and Sergey Levine.
\newblock Conservative q-learning for offline reinforcement learning.
\newblock \emph{Advances in Neural Information Processing Systems}, 33:\penalty0 1179--1191, 2020.

\bibitem[Kumar et~al.(2021)Kumar, Singh, Tian, Finn, and Levine]{kumar2021a}
Aviral Kumar, Anikait Singh, Stephen Tian, Chelsea Finn, and Sergey Levine.
\newblock A workflow for offline model-free robotic reinforcement learning.
\newblock In \emph{5th Annual Conference on Robot Learning}, 2021.
\newblock URL \url{https://openreview.net/forum?id=fy4ZBWxYbIo}.

\bibitem[Kurenkov \& Kolesnikov(2022)Kurenkov and Kolesnikov]{pmlr-v162-kurenkov22a}
Vladislav Kurenkov and Sergey Kolesnikov.
\newblock Showing your offline reinforcement learning work: Online evaluation budget matters.
\newblock In Kamalika Chaudhuri, Stefanie Jegelka, Le~Song, Csaba Szepesvari, Gang Niu, and Sivan Sabato (eds.), \emph{Proceedings of the 39th International Conference on Machine Learning}, volume 162 of \emph{Proceedings of Machine Learning Research}, pp.\  11729--11752. PMLR, 17--23 Jul 2022.
\newblock URL \url{https://proceedings.mlr.press/v162/kurenkov22a.html}.

\bibitem[Levine et~al.(2020)Levine, Kumar, Tucker, and Fu]{levine2020offline}
Sergey Levine, Aviral Kumar, George Tucker, and Justin Fu.
\newblock Offline reinforcement learning: Tutorial, review, and perspectives on open problems.
\newblock \emph{arXiv preprint arXiv:2005.01643}, 2020.

\bibitem[Liang et~al.(2018)Liang, Liaw, Nishihara, Moritz, Fox, Goldberg, Gonzalez, Jordan, and Stoica]{liang2018rllib}
Eric Liang, Richard Liaw, Robert Nishihara, Philipp Moritz, Roy Fox, Ken Goldberg, Joseph~E. Gonzalez, Michael~I. Jordan, and Ion Stoica.
\newblock {RLlib}: Abstractions for distributed reinforcement learning.
\newblock In \emph{International Conference on Machine Learning ({ICML})}, 2018.

\bibitem[Liu et~al.(2021)Liu, Li, Wang, and Zheng]{erl}
Xiao-Yang Liu, Zechu Li, Zhaoran Wang, and Jiahao Zheng.
\newblock {ElegantRL}: Massively parallel framework for cloud-native deep reinforcement learning.
\newblock \url{https://github.com/AI4Finance-Foundation/ElegantRL}, 2021.

\bibitem[Loshchilov \& Hutter(2017)Loshchilov and Hutter]{adamw}
Ilya Loshchilov and Frank Hutter.
\newblock Decoupled weight decay regularization.
\newblock \emph{arXiv preprint arXiv:1711.05101}, 2017.

\bibitem[Mnih et~al.(2015)Mnih, Kavukcuoglu, Silver, Rusu, Veness, Bellemare, Graves, Riedmiller, Fidjeland, Ostrovski, Petersen, Beattie, Sadik, Antonoglou, King, Kumaran, Wierstra, Legg, and Hassabis]{mnihHumanlevelControlDeep2015}
Volodymyr Mnih, Koray Kavukcuoglu, David Silver, Andrei~A. Rusu, Joel Veness, Marc~G. Bellemare, Alex Graves, Martin Riedmiller, Andreas~K. Fidjeland, Georg Ostrovski, Stig Petersen, Charles Beattie, Amir Sadik, Ioannis Antonoglou, Helen King, Dharshan Kumaran, Daan Wierstra, Shane Legg, and Demis Hassabis.
\newblock Human-level control through deep reinforcement learning.
\newblock \emph{Nature}, 518\penalty0 (7540):\penalty0 529--533, February 2015.
\newblock ISSN 1476-4687.
\newblock \doi{10.1038/nature14236}.

\bibitem[Nair et~al.(2020)Nair, Dalal, Gupta, and Levine]{nair2020accelerating}
Ashvin Nair, Murtaza Dalal, Abhishek Gupta, and Sergey Levine.
\newblock Accelerating online reinforcement learning with offline datasets.
\newblock \emph{arXiv preprint arXiv:2006.09359}, 2020.

\bibitem[Nakamoto et~al.(2023)Nakamoto, Zhai, Singh, Mark, Ma, Finn, Kumar, and Levine]{nakamoto2023cal}
Mitsuhiko Nakamoto, Yuexiang Zhai, Anikait Singh, Max~Sobol Mark, Yi~Ma, Chelsea Finn, Aviral Kumar, and Sergey Levine.
\newblock Cal-ql: Calibrated offline rl pre-training for efficient online fine-tuning.
\newblock \emph{arXiv preprint arXiv:2303.05479}, 2023.

\bibitem[Smith et~al.(2022)Smith, Kostrikov, and Levine]{smithWalkParkLearning2022}
Laura Smith, Ilya Kostrikov, and Sergey Levine.
\newblock A {{Walk}} in the {{Park}}: {{Learning}} to {{Walk}} in 20 {{Minutes With Model-Free Reinforcement Learning}}, August 2022.

\bibitem[Stooke \& Abbeel(2019)Stooke and Abbeel]{1909.01500}
Adam Stooke and Pieter Abbeel.
\newblock rlpyt: A research code base for deep reinforcement learning in pytorch, 2019.
\newblock URL \url{https://arxiv.org/abs/1909.01500}.

\bibitem[Takuma~Seno(2021)]{seno2021d3rlpy}
Michita~Imai Takuma~Seno.
\newblock d3rlpy: An offline deep reinforcement library.
\newblock In \emph{NeurIPS 2021 Offline Reinforcement Learning Workshop}, December 2021.

\bibitem[Tarasov et~al.(2023)Tarasov, Kurenkov, Nikulin, and Kolesnikov]{tarasov2023revisiting}
Denis Tarasov, Vladislav Kurenkov, Alexander Nikulin, and Sergey Kolesnikov.
\newblock Revisiting the minimalist approach to offline reinforcement learning.
\newblock \emph{arXiv preprint arXiv:2305.09836}, 2023.

\bibitem[Weng et~al.(2021)Weng, Chen, Yan, You, Duburcq, Zhang, Su, Su, and Zhu]{tianshou}
Jiayi Weng, Huayu Chen, Dong Yan, Kaichao You, Alexis Duburcq, Minghao Zhang, Yi~Su, Hang Su, and Jun Zhu.
\newblock Tianshou: A highly modularized deep reinforcement learning library.
\newblock \emph{arXiv preprint arXiv:2107.14171}, 2021.

\bibitem[Weng et~al.(2022)Weng, Lin, Huang, Liu, Makoviichuk, Makoviychuk, Liu, Song, Luo, Jiang, Xu, and Yan]{weng2022envpool}
Jiayi Weng, Min Lin, Shengyi Huang, Bo~Liu, Denys Makoviichuk, Viktor Makoviychuk, Zichen Liu, Yufan Song, Ting Luo, Yukun Jiang, Zhongwen Xu, and Shuicheng Yan.
\newblock Env{P}ool: A highly parallel reinforcement learning environment execution engine.
\newblock In S.~Koyejo, S.~Mohamed, A.~Agarwal, D.~Belgrave, K.~Cho, and A.~Oh (eds.), \emph{Advances in Neural Information Processing Systems}, volume~35, pp.\  22409--22421. Curran Associates, Inc., 2022.
\newblock URL \url{https://proceedings.neurips.cc/paper_files/paper/2022/file/8caaf08e49ddbad6694fae067442ee21-Paper-Datasets_and_Benchmarks.pdf}.

\bibitem[Wu et~al.(2022)Wu, Wu, Qiu, Wang, and Long]{wu2022supported}
Jialong Wu, Haixu Wu, Zihan Qiu, Jianmin Wang, and Mingsheng Long.
\newblock Supported policy optimization for offline reinforcement learning.
\newblock \emph{arXiv preprint arXiv:2202.06239}, 2022.

\end{thebibliography}
\bibliographystyle{iclr2023_conference}
\clearpage
%%%%%%%%%%%%%%%%%%%%%%%%%%%%%%%%%%%%%%%%%%%%%%%%%%%%%%%%%%%%
%%%%%%%%%%%%%%%%%%%%%%%%%%%%%%%%%%%%%%%%%%%%%%%%%%%%%%%%%%%%
\section*{Checklist}

\begin{enumerate}

\item For all authors...
\begin{enumerate}
  \item Do the main claims made in the abstract and introduction accurately reflect the paper's contributions and scope?
    \answerYes{}
  \item Did you describe the limitations of your work?
    \answerYes{See \autoref{method}}
  \item Did you discuss any potential negative societal impacts of your work?
    \answerNA{}
  \item Have you read the ethics review guidelines and ensured that your paper conforms to them?
    \answerYes{}
\end{enumerate}

\item If you are including theoretical results...
\begin{enumerate}
  \item Did you state the full set of assumptions of all theoretical results?
    \answerNA{}
	\item Did you include complete proofs of all theoretical results?
    \answerNA{}
\end{enumerate}

\item If you ran experiments (e.g. for benchmarks)...
\begin{enumerate}
  \item Did you include the code, data, and instructions needed to reproduce the main experimental results (either in the supplemental material or as a URL)?
    \answerYes{We release our codebase, configs, and in-depth reports at \url{https://github.com/corl-team/CORL}}
  \item Did you specify all the training details (e.g., data splits, hyperparameters, how they were chosen)?
    \answerYes{See \autoref{appendix:hyperparameters}}
	\item Did you report error bars (e.g., with respect to the random seed after running experiments multiple times)?
    \answerYes{}
	\item Did you include the total amount of compute and the type of resources used (e.g., type of GPUs, internal cluster, or cloud provider)?
    \answerYes{See \autoref{appendix:hyperparameters}}
\end{enumerate}

\item If you are using existing assets (e.g., code, data, models) or curating/releasing new assets...
\begin{enumerate}
  \item If your work uses existing assets, did you cite the creators?
    \answerYes{}
  \item Did you mention the license of the assets?
    \answerYes{See \autoref{appendix:license}}
  \item Did you include any new assets either in the supplemental material or as a URL?
    \answerNA{}
  \item Did you discuss whether and how consent was obtained from people whose data you're using/curating?
    \answerNA{}
  \item Did you discuss whether the data you are using/curating contains personally identifiable information or offensive content?
    \answerNA{}
\end{enumerate}

\item If you used crowdsourcing or conducted research with human subjects...
\begin{enumerate}
  \item Did you include the full text of instructions given to participants and screenshots, if applicable?
    \answerNA{}
  \item Did you describe any potential participant risks, with links to Institutional Review Board (IRB) approvals, if applicable?
    \answerNA{}
  \item Did you include the estimated hourly wage paid to participants and the total amount spent on participant compensation?
    \answerNA{}
\end{enumerate}

\end{enumerate}

%%%%%%%%%%%%%%%%%%%%%%%%%%%%%%%%%%%%%%%%%%%%%%%%%%%%%%%%%%%%

\clearpage

\appendix

\section{Additional Benchmark Information}
        \label{appendix:results}
\subsection{Offline}
\begin{figure}[ht]
\centering
\captionsetup{justification=centering}
     \centering
     \begin{subfigure}[b]{0.49\textwidth}
         \centering
         \includegraphics[width=\textwidth]{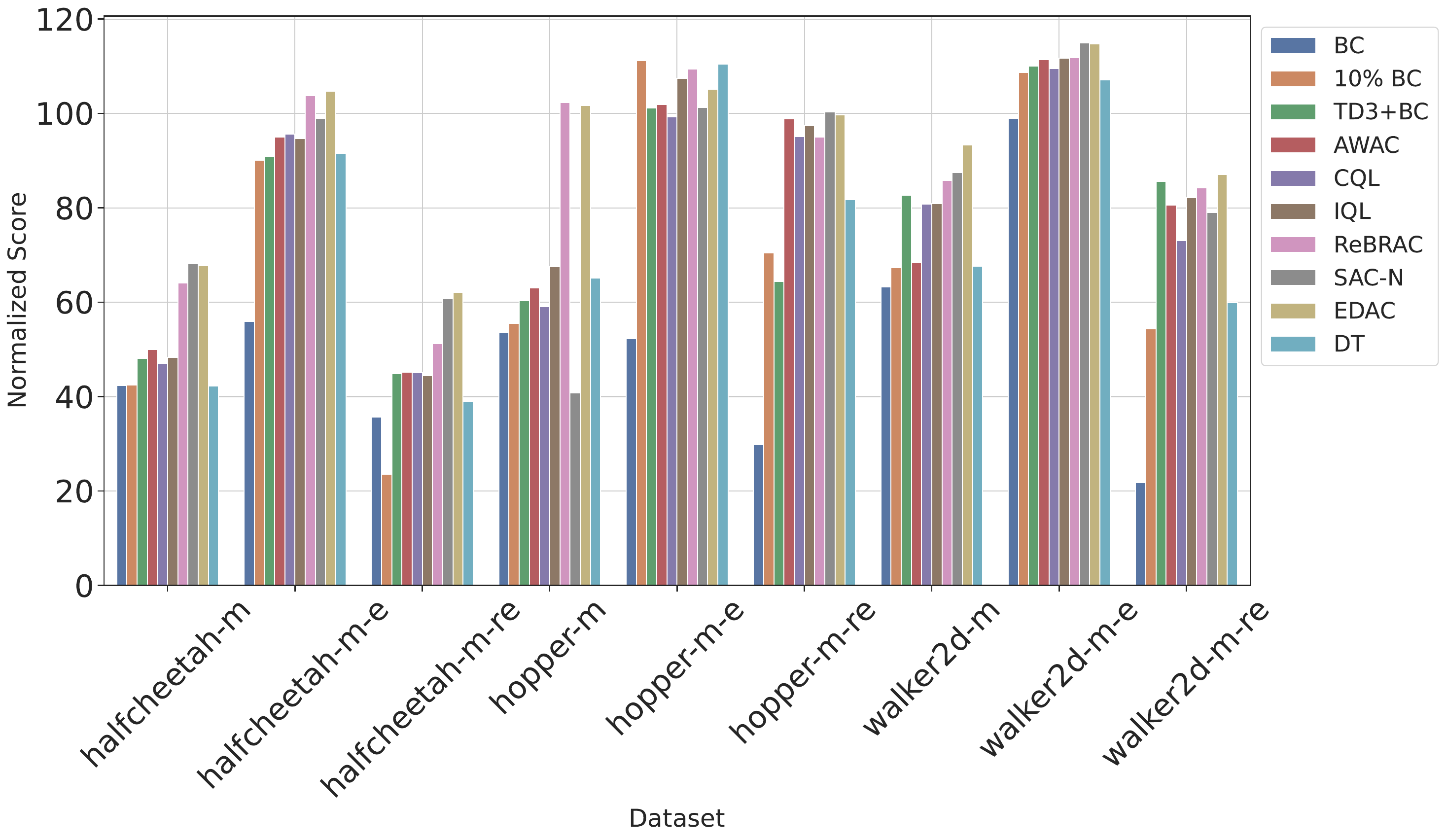}
         \caption{}
         \label{fig:loco-last-bars}
     \end{subfigure}
     \hfill
     \begin{subfigure}[b]{0.49\textwidth}
         \centering
         \includegraphics[width=\textwidth]{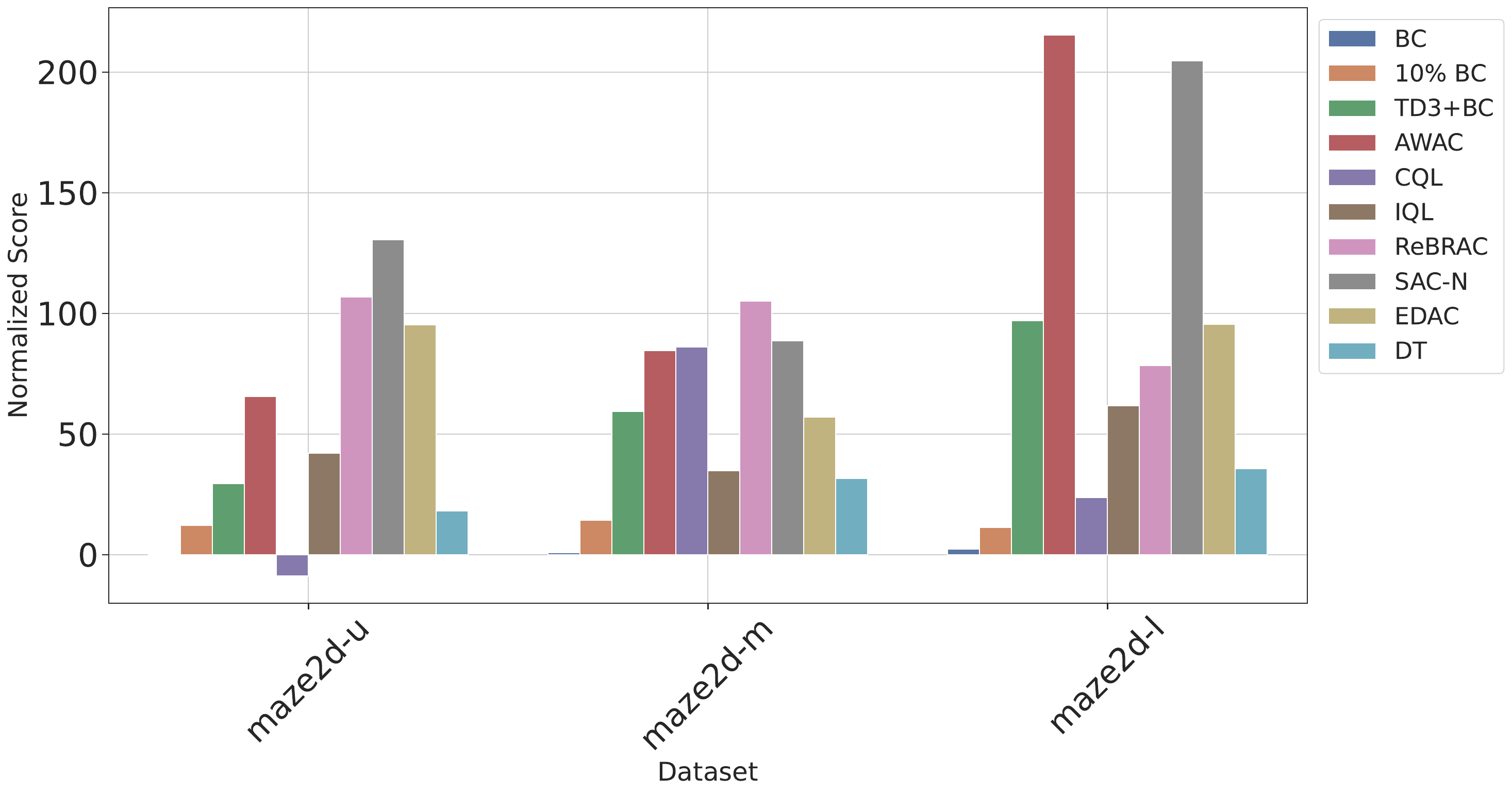}
         \caption{}
         \label{fig:maze-last-bars}
     \end{subfigure}
     \begin{subfigure}[b]{0.49\textwidth}
         \centering
         \includegraphics[width=\textwidth]{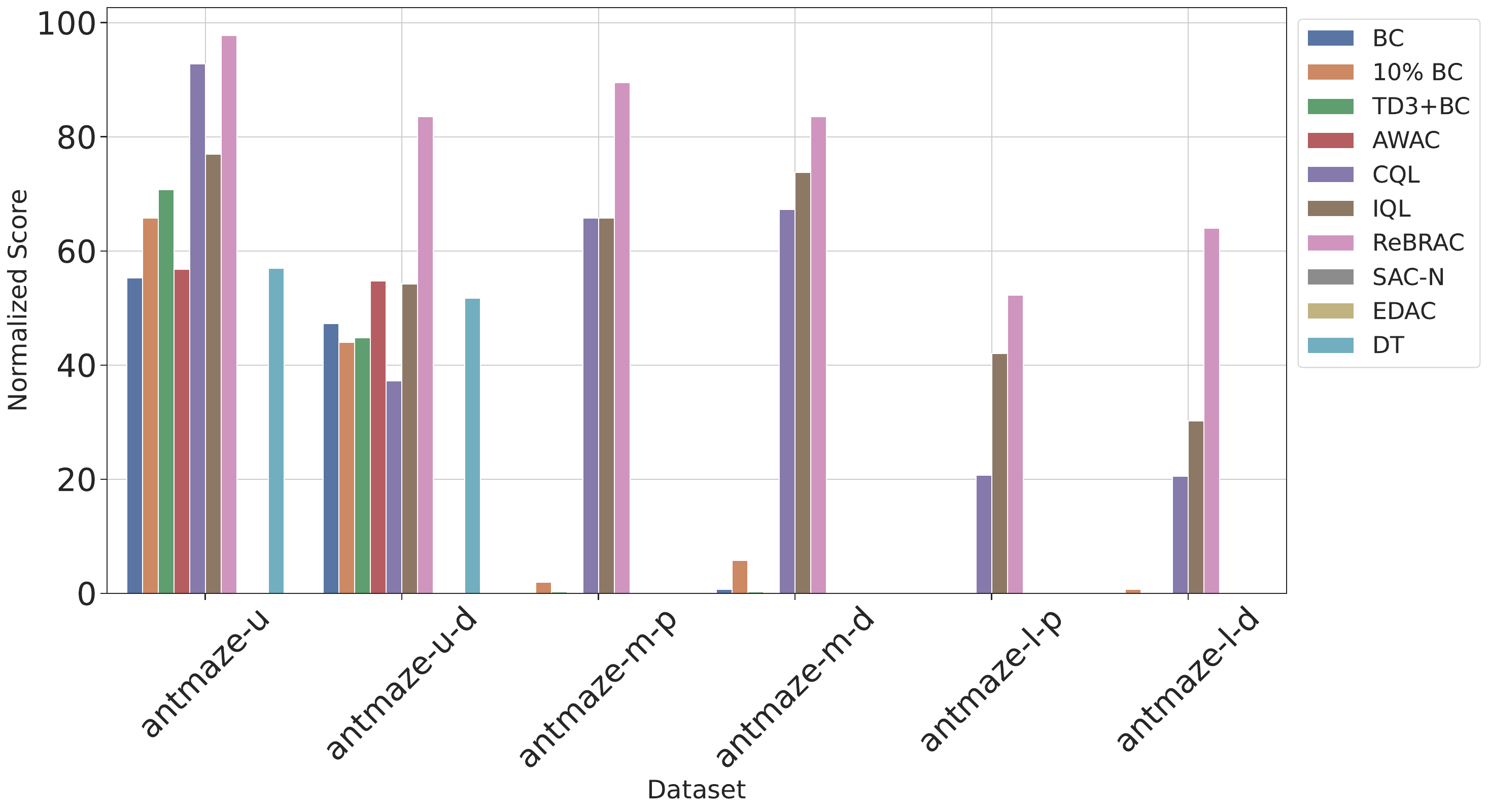}
         \caption{}
         \label{fig:ant-last-bars}
     \end{subfigure}
     \begin{subfigure}[b]{0.49\textwidth}
         \centering
         \includegraphics[width=\textwidth]{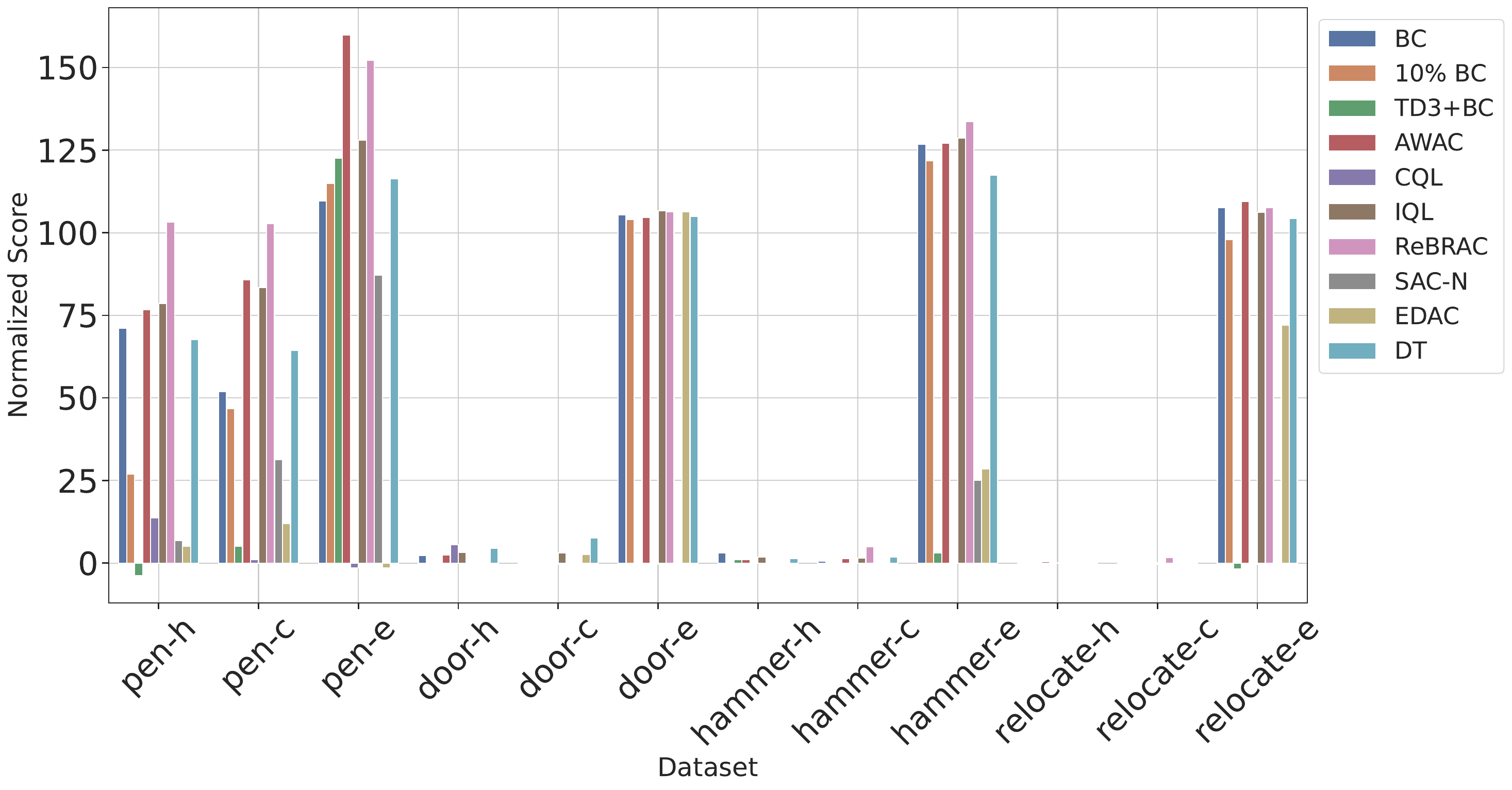}
         \caption{}
         \label{fig:adroit-last-bars}
     \end{subfigure}
        \caption{Graphical representation of the normalized performance of the last trained policy on D4RL averaged over 4 random seeds.
        % normalized average over last scores on D4RL Gym tasks, averaged over 4 random seeds.
        (a) Gym-MuJoCo datasets. (b) Maze2d datasets (c) AntMaze datasets (d) Adroit datasets}
        \label{fig:last_bars}
\end{figure}        

\begin{figure}[ht]
\centering
\captionsetup{justification=centering}
     \centering
     \begin{subfigure}[b]{0.49\textwidth}
         \centering
         \includegraphics[width=\textwidth]{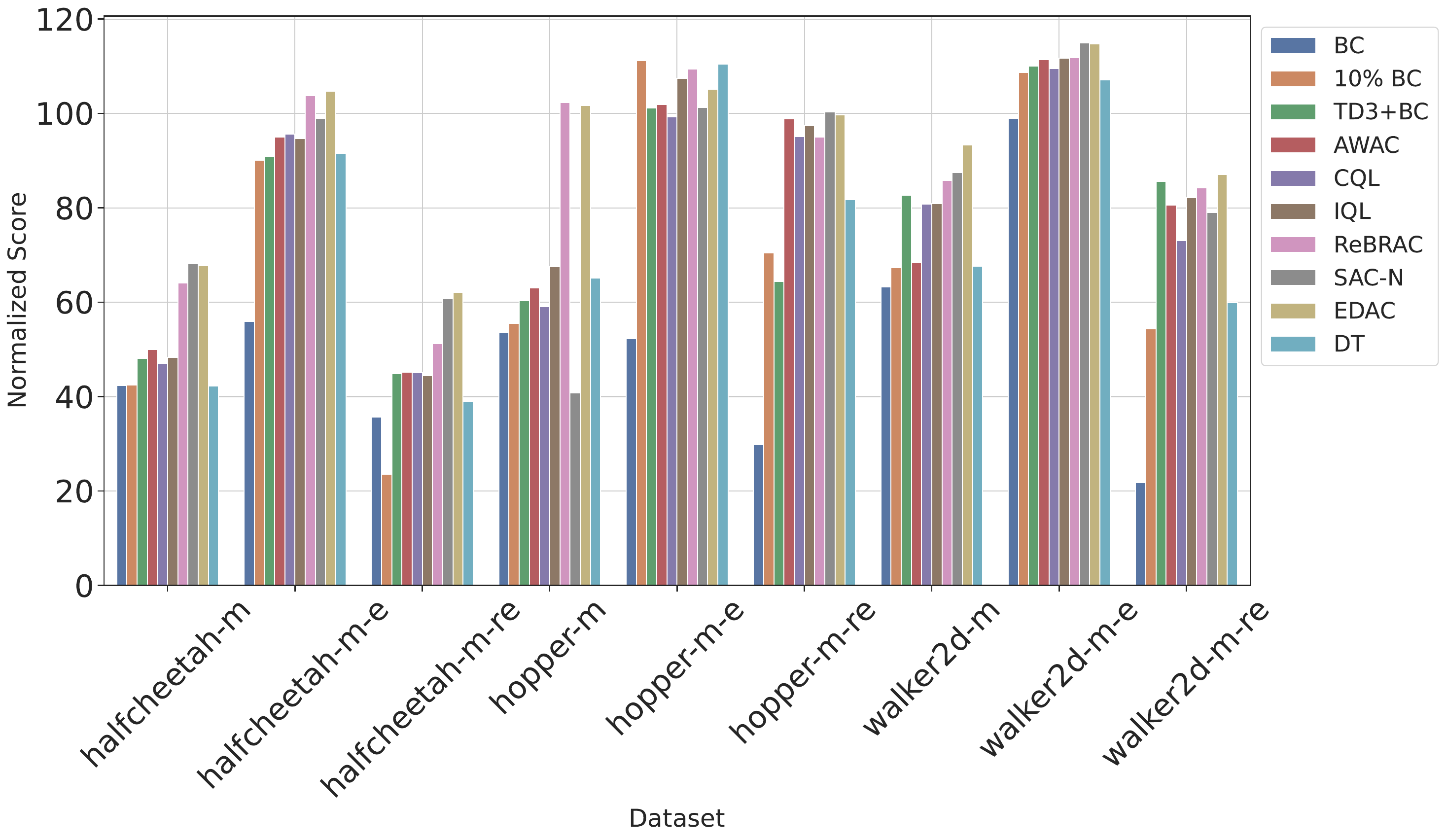}
         \caption{}
         \label{fig:loco-max-bars}
     \end{subfigure}
     \hfill
     \begin{subfigure}[b]{0.49\textwidth}
         \centering
         \includegraphics[width=\textwidth]{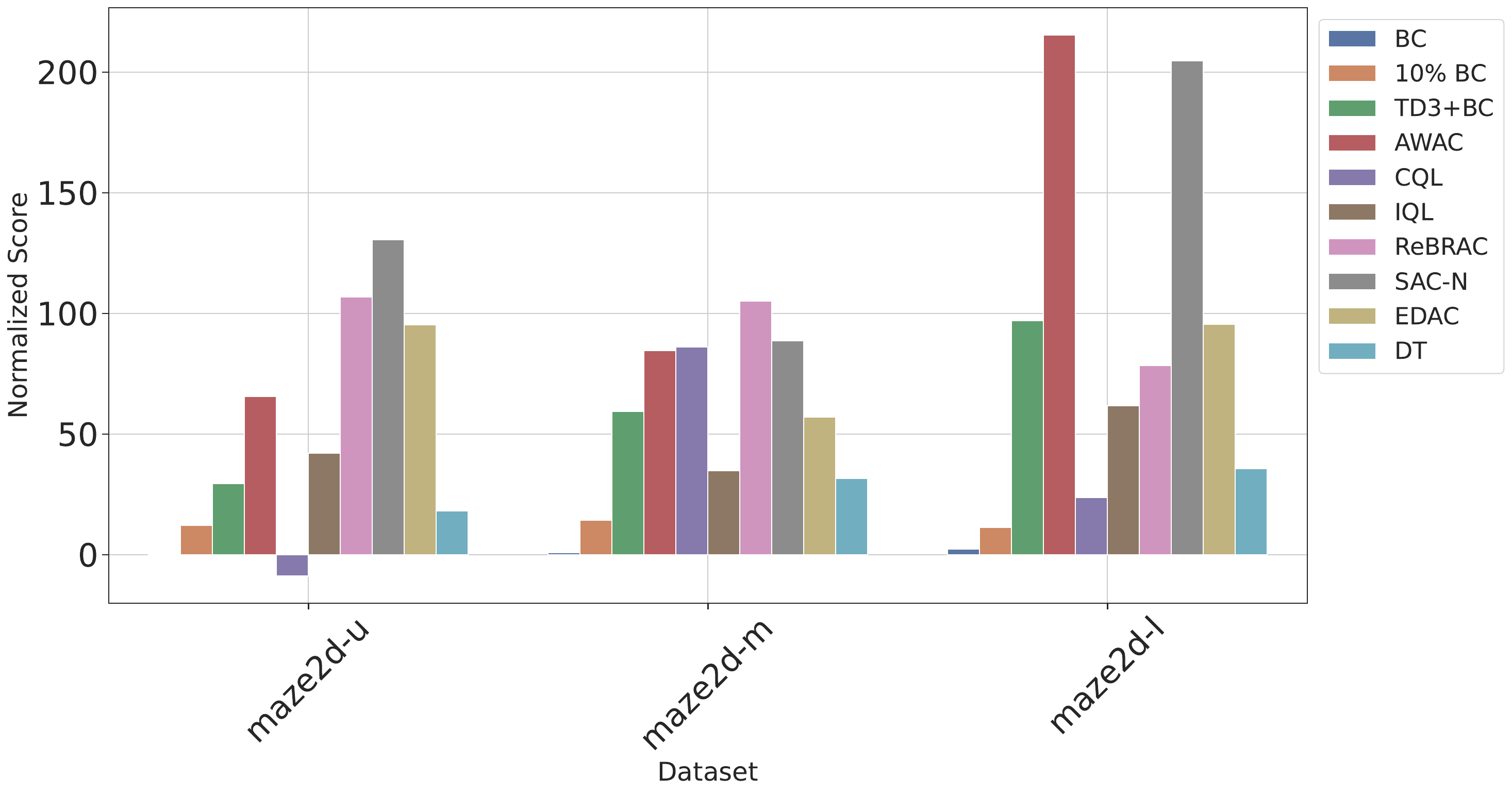}
         \caption{}
         \label{fig:maze-max-bars}
     \end{subfigure}
      \begin{subfigure}[b]{0.49\textwidth}
         \centering
         \includegraphics[width=\textwidth]{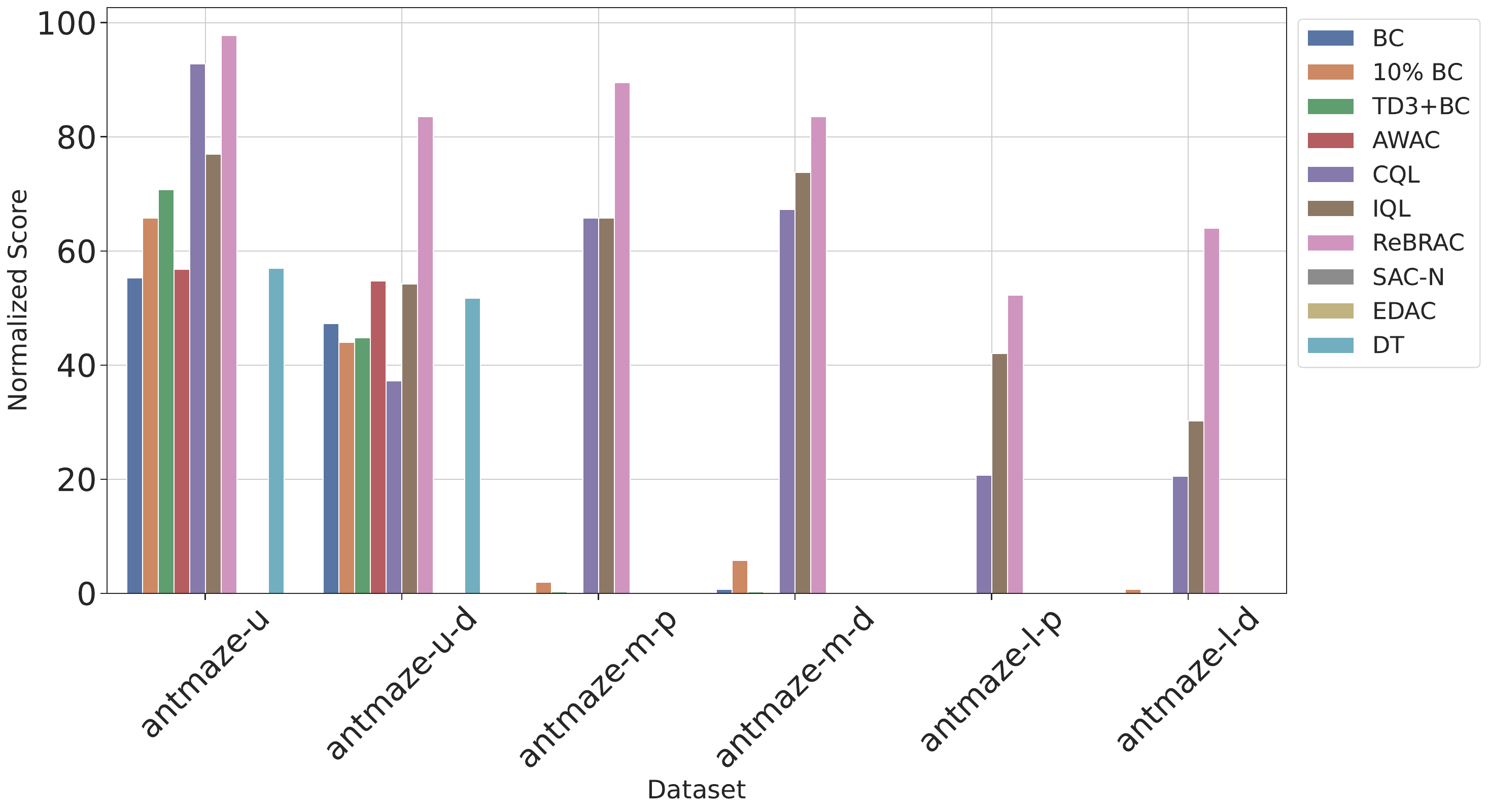}
         \caption{}
         \label{fig:ant-max-bars}
     \end{subfigure}
     \begin{subfigure}[b]{0.49\textwidth}
         \centering
         \includegraphics[width=\textwidth]{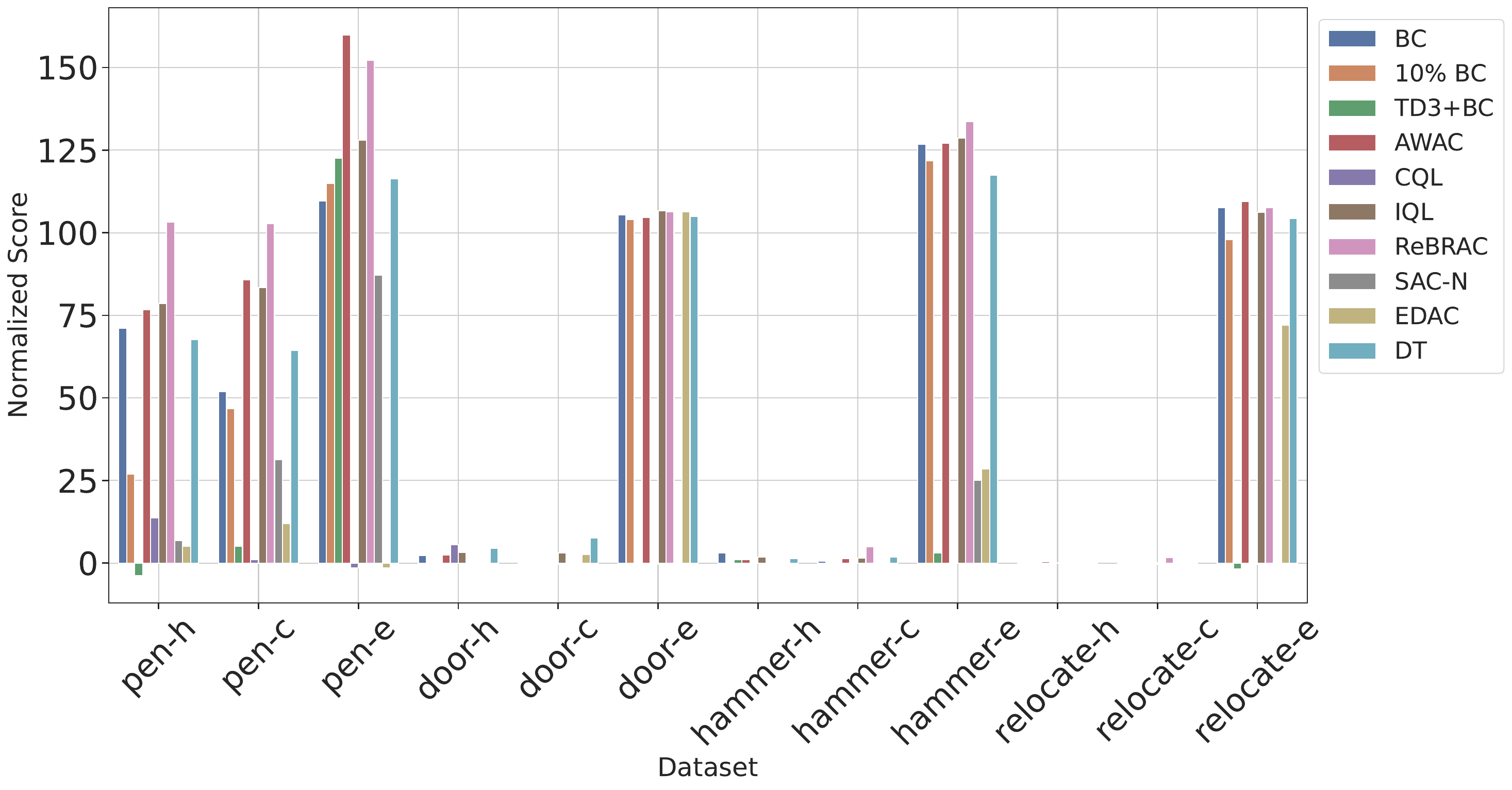}
         \caption{}
         \label{fig:adroit-max-bars}
     \end{subfigure}
        \caption{Graphical representation of the normalized performance of the best trained policy on D4RL averaged over 4 random seeds.
        % normalized average over maximal scores on D4RL Gym tasks, averaged over 4 random seeds. 
        (a) Gym-MuJoCo datasets. (b) Maze2d datasets (c) AntMaze datasets (d) Adroit datasets}
        \label{fig:max_bars}
\end{figure}

\begin{figure}[ht]
\centering
\captionsetup{justification=centering}
     \centering
     \begin{subfigure}[b]{0.32\textwidth}
         \centering
         \includegraphics[width=\textwidth]{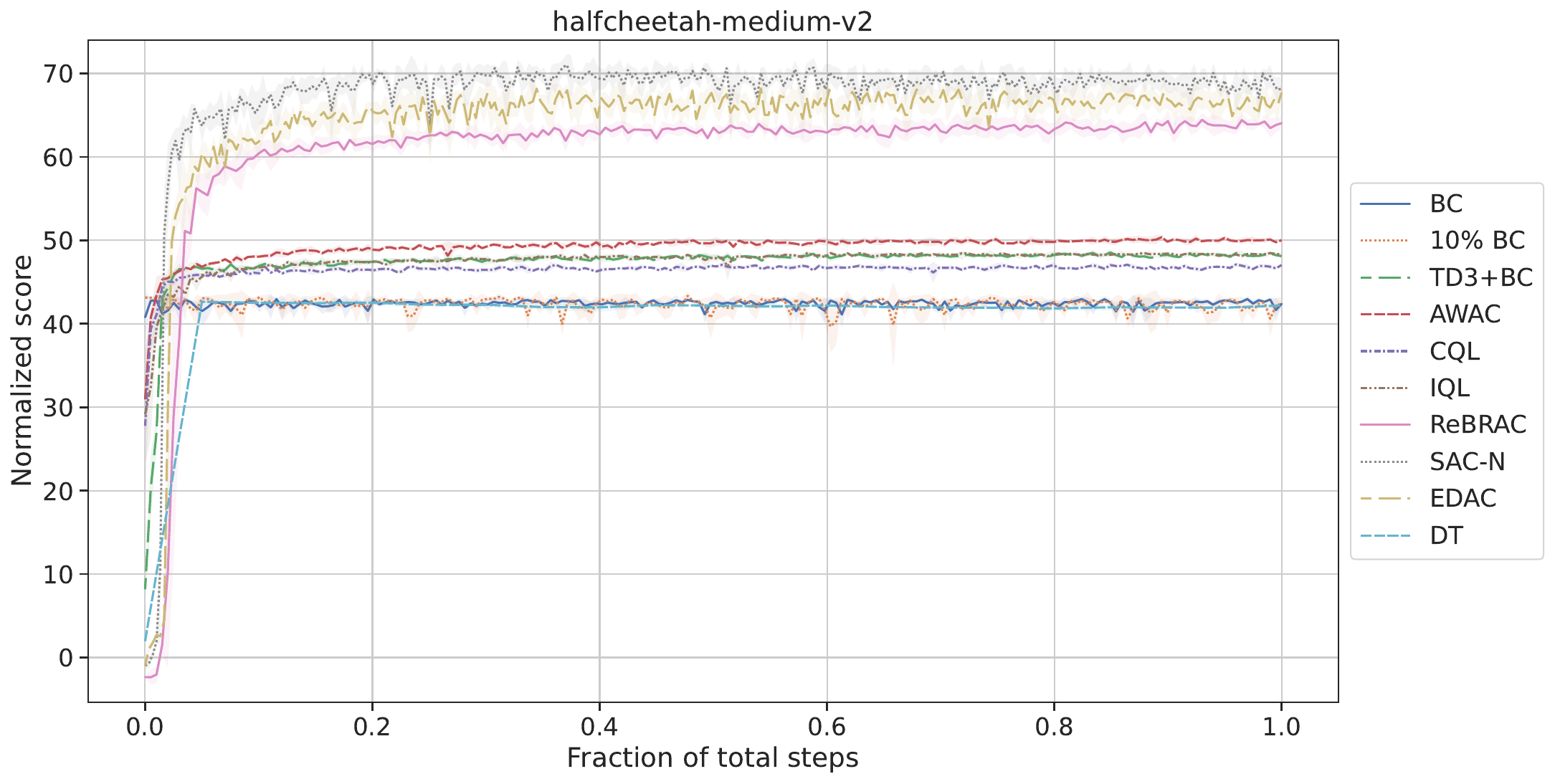}
         \caption{}
         \label{fig:halfcheetah-m}
     \end{subfigure}
     \hfill
     \begin{subfigure}[b]{0.32\textwidth}
         \centering
         \includegraphics[width=\textwidth]{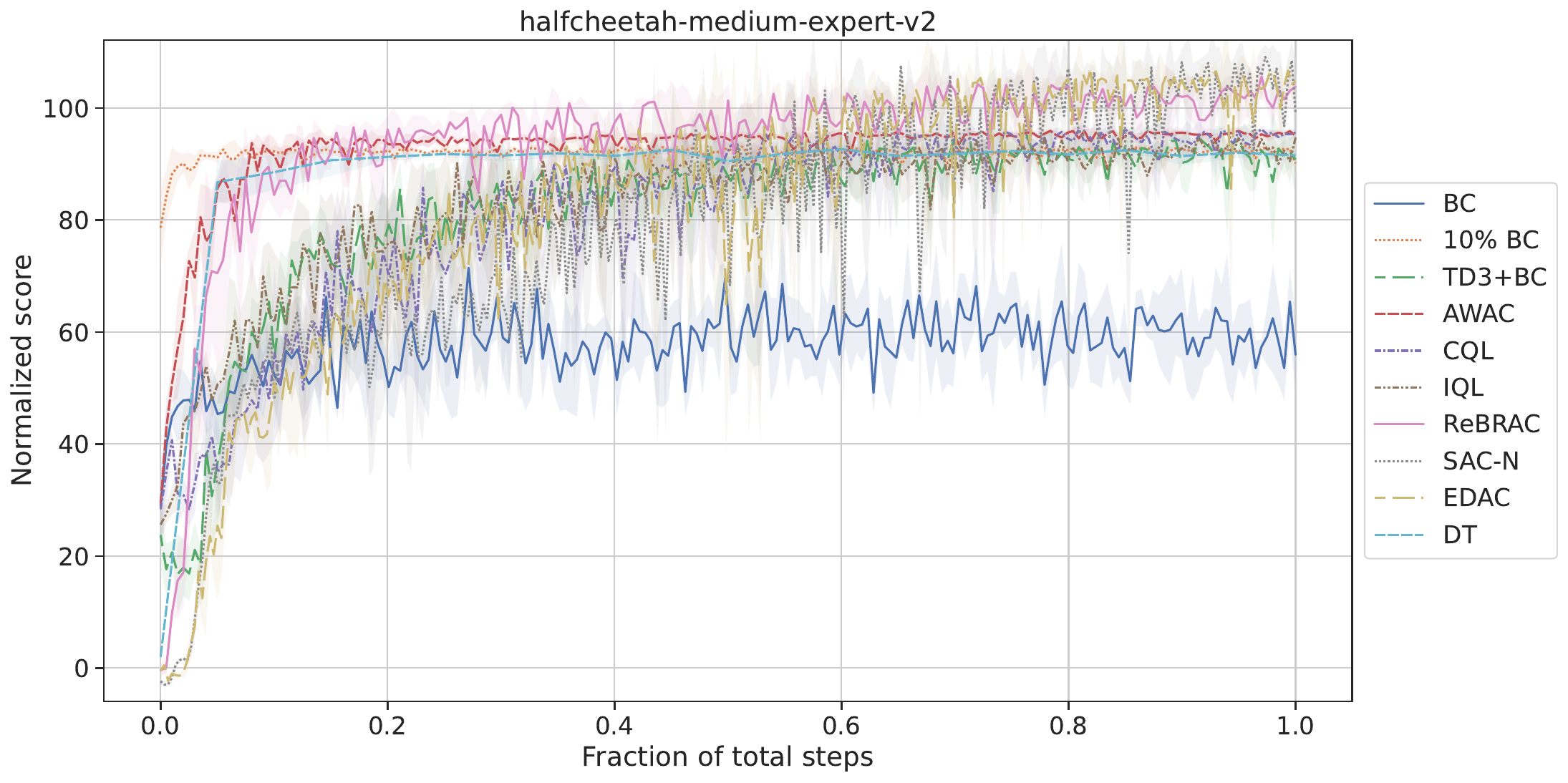}
         \caption{}
         \label{fig:halfcheetah-m-e}
     \end{subfigure}
     \begin{subfigure}[b]{0.32\textwidth}
         \centering
         \includegraphics[width=\textwidth]{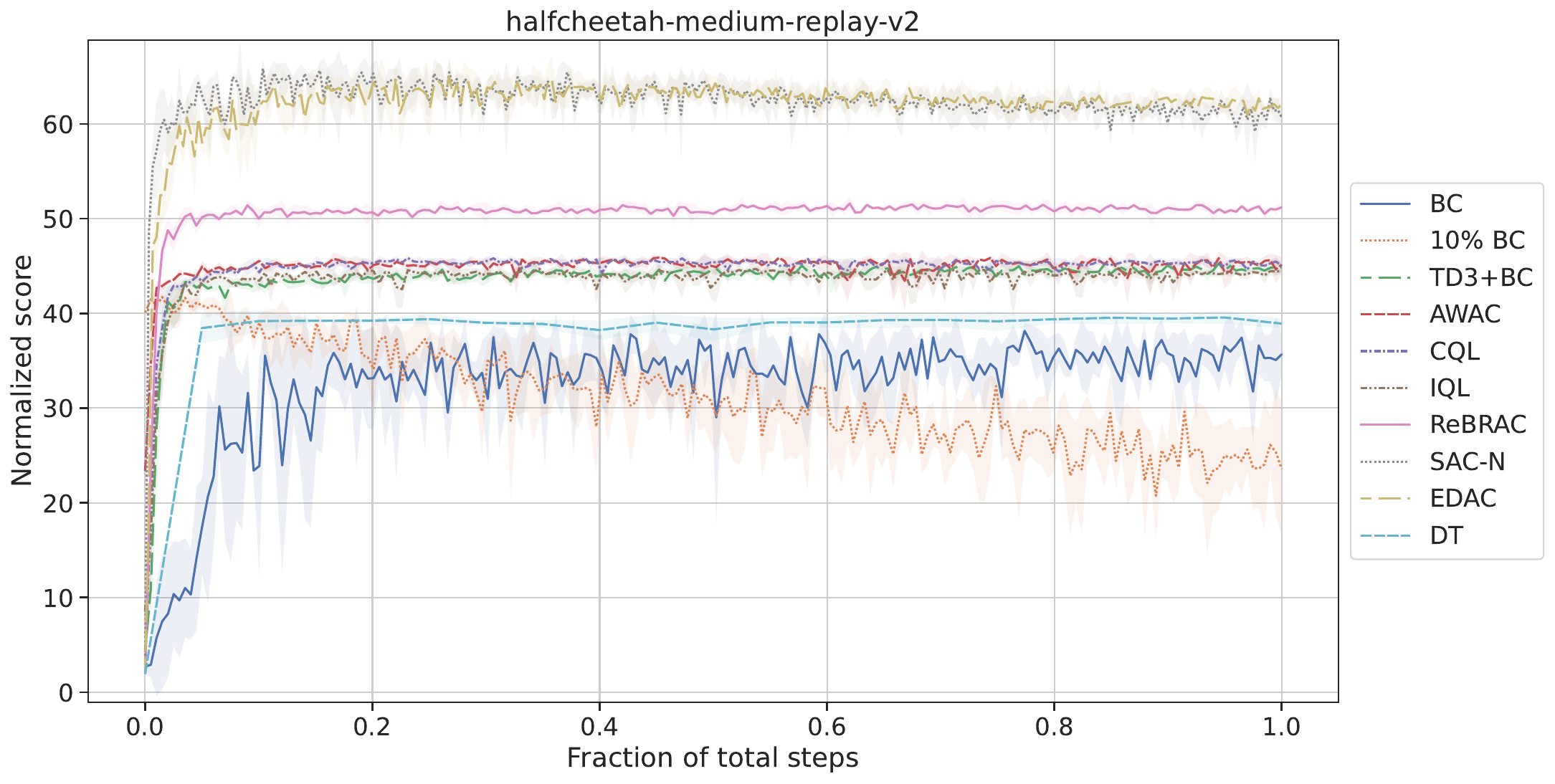}
         \caption{}
         \label{fig:halfcheetah-m-re}
     \end{subfigure}
    \caption{Training curves for HalfCheetah task. \\ (a) Medium dataset, (b) Medium-expert dataset, (c) Medium-replay dataset}
        \label{fig:halfcheetah_curves}
\end{figure}

\begin{figure}[!ht]
\centering
\captionsetup{justification=centering}
     \centering
     \begin{subfigure}[b]{0.32\textwidth}
         \centering
         \includegraphics[width=\textwidth]{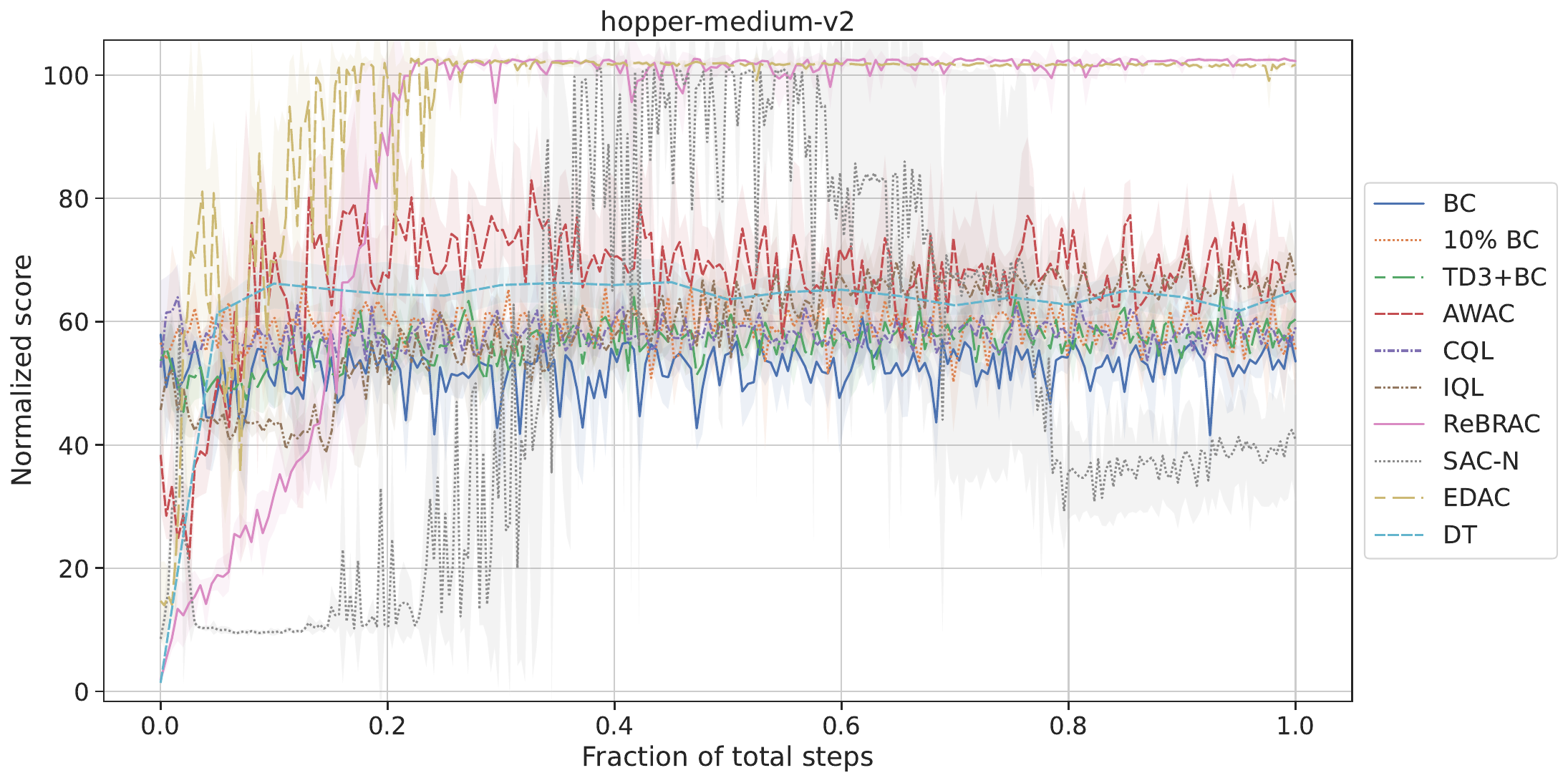}
         \caption{}
         \label{fig:hopper-m}
     \end{subfigure}
     \hfill
     \begin{subfigure}[b]{0.32\textwidth}
         \centering
         \includegraphics[width=\textwidth]{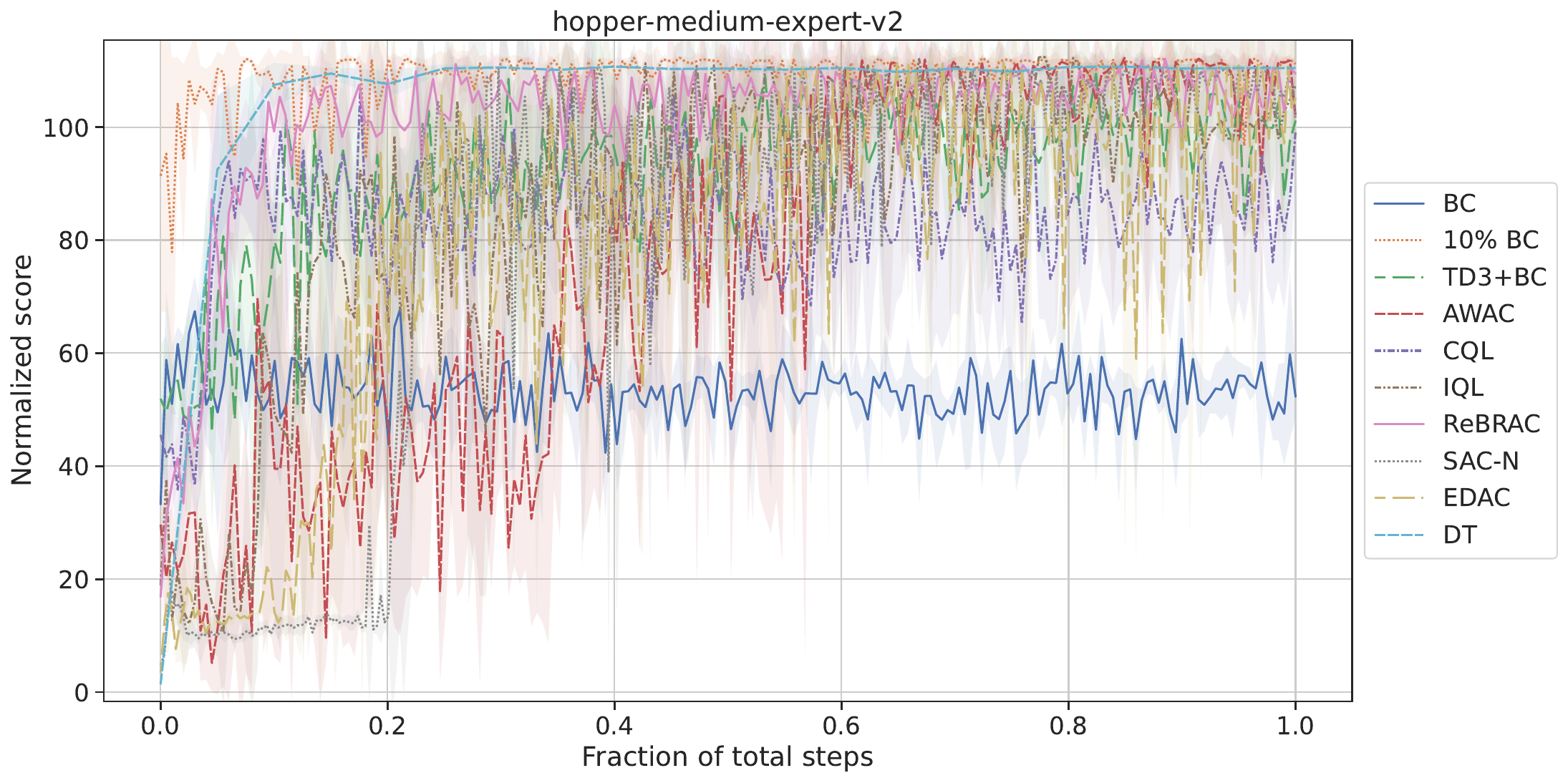}
         \caption{}
         \label{fig:hopper-m-e}
     \end{subfigure}
     \begin{subfigure}[b]{0.32\textwidth}
         \centering
         \includegraphics[width=\textwidth]{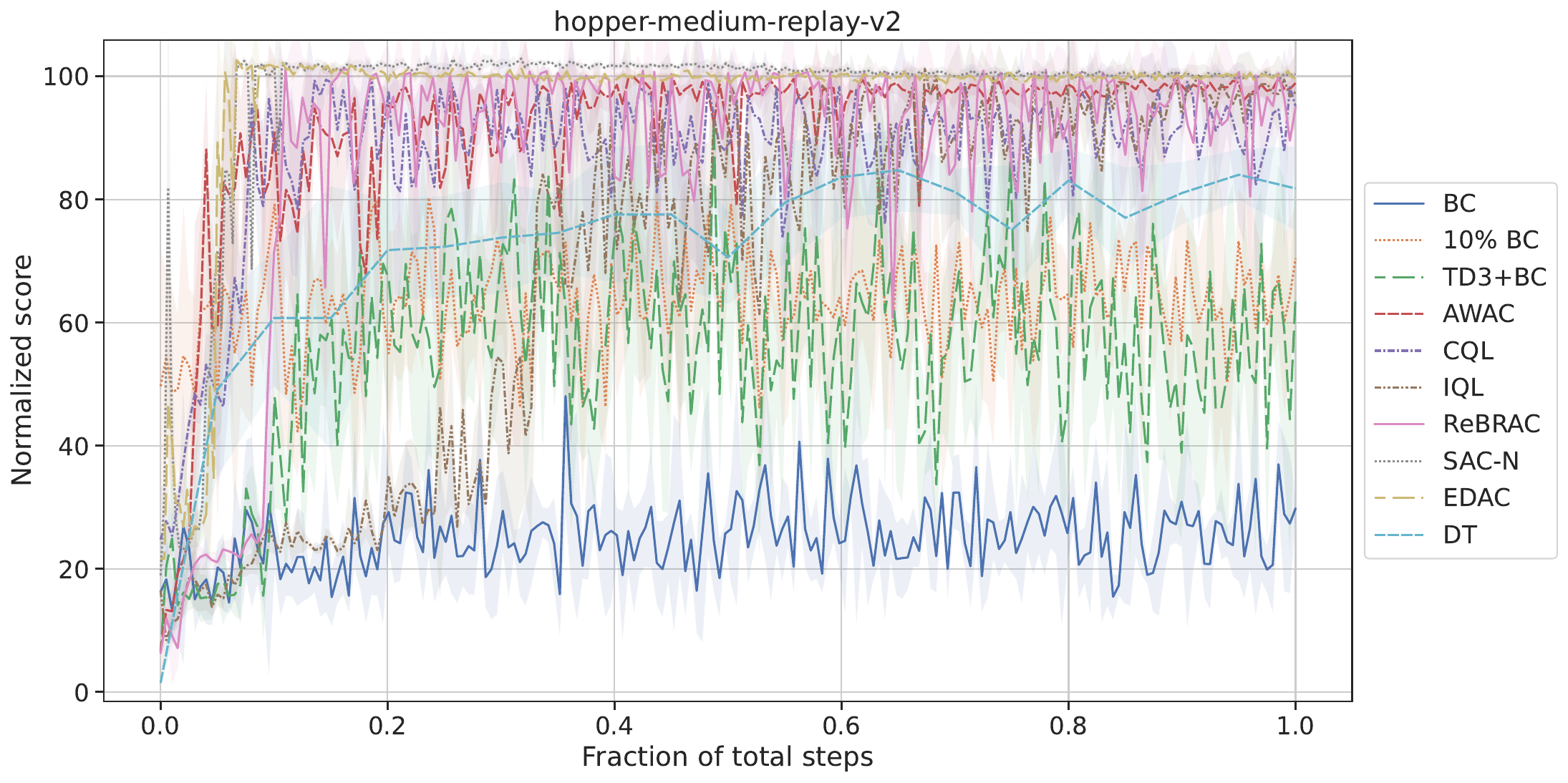}
         \caption{}
         \label{fig:hopper-m-re}
     \end{subfigure}
    \caption{Training curves for Hopper task.\\ (a) Medium dataset, (b) Medium-expert dataset, (c) Medium-replay dataset}
        \label{fig:hopper_curves}
\end{figure}

\begin{figure}[!ht]
\centering
\captionsetup{justification=centering}
     \centering
     \begin{subfigure}[b]{0.32\textwidth}
         \centering
         \includegraphics[width=\textwidth]{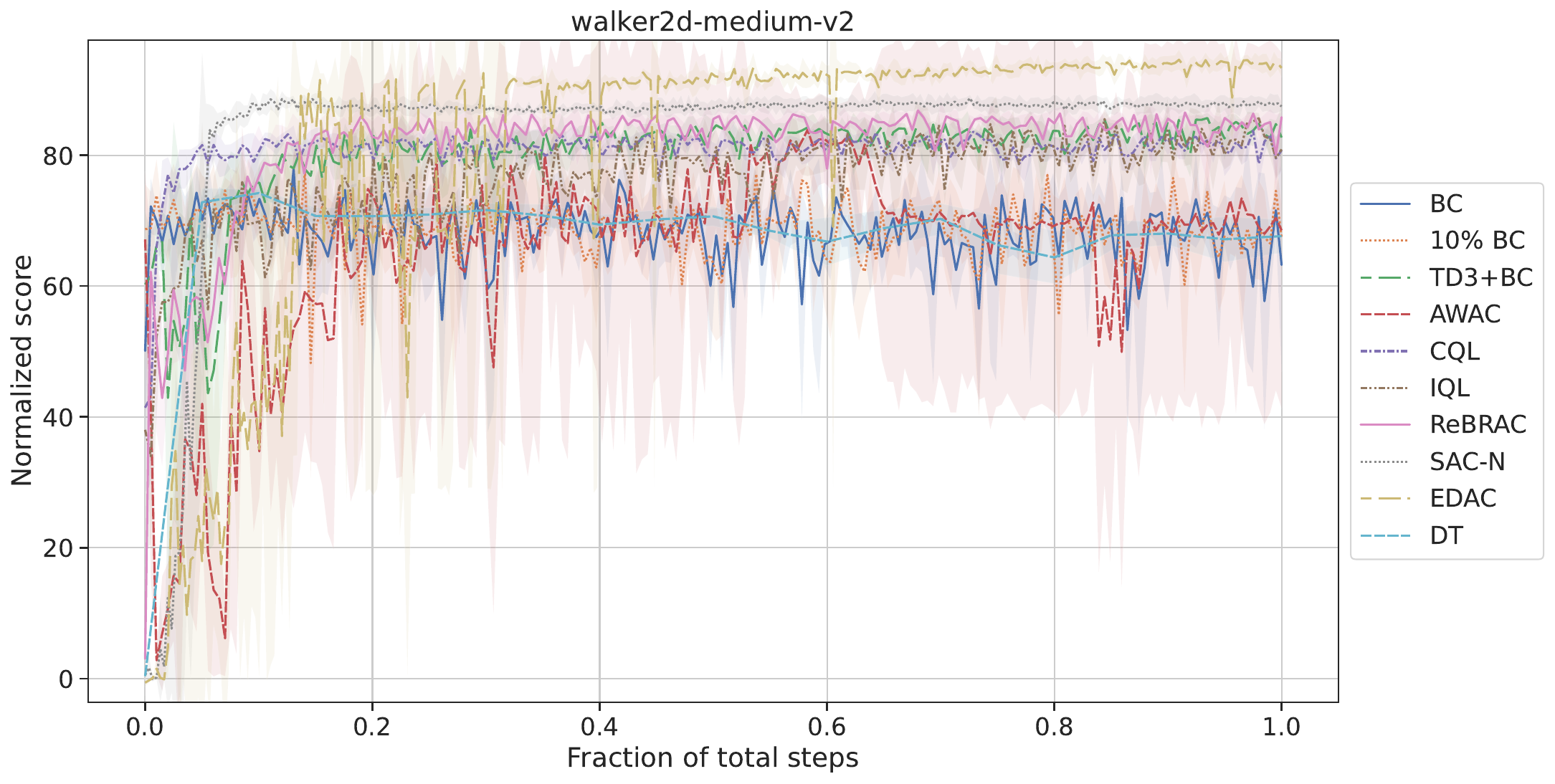}
         \caption{}
         \label{fig:walker2d-m}
     \end{subfigure}
     \hfill
     \begin{subfigure}[b]{0.32\textwidth}
         \centering
         \includegraphics[width=\textwidth]{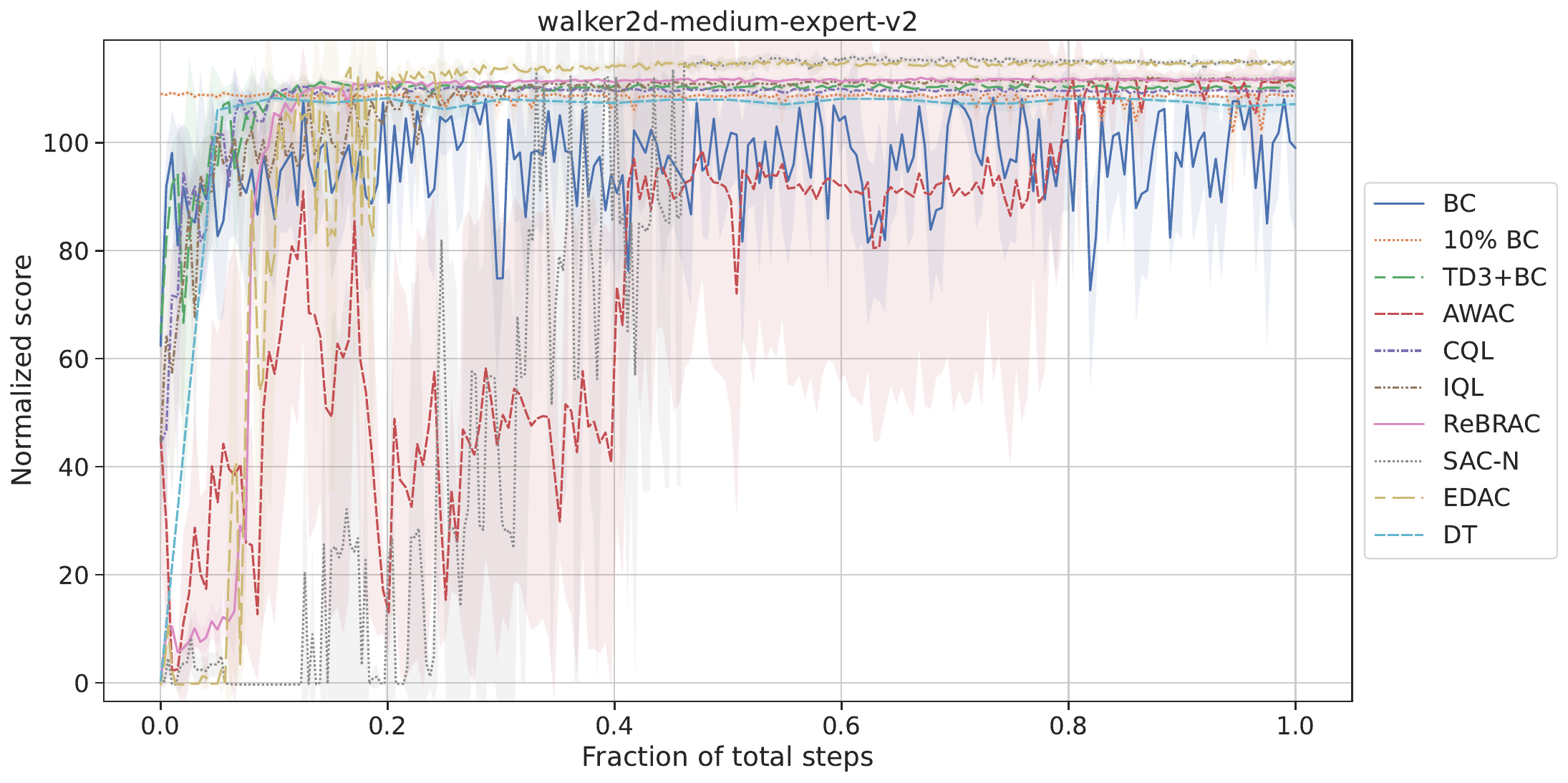}
         \caption{}
         \label{fig:walker2d-m-e}
     \end{subfigure}
     \begin{subfigure}[b]{0.32\textwidth}
         \centering
         \includegraphics[width=\textwidth]{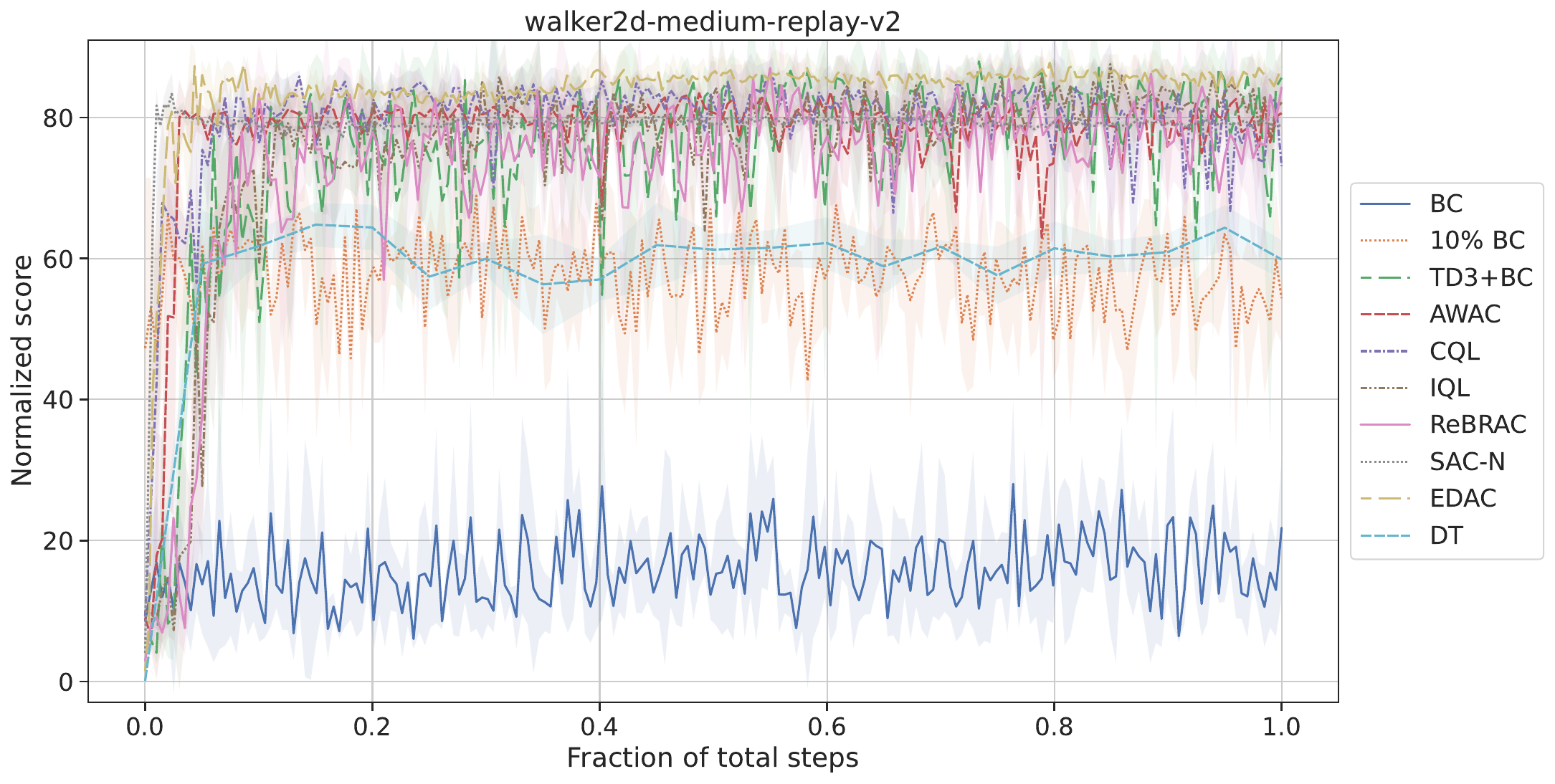}
         \caption{}
         \label{fig:walker2d-m-re}
     \end{subfigure}
    \caption{Training curves for Walker2d task.\\ (a) Medium dataset, (b) Medium-expert dataset, (c) Medium-replay dataset}
        \label{fig:walker2d_curves}
\end{figure}

\begin{figure}[!ht]
\centering
\captionsetup{justification=centering}
     \centering
     \begin{subfigure}[b]{0.32\textwidth}
         \centering
         \includegraphics[width=\textwidth]{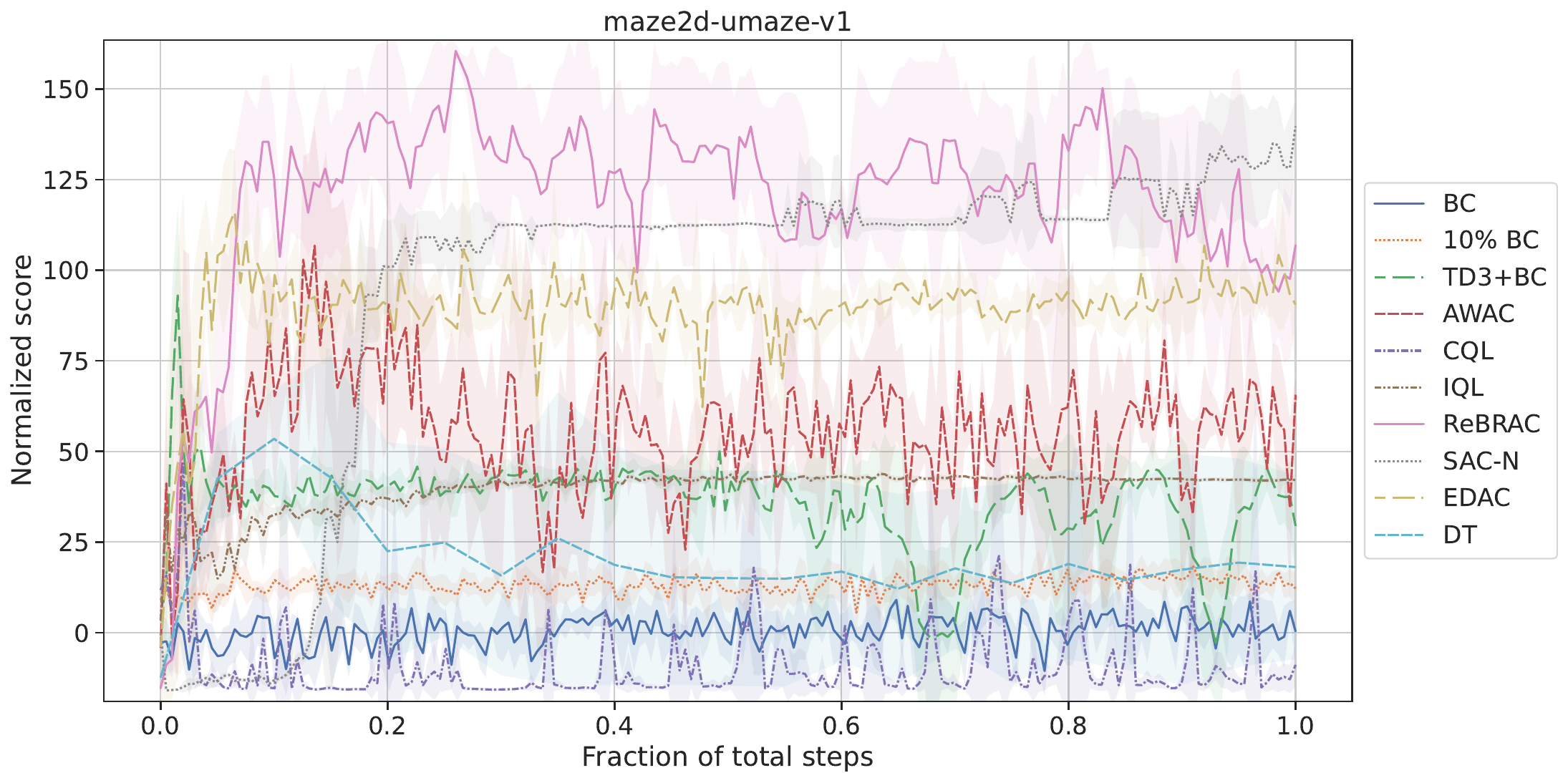}
         \caption{}
         \label{fig:maze-u}
     \end{subfigure}
     \hfill
     \begin{subfigure}[b]{0.32\textwidth}
         \centering
         \includegraphics[width=\textwidth]{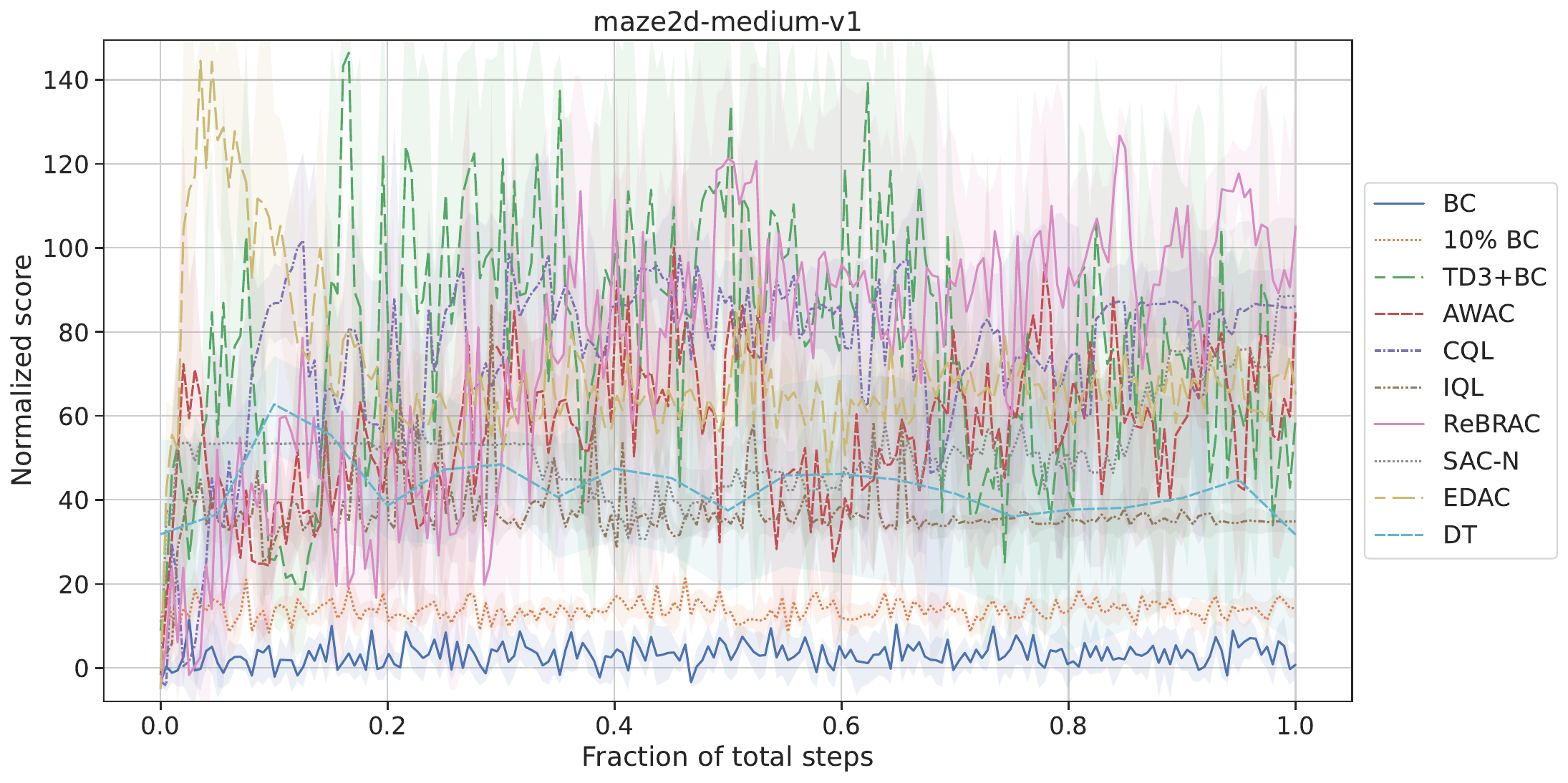}
         \caption{}
         \label{fig:maze-m}
     \end{subfigure}
     \begin{subfigure}[b]{0.32\textwidth}
         \centering
         \includegraphics[width=\textwidth]{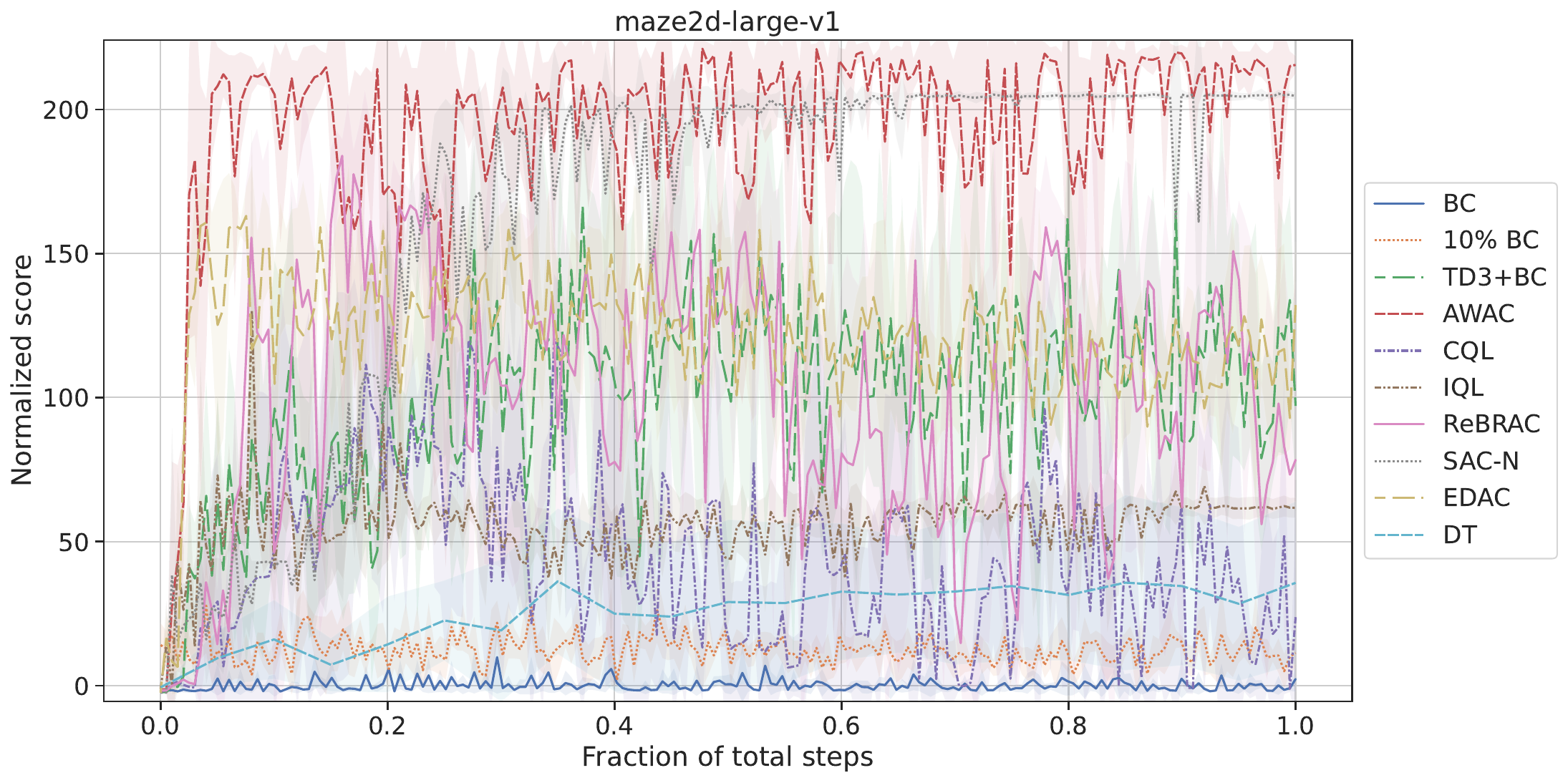}
         \caption{}
         \label{fig:maze-l}
     \end{subfigure}
    \caption{Training curves for Maze2d task.\\ (a) Medium dataset, (b) Medium-expert dataset, (c) Medium-replay dataset}
        \label{fig:maze_curves}
\end{figure}

\begin{figure}[ht]
\centering
\captionsetup{justification=centering}
     \centering
     \begin{subfigure}[b]{0.32\textwidth}
         \centering
         \includegraphics[width=\textwidth]{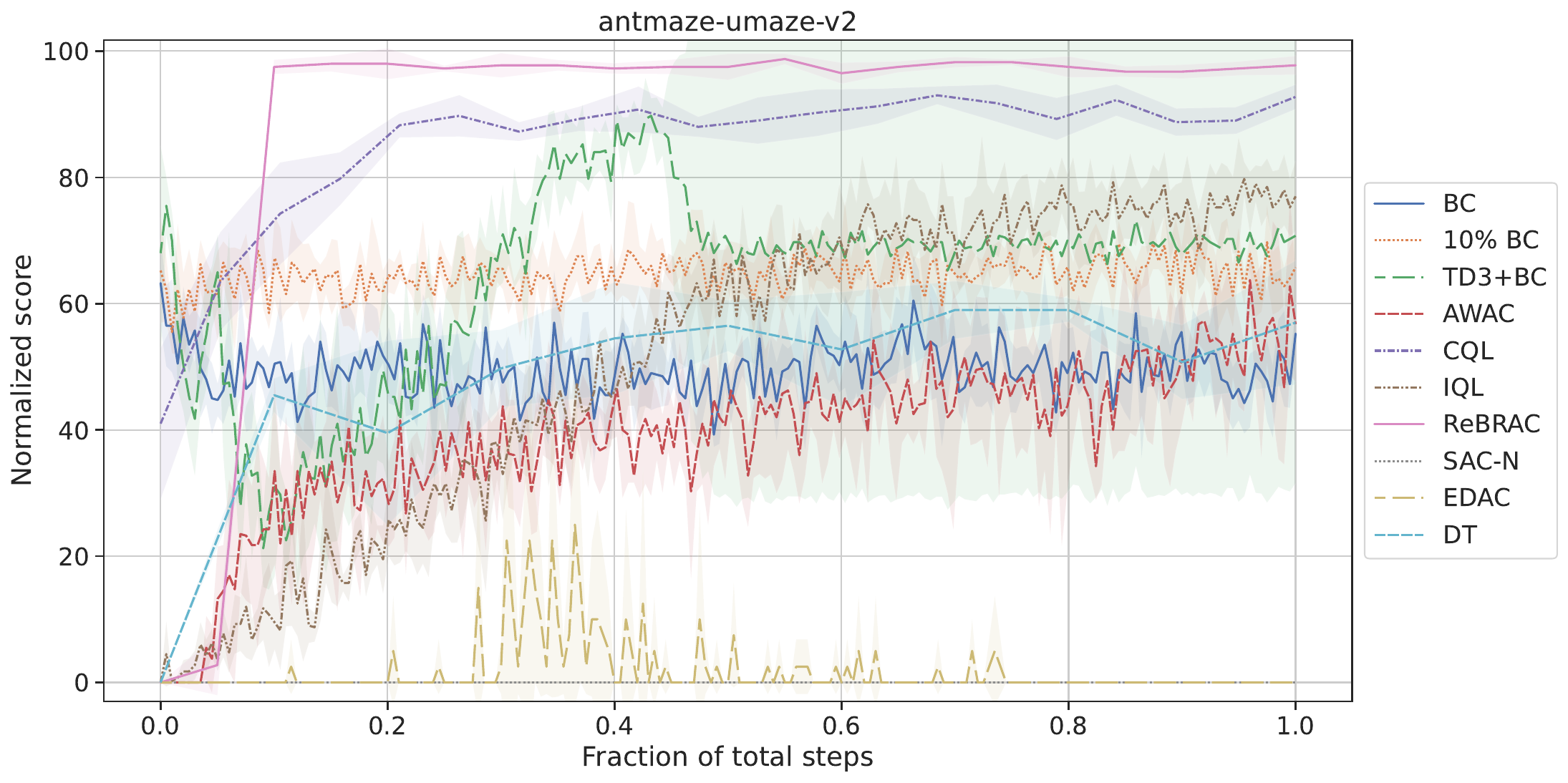}
         \caption{}
         \label{fig:ant-u}
     \end{subfigure}
     \hfill
     \begin{subfigure}[b]{0.32\textwidth}
         \centering
         \includegraphics[width=\textwidth]{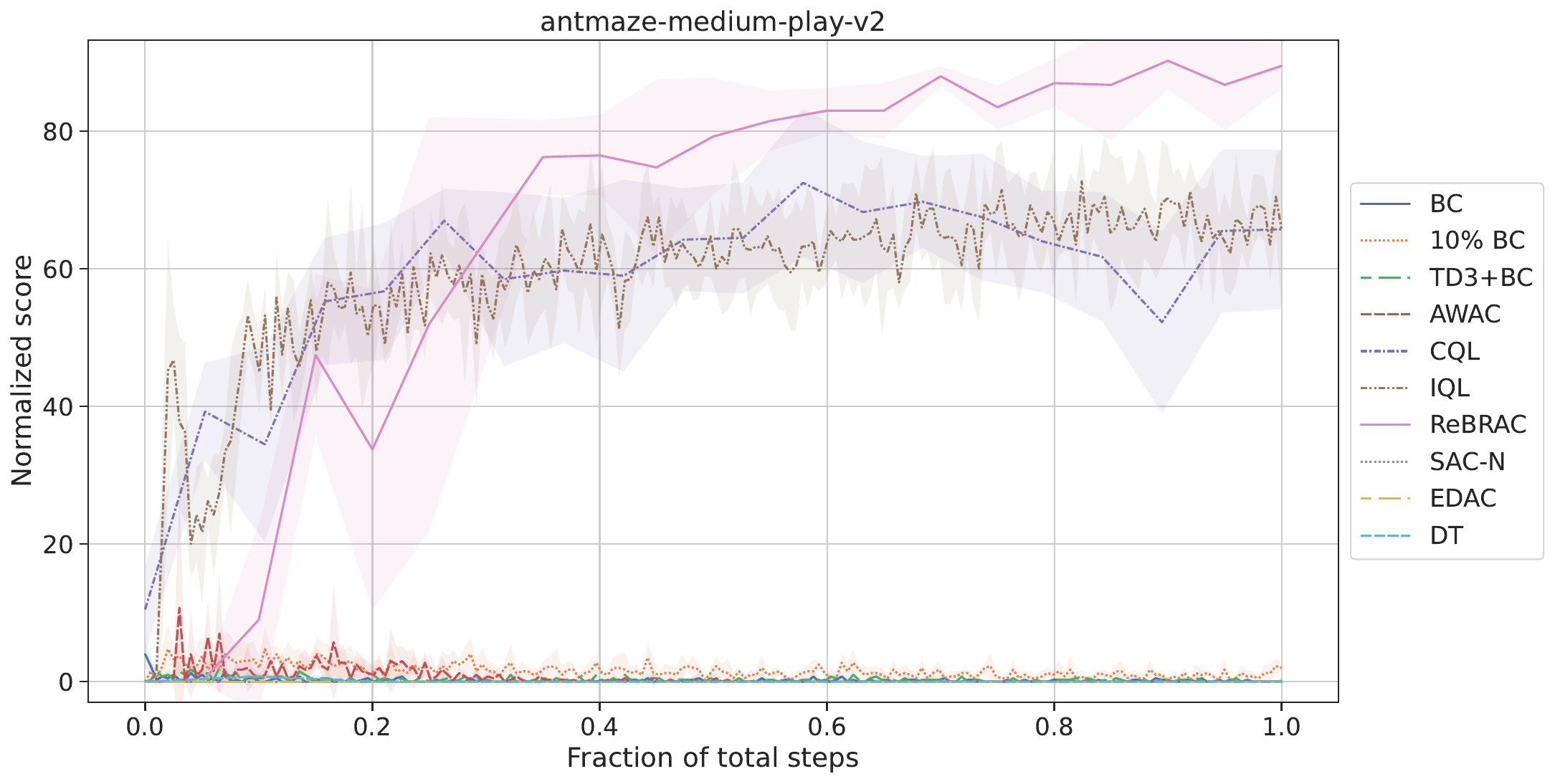}
         \caption{}
         \label{fig:ant-mp}
     \end{subfigure}
     \begin{subfigure}[b]{0.32\textwidth}
         \centering
         \includegraphics[width=\textwidth]{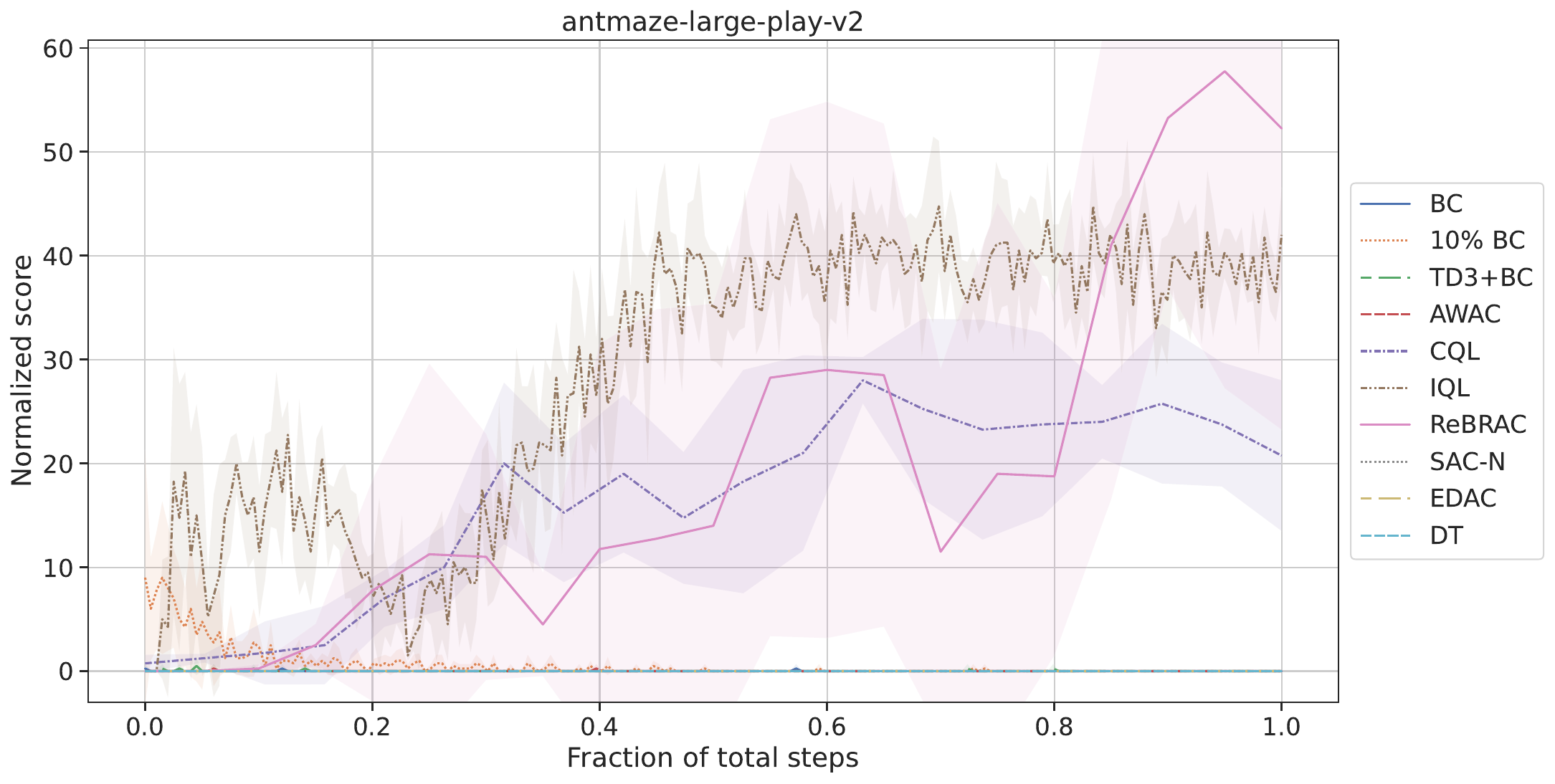}
         \caption{}
         \label{fig:ant-lp}
     \end{subfigure}
     \begin{subfigure}[b]{0.32\textwidth}
         \centering
         \includegraphics[width=\textwidth]{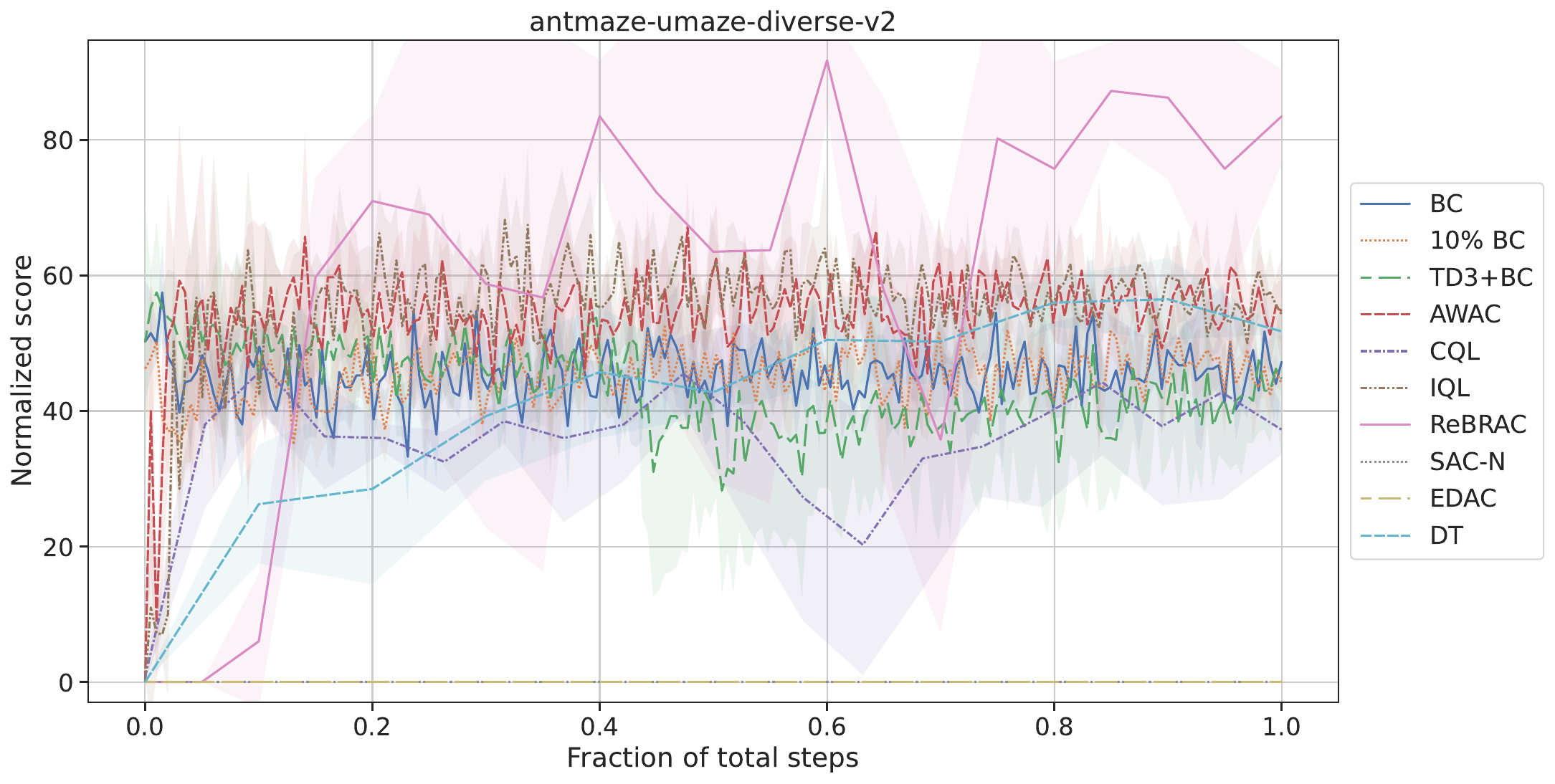}
         \caption{}
         \label{fig:ant-ud}
     \end{subfigure}
     \hfill
     \begin{subfigure}[b]{0.32\textwidth}
         \centering
         \includegraphics[width=\textwidth]{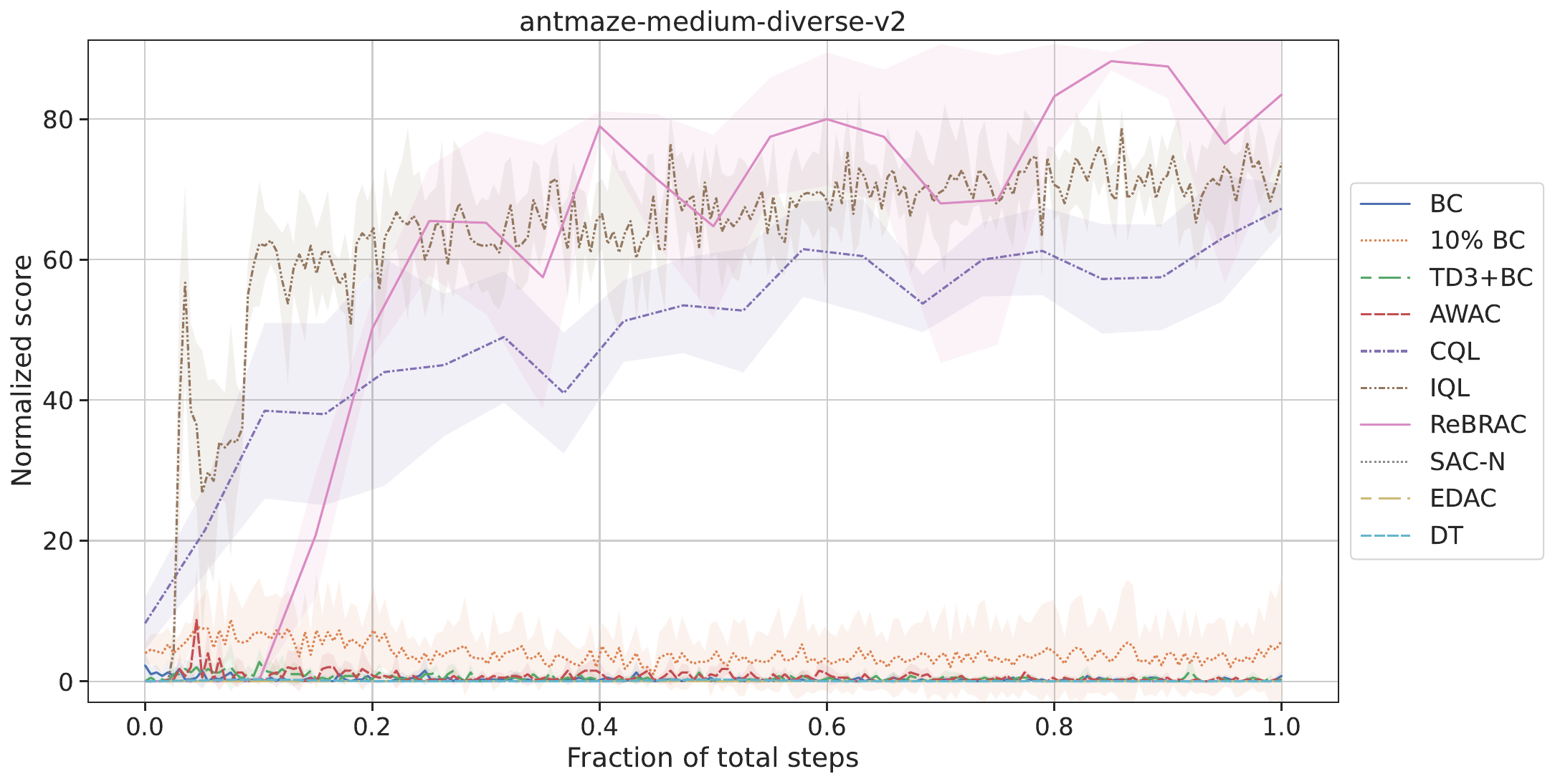}
         \caption{}
         \label{fig:ant-md}
     \end{subfigure}
     \begin{subfigure}[b]{0.32\textwidth}
         \centering
         \includegraphics[width=\textwidth]{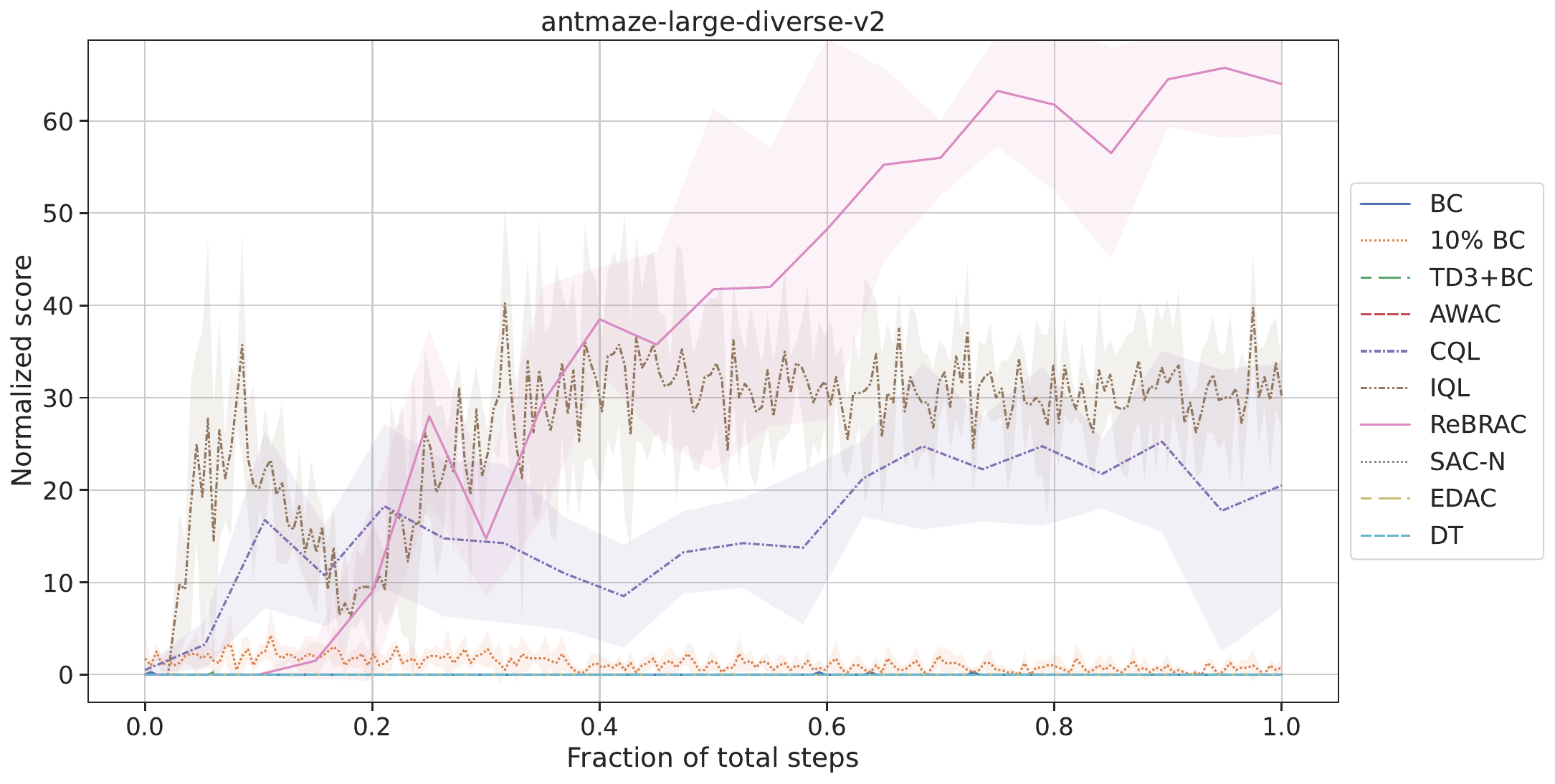}
         \caption{}
         \label{fig:ant-ld}
     \end{subfigure}
    \caption{Training curves for AntMaze task.\\ (a) Umaze dataset, (b) Medium-play dataset, (c) Large-play dataset, (d) Umaze-diverse dataset, (e) Medium-diverse dataset, (f) Large-diverse dataset}
        \label{fig:ant_curves}
\end{figure}

\begin{figure}[!ht]
\centering
\captionsetup{justification=centering}
     \centering
     \begin{subfigure}[b]{0.32\textwidth}
         \centering
         \includegraphics[width=\textwidth]{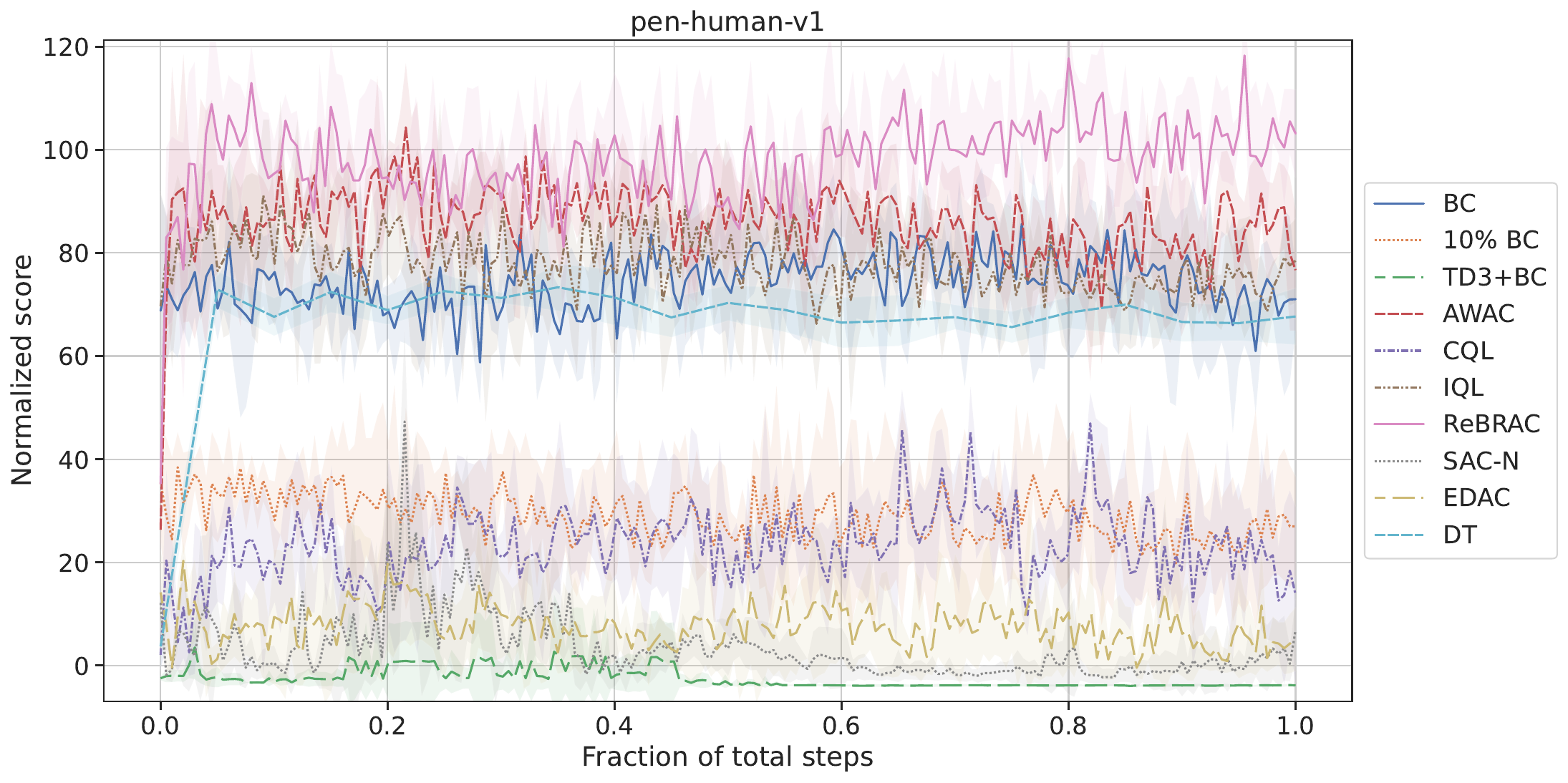}
         \caption{}
         \label{fig:pen-h}
     \end{subfigure}
     \hfill
     \begin{subfigure}[b]{0.32\textwidth}
         \centering
         \includegraphics[width=\textwidth]{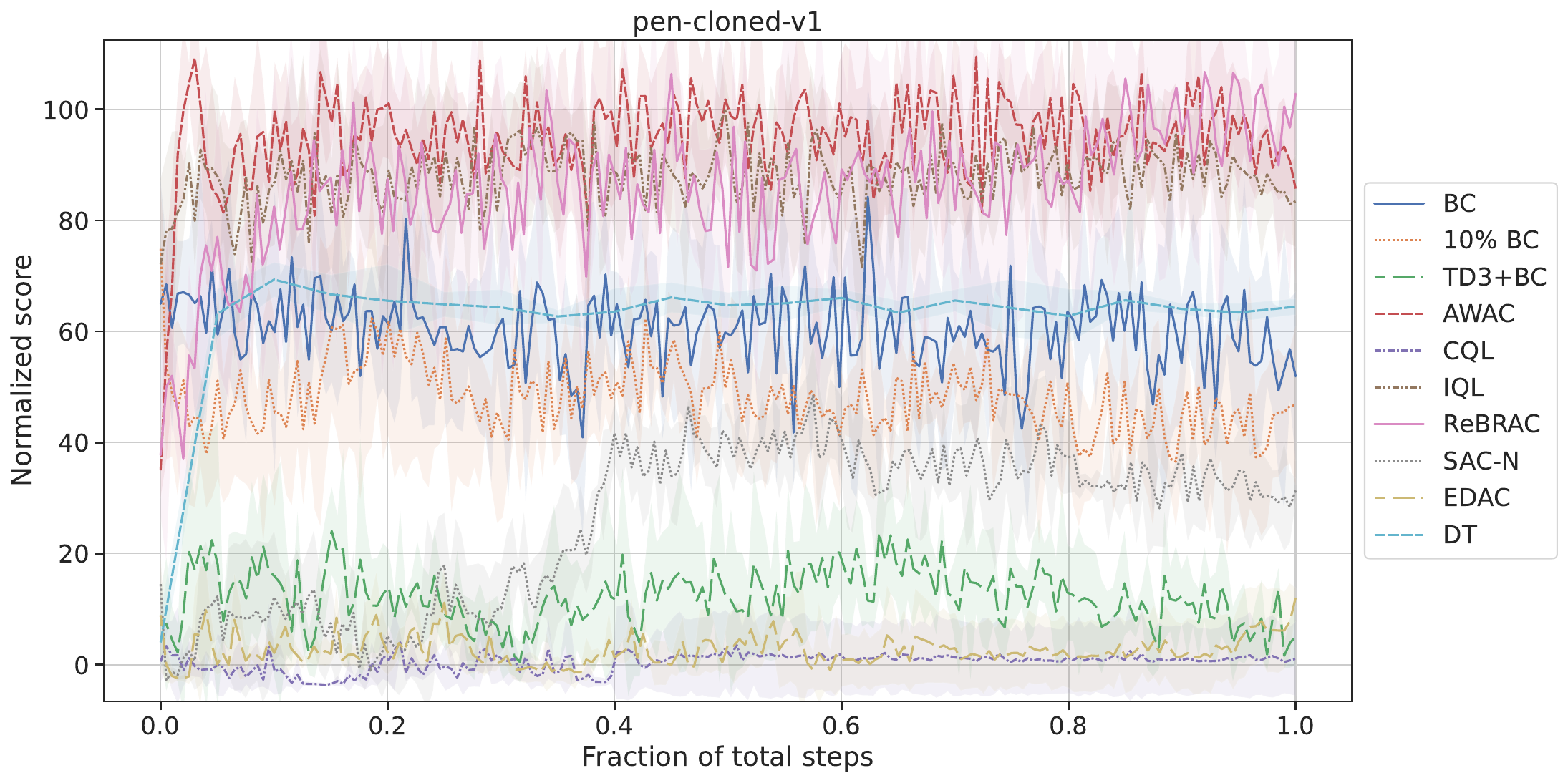}
         \caption{}
         \label{fig:pen-c}
     \end{subfigure}
     \begin{subfigure}[b]{0.32\textwidth}
         \centering
         \includegraphics[width=\textwidth]{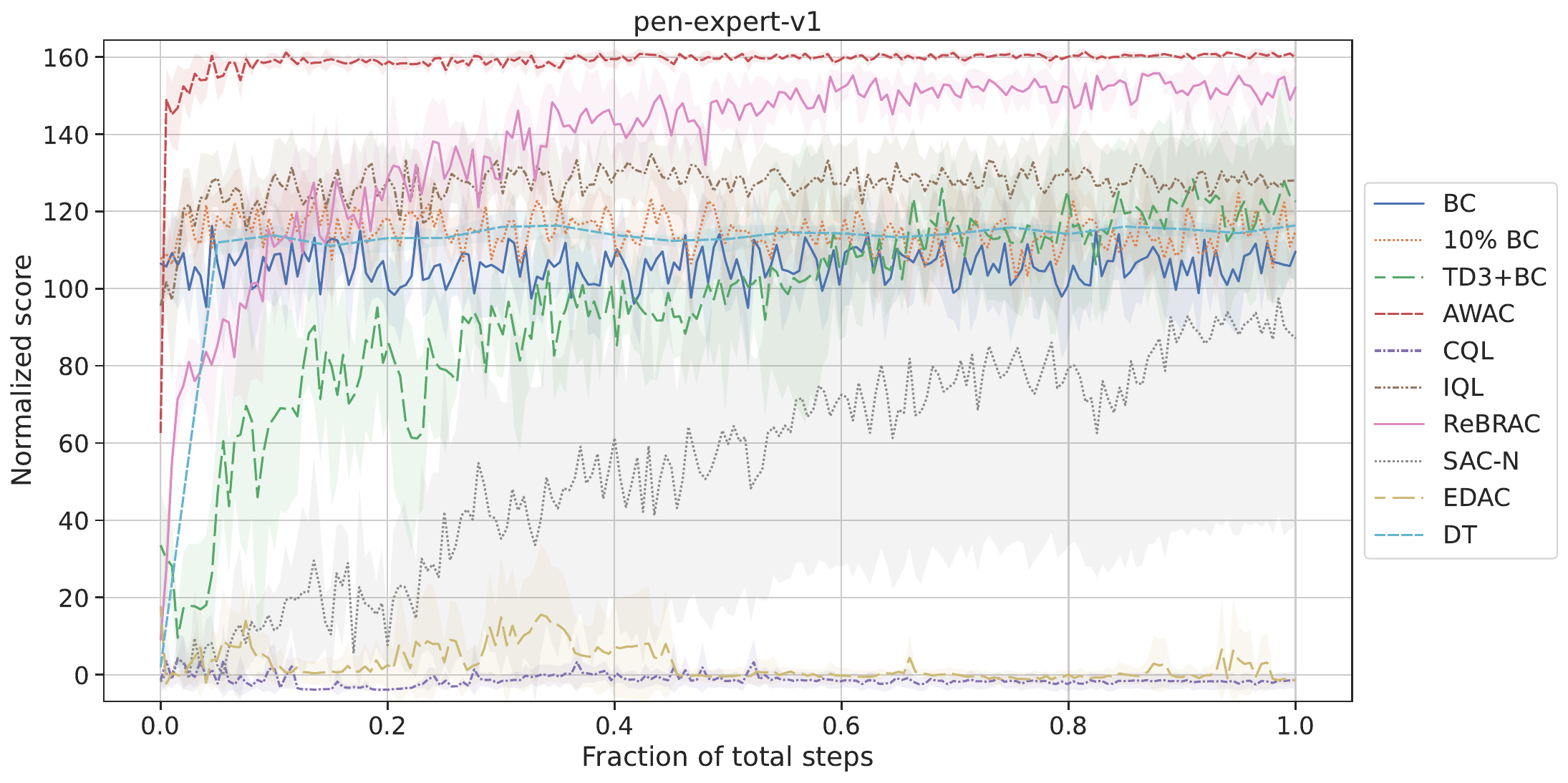}
         \caption{}
         \label{fig:pen-e}
     \end{subfigure}
    \caption{Training curves for Pen task.\\ (a) Human dataset, (b) Colned dataset, (c) Expert dataset}
        \label{fig:pen_curves}
\end{figure}

\begin{figure}[!ht]
\centering
\captionsetup{justification=centering}
     \centering
     \begin{subfigure}[b]{0.32\textwidth}
         \centering
         \includegraphics[width=\textwidth]{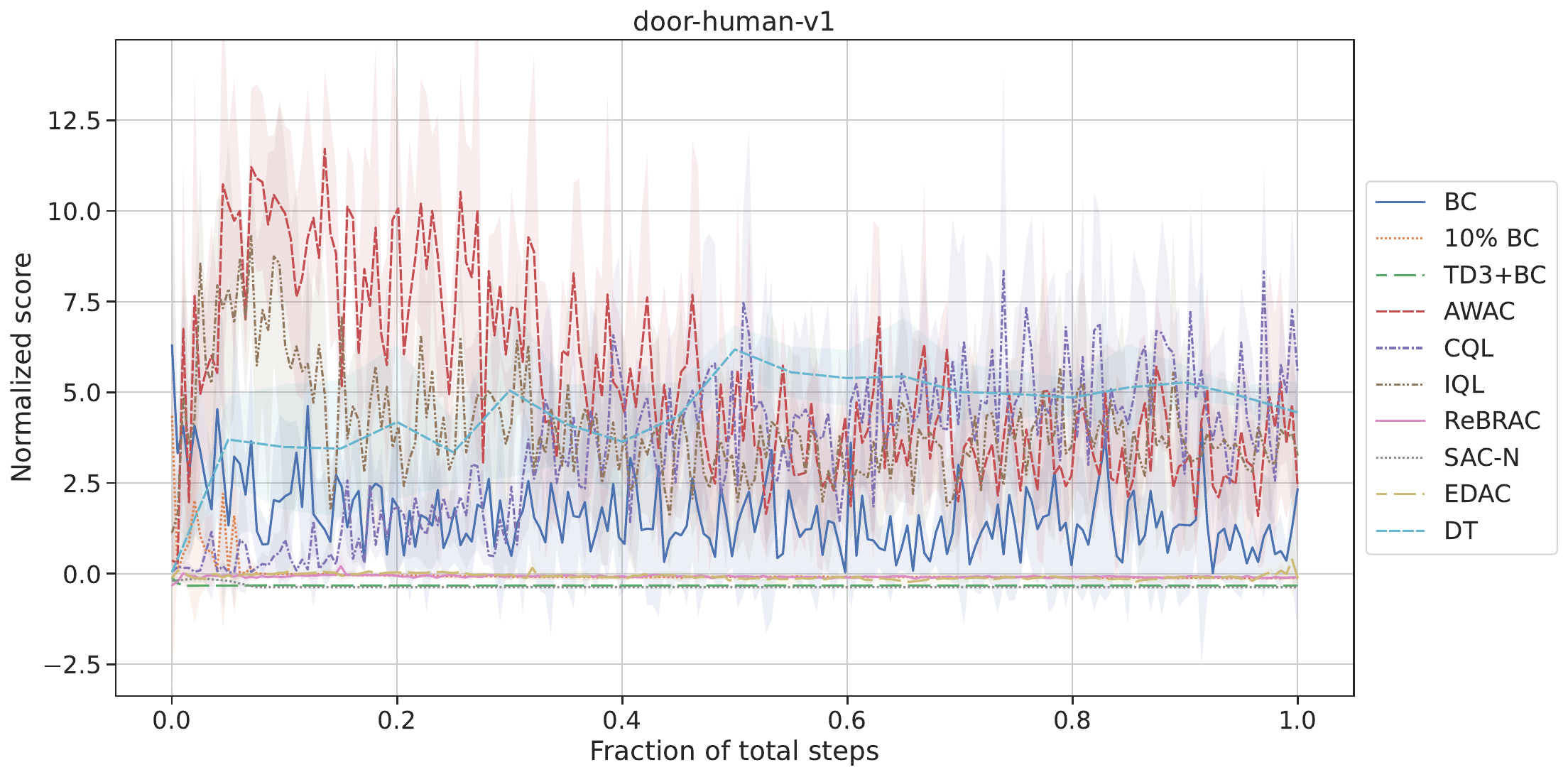}
         \caption{}
         \label{fig:door-h}
     \end{subfigure}
     \hfill
     \begin{subfigure}[b]{0.32\textwidth}
         \centering
         \includegraphics[width=\textwidth]{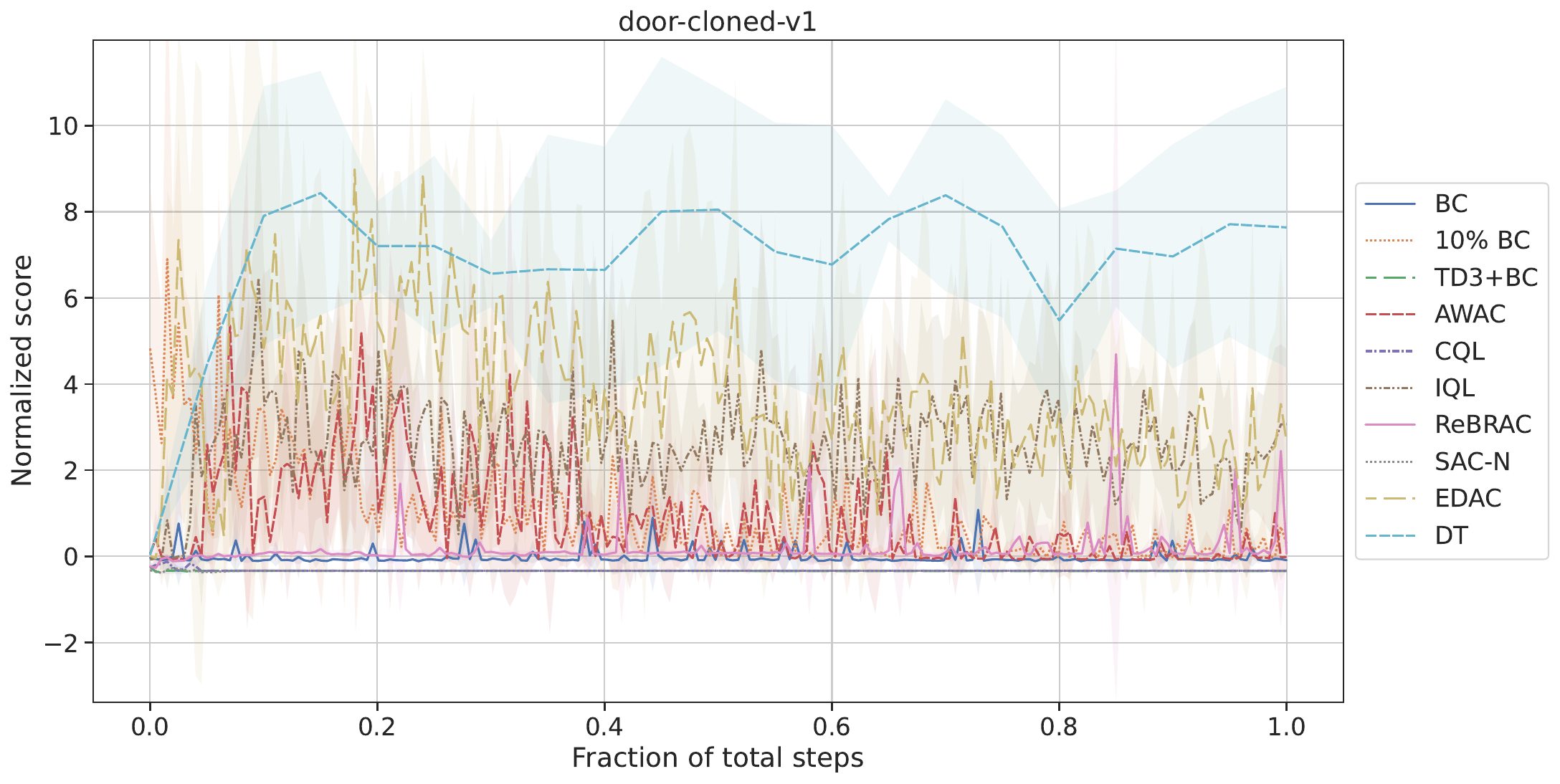}
         \caption{}
         \label{fig:door-c}
     \end{subfigure}
     \begin{subfigure}[b]{0.32\textwidth}
         \centering
         \includegraphics[width=\textwidth]{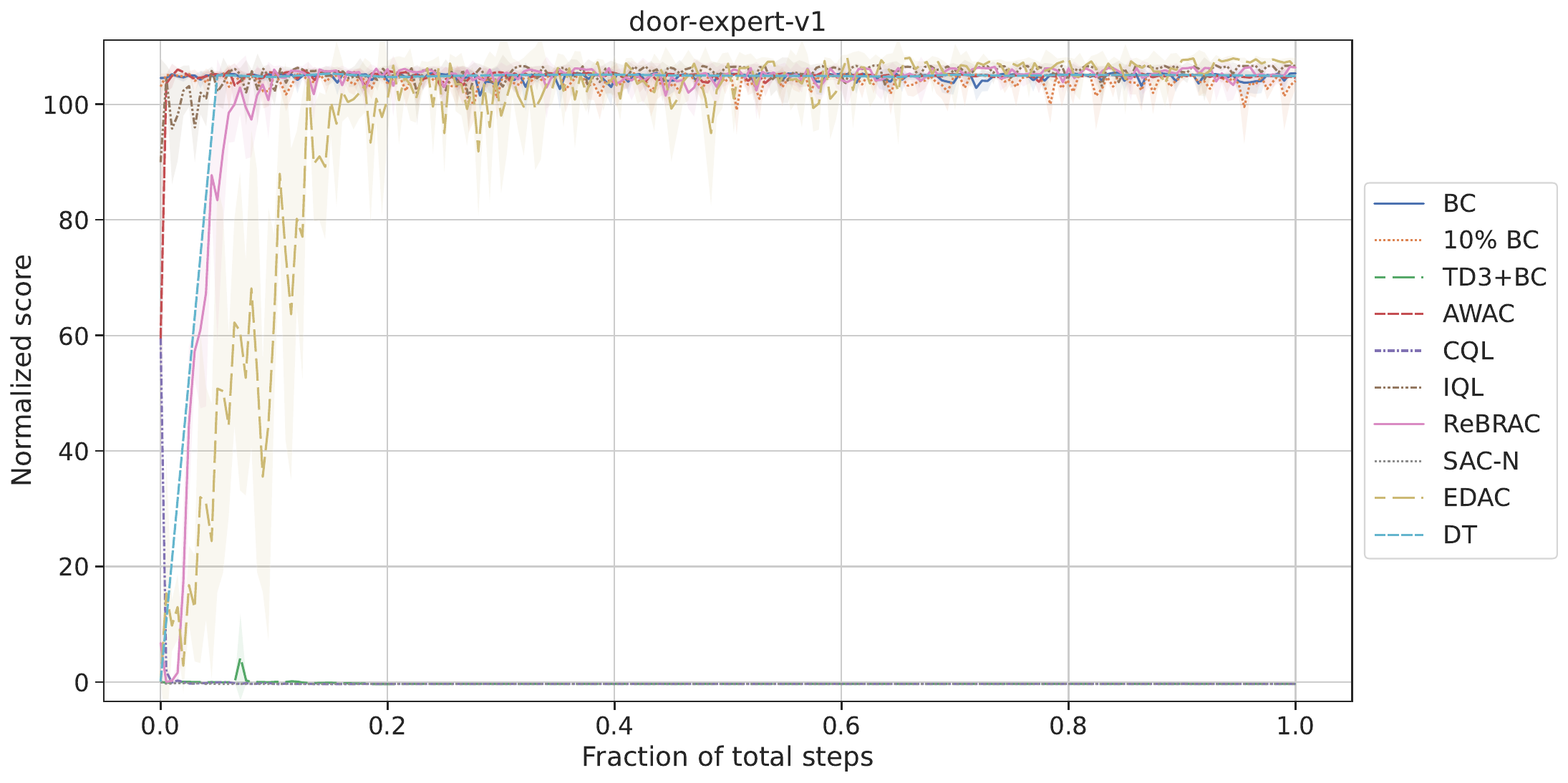}
         \caption{}
         \label{fig:door-e}
     \end{subfigure}
    \caption{Training curves for Door task.\\ (a) Human dataset, (b) Colned dataset, (c) Expert dataset}
        \label{fig:door_curves}
\end{figure}

\begin{figure}[!ht]
\centering
\captionsetup{justification=centering}
     \centering
     \begin{subfigure}[b]{0.32\textwidth}
         \centering
         \includegraphics[width=\textwidth]{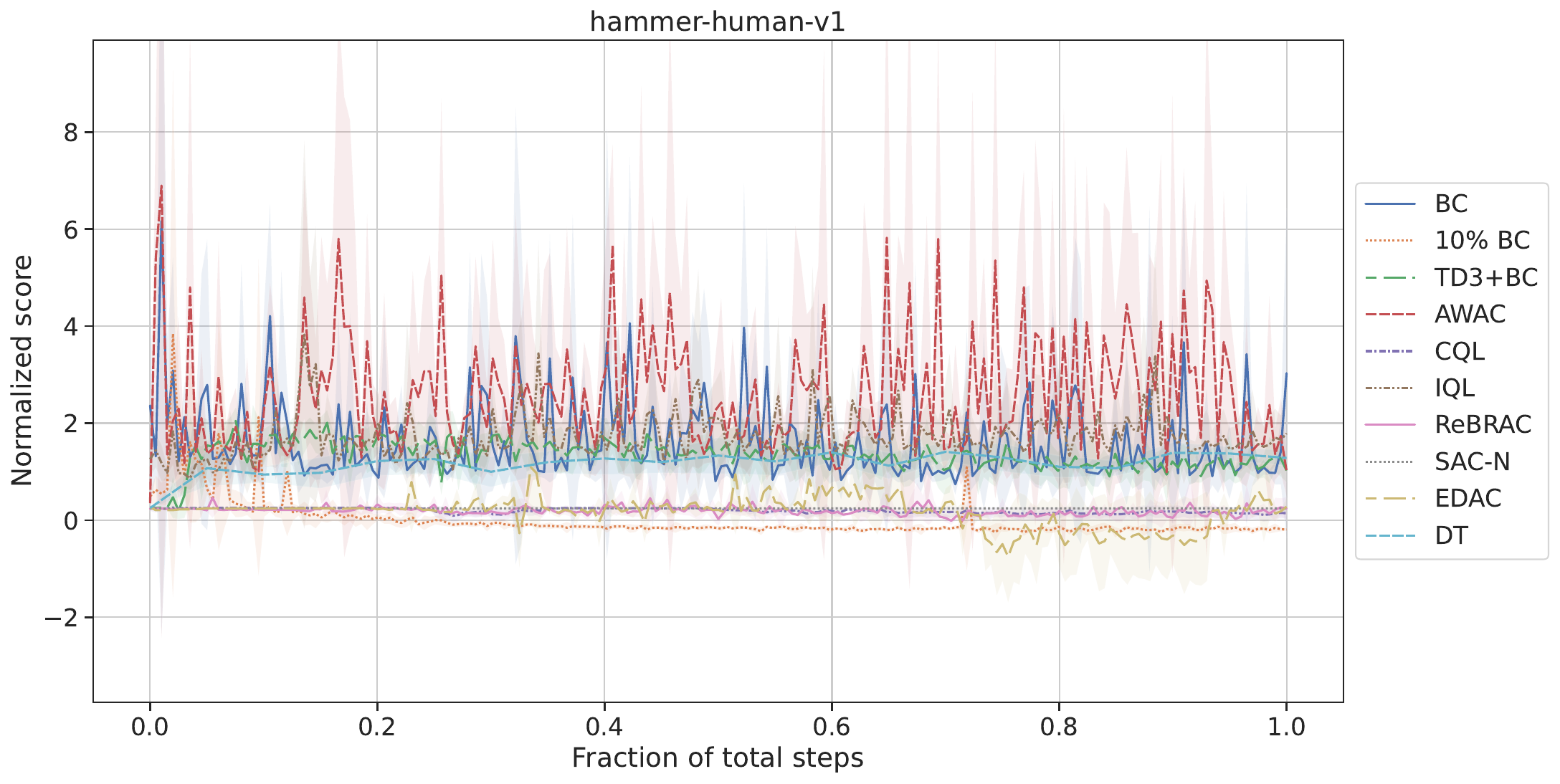}
         \caption{}
         \label{fig:hammer-h}
     \end{subfigure}
     \hfill
     \begin{subfigure}[b]{0.32\textwidth}
         \centering
         \includegraphics[width=\textwidth]{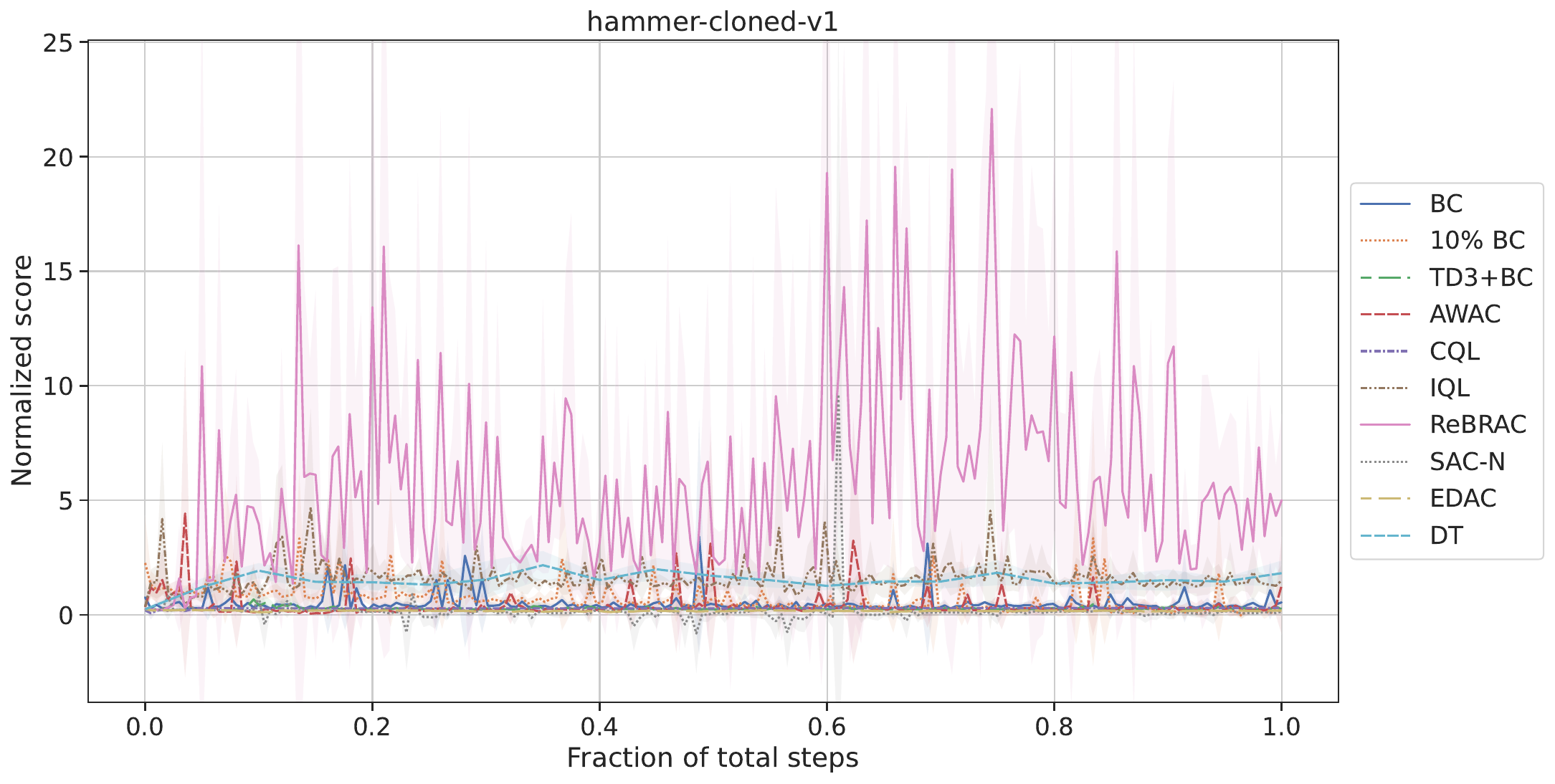}
         \caption{}
         \label{fig:hammer-c}
     \end{subfigure}
     \begin{subfigure}[b]{0.32\textwidth}
         \centering
         \includegraphics[width=\textwidth]{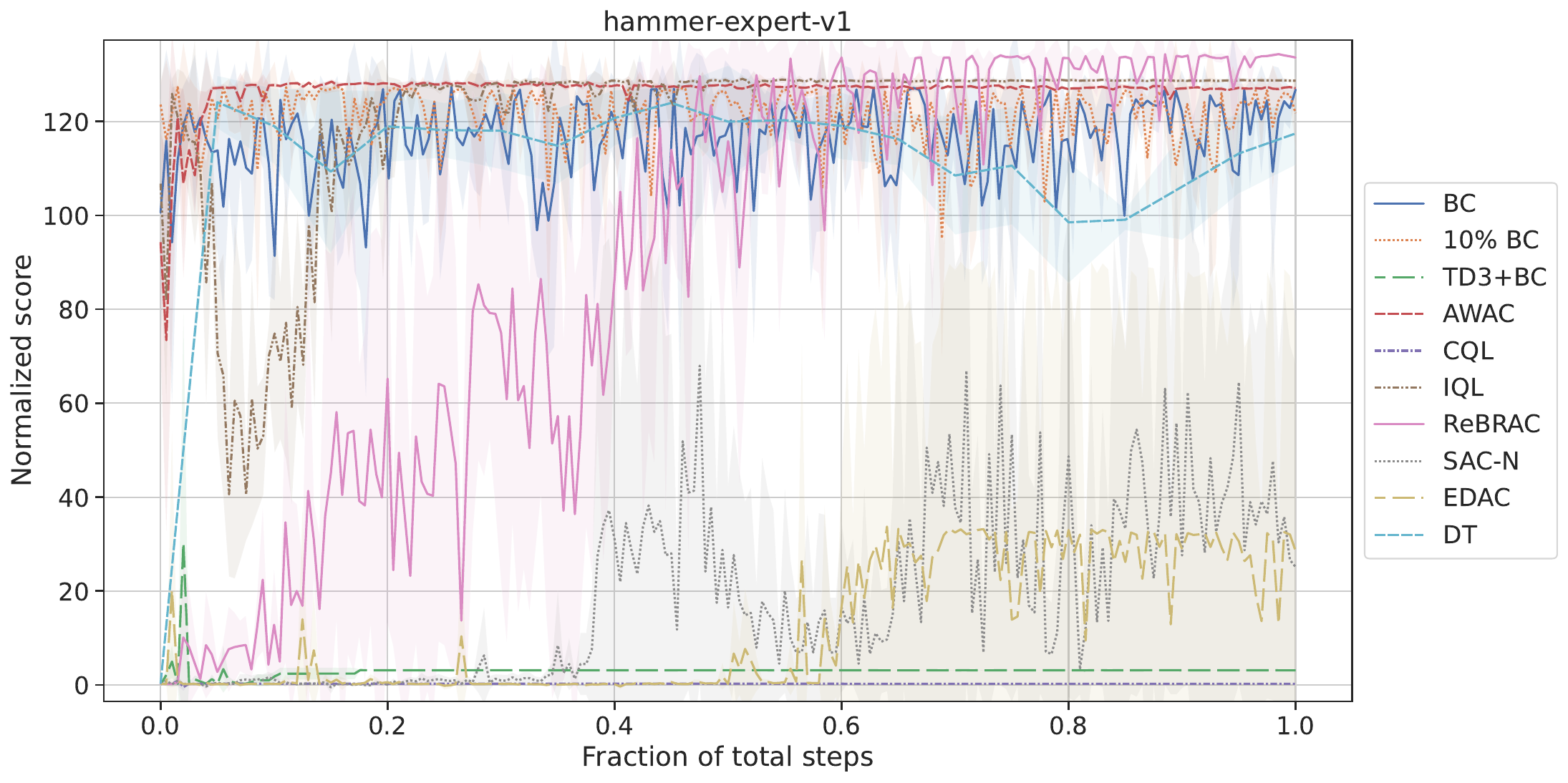}
         \caption{}
         \label{fig:hammer-e}
     \end{subfigure}
    \caption{Training curves for Hammer task.\\ (a) Human dataset, (b) Colned dataset, (c) Expert dataset}
        \label{fig:hammer_curves}
\end{figure}

\begin{figure}[!ht]
\centering
\captionsetup{justification=centering}
     \centering
     \begin{subfigure}[b]{0.32\textwidth}
         \centering
         \includegraphics[width=\textwidth]{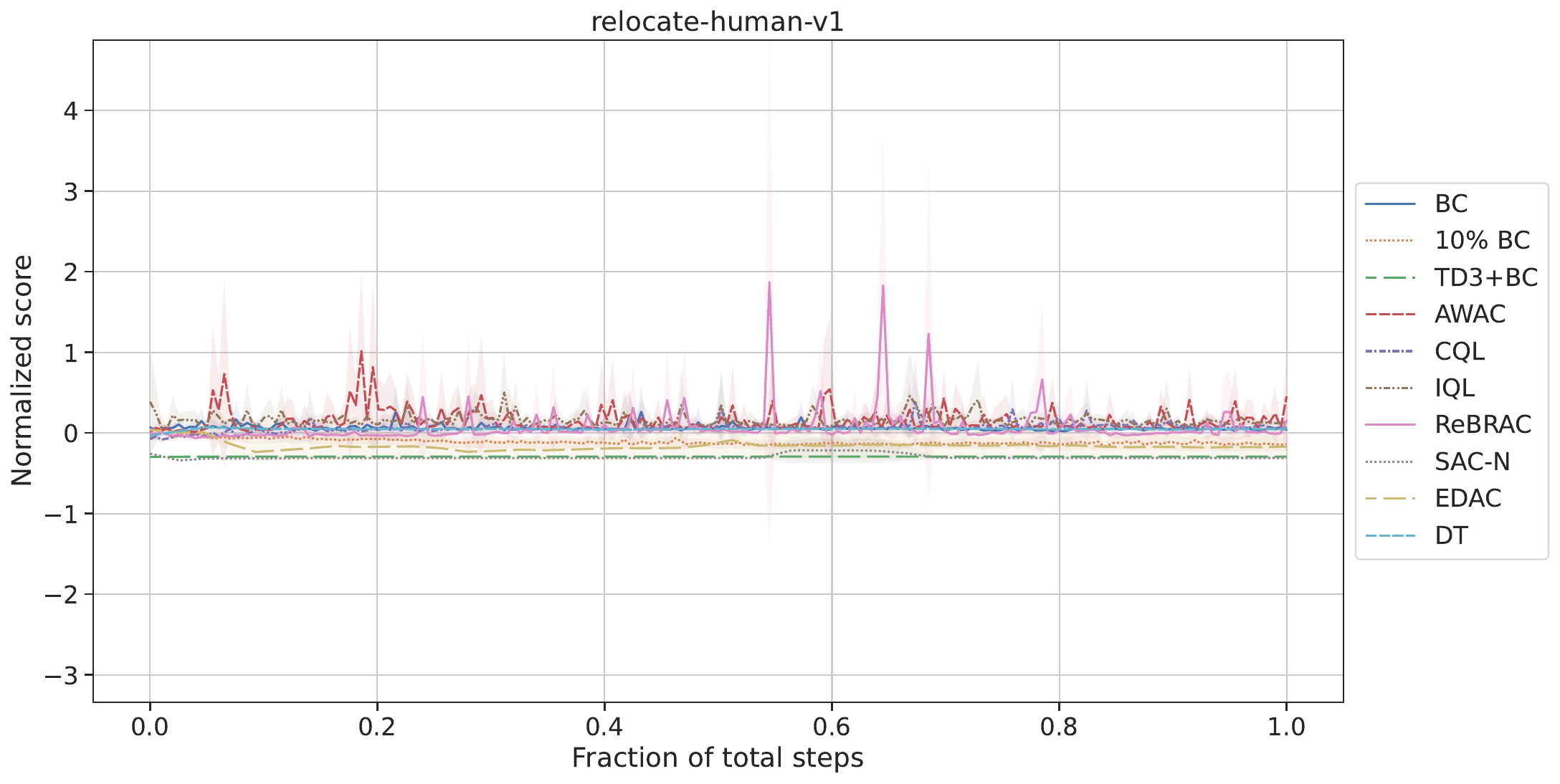}
         \caption{}
         \label{fig:relocate-h}
     \end{subfigure}
     \hfill
     \begin{subfigure}[b]{0.32\textwidth}
         \centering
         \includegraphics[width=\textwidth]{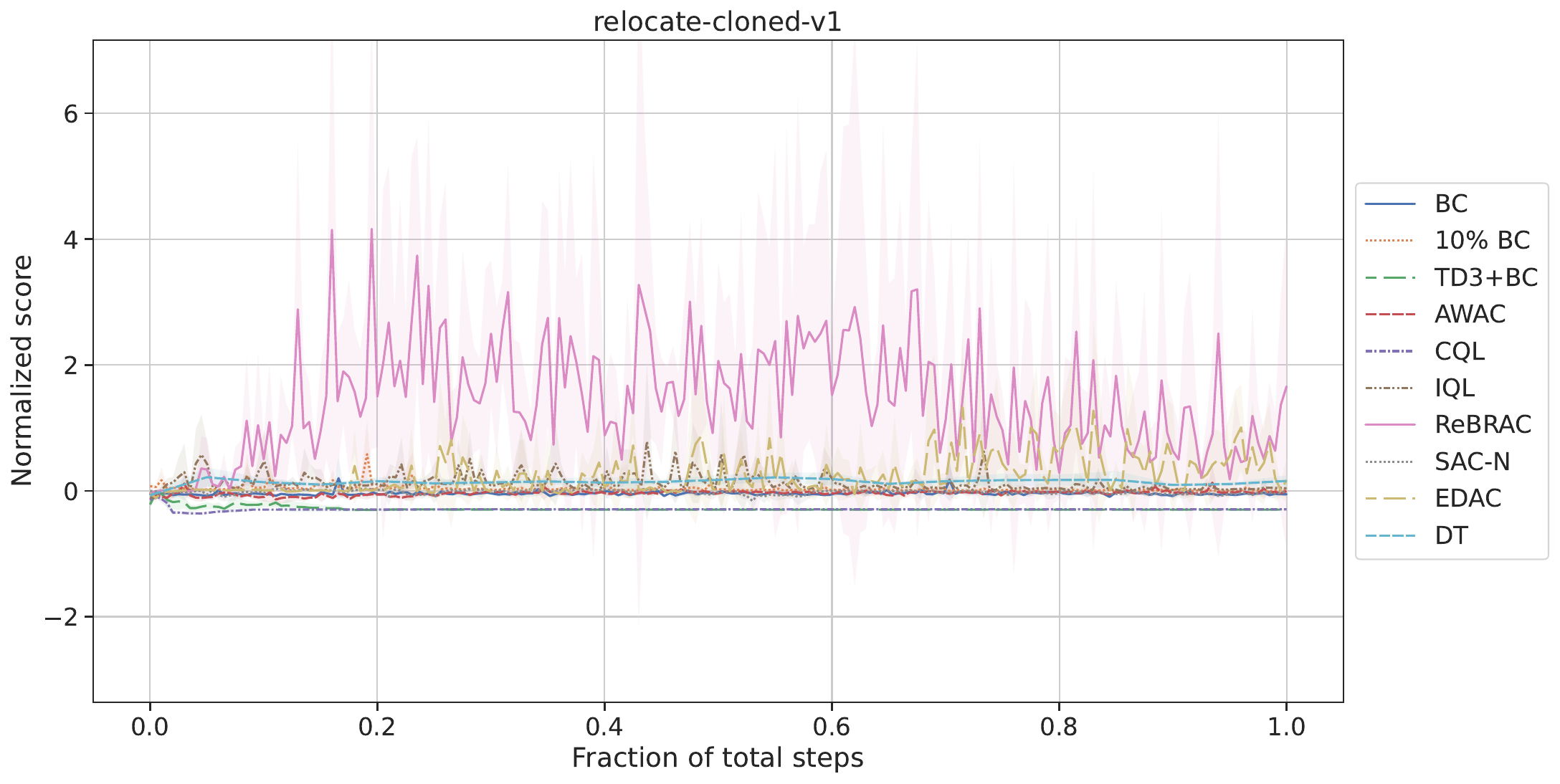}
         \caption{}
         \label{fig:relocate-c}
     \end{subfigure}
     \begin{subfigure}[b]{0.32\textwidth}
         \centering
         \includegraphics[width=\textwidth]{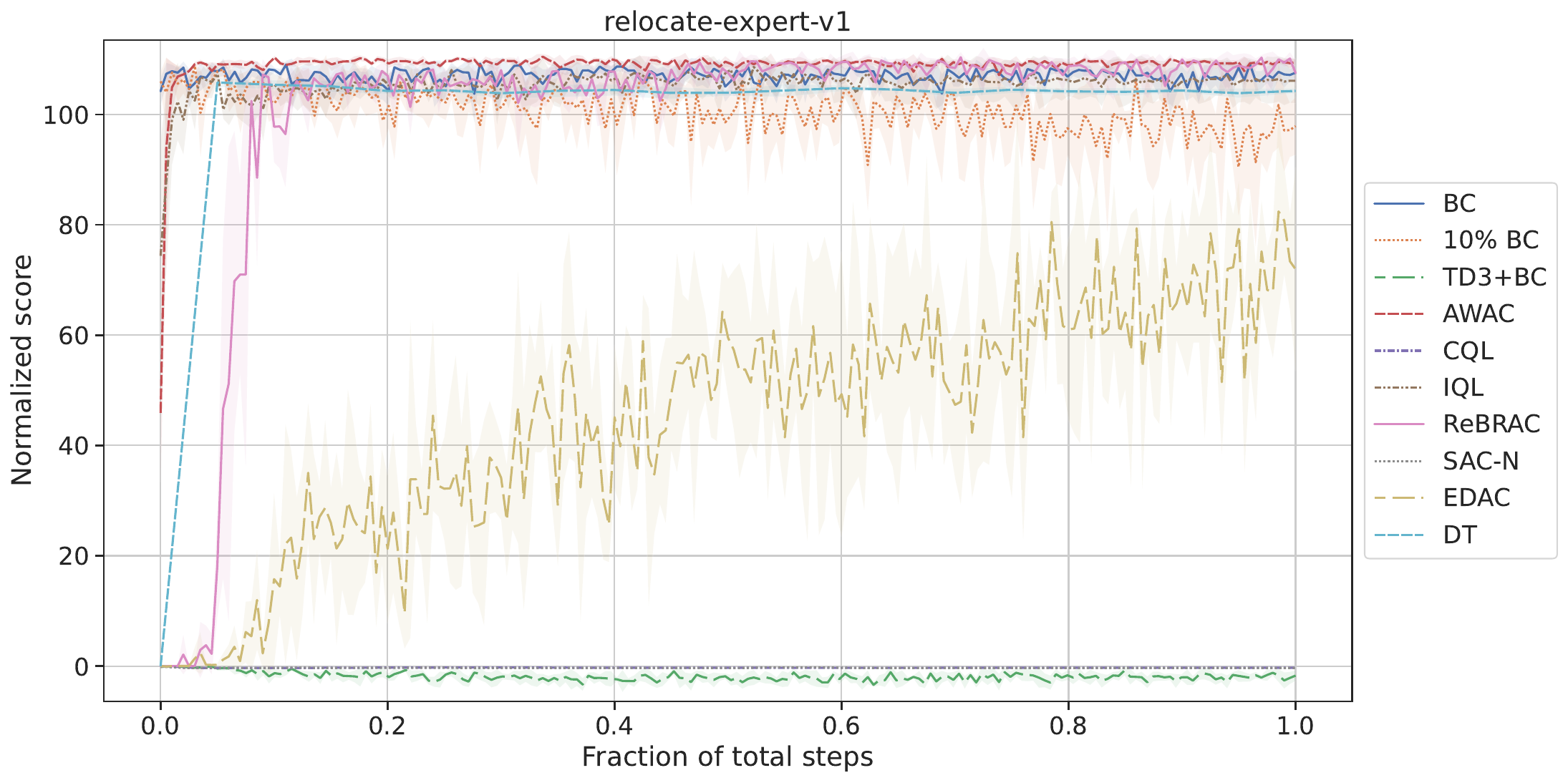}
         \caption{}
         \label{fig:relocate-e}
     \end{subfigure}
    \caption{Training curves for Relocate task.\\ (a) Human dataset, (b) Colned dataset, (c) Expert dataset}
        \label{fig:relocate_curves}
\end{figure}
\clearpage
\subsection{Offline-to-online}
\begin{figure}[ht]
\centering
\captionsetup{justification=centering}
     \centering
     \begin{subfigure}[b]{0.49\textwidth}
         \centering
         \includegraphics[width=\textwidth]{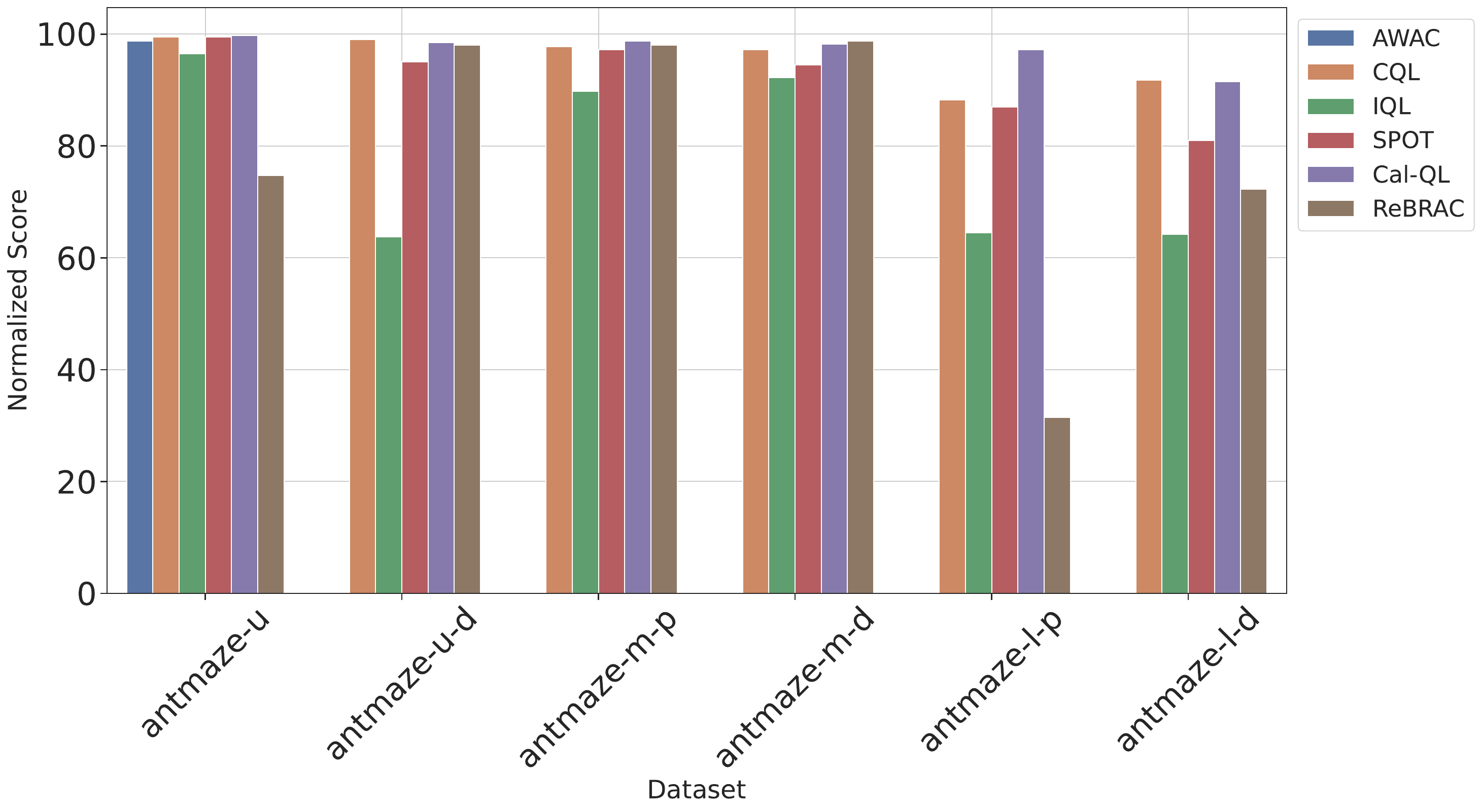}
         \caption{}
     \end{subfigure}
     \hfill
     \begin{subfigure}[b]{0.49\textwidth}
         \centering
         \includegraphics[width=\textwidth]{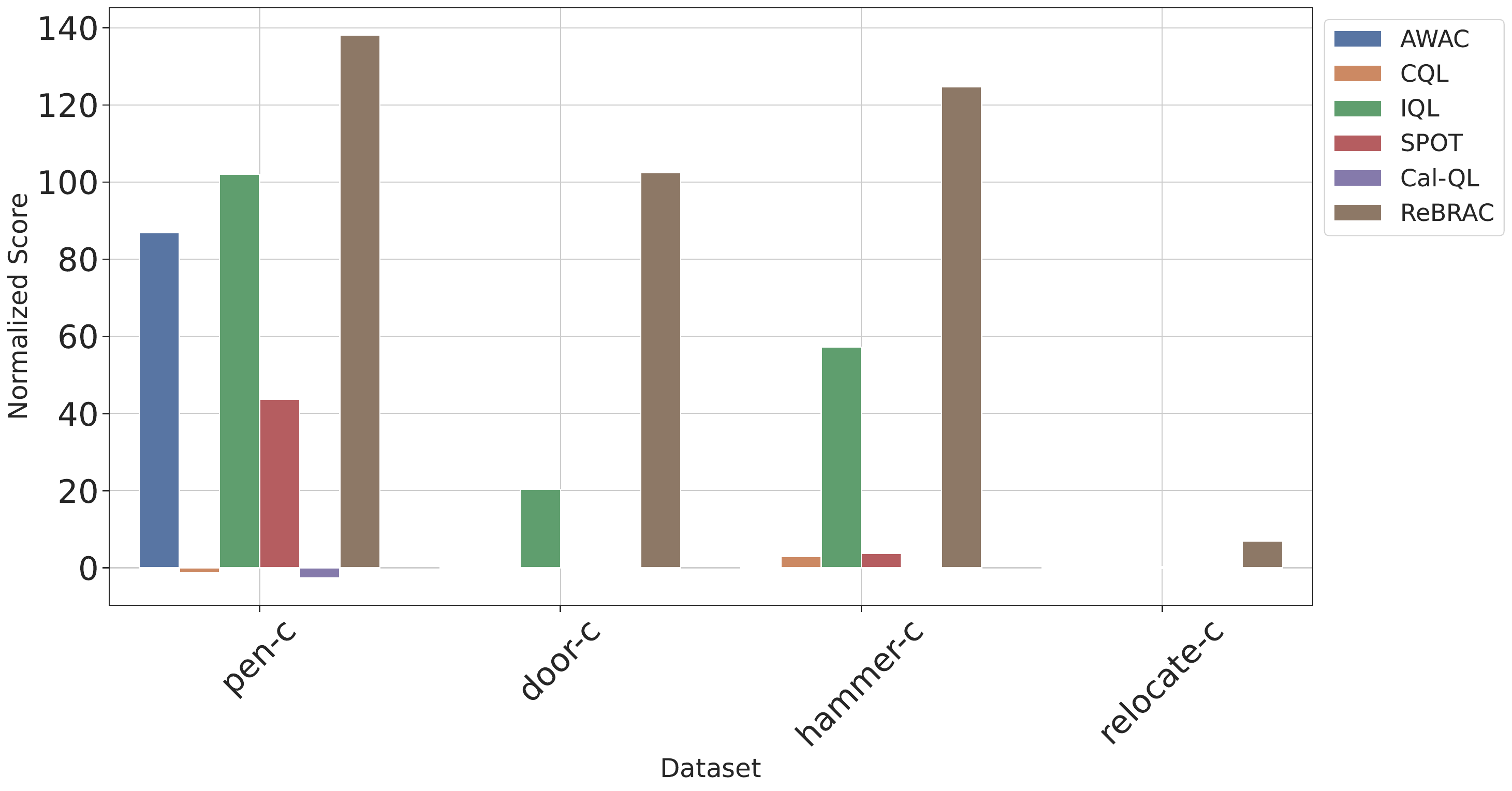}
         \caption{}
     \end{subfigure}
        \caption{Graphical representation of the normalized performance of the last trained policy on D4RL after online tuning averaged over 4 random seeds.\\
        % normalized average over last scores on D4RL Gym tasks, averaged over 4 random seeds.
        (a) AntMaze datasets (b) Adroit datasets}
        \label{fig:last_bars_online}
\end{figure}        

\begin{figure}[ht]
\centering
\captionsetup{justification=centering}
     \centering
     \begin{subfigure}[b]{0.32\textwidth}
         \centering
         \includegraphics[width=\textwidth]{img/antmaze-umaze-v2.pdf}
         \caption{}
     \end{subfigure}
     \hfill
     \begin{subfigure}[b]{0.32\textwidth}
         \centering
         \includegraphics[width=\textwidth]{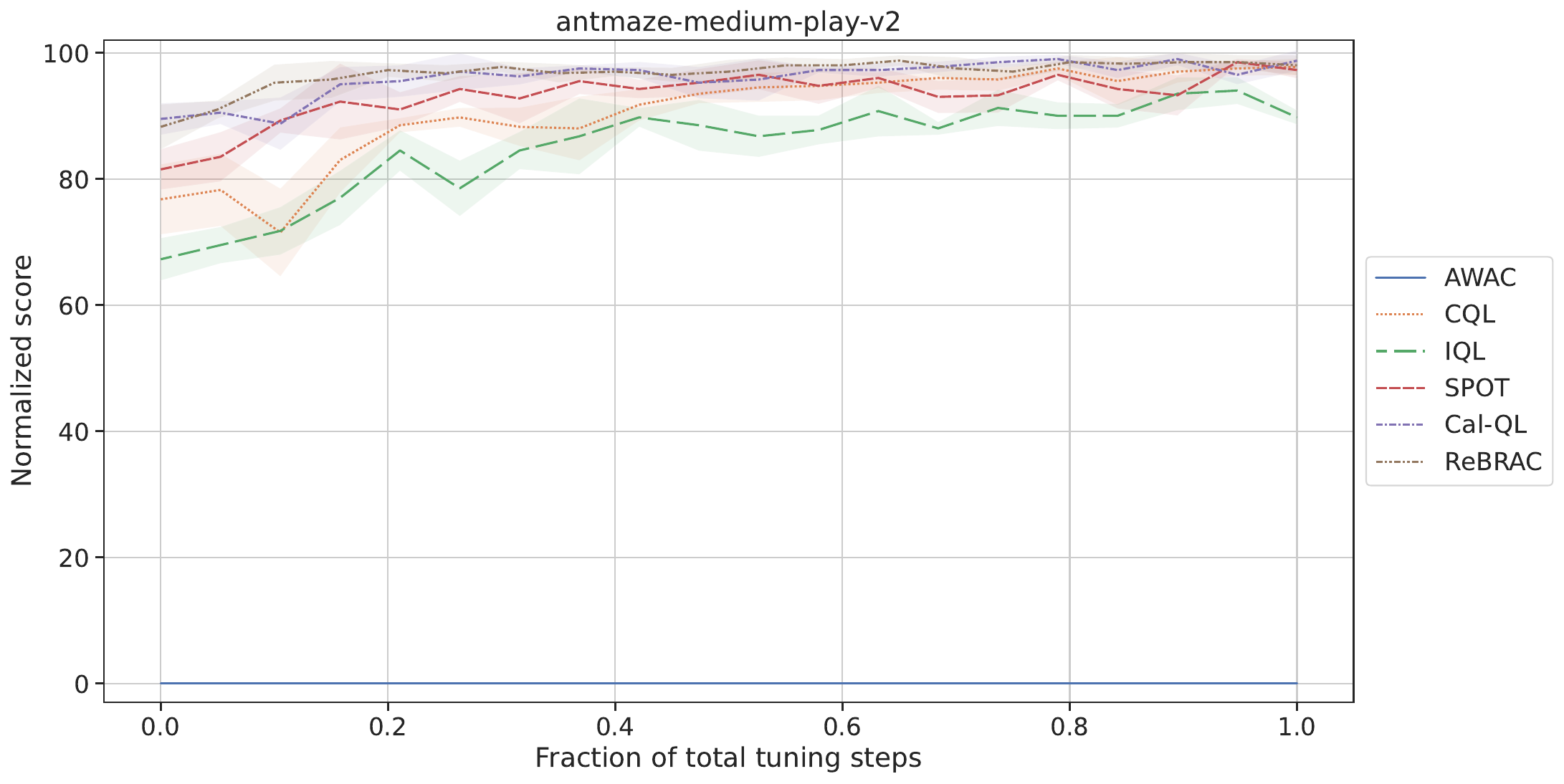}
         \caption{}
     \end{subfigure}
     \begin{subfigure}[b]{0.32\textwidth}
         \centering
         \includegraphics[width=\textwidth]{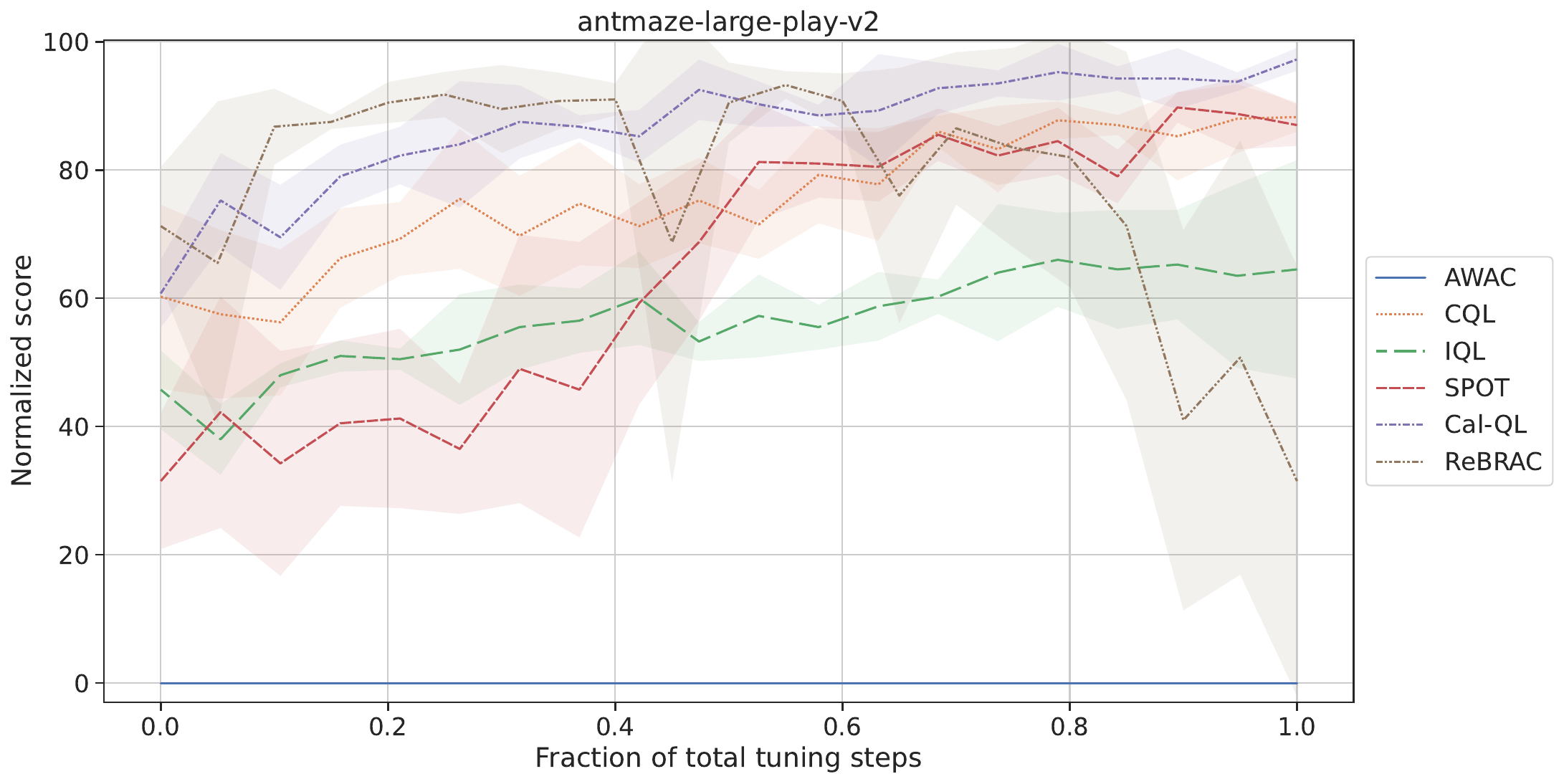}
         \caption{}
     \end{subfigure}
     \begin{subfigure}[b]{0.32\textwidth}
         \centering
         \includegraphics[width=\textwidth]{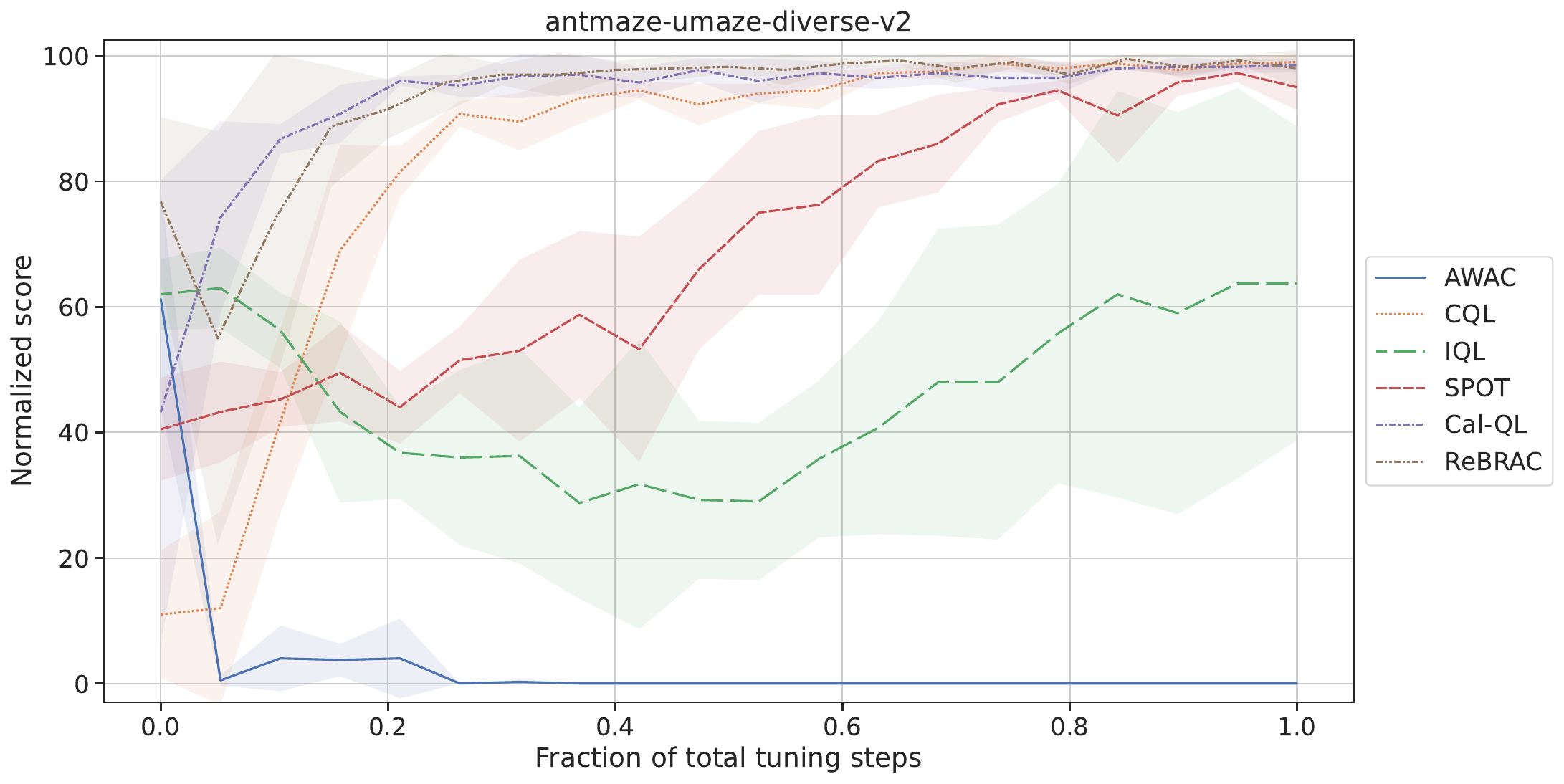}
         \caption{}
     \end{subfigure}
     \hfill
     \begin{subfigure}[b]{0.32\textwidth}
         \centering
         \includegraphics[width=\textwidth]{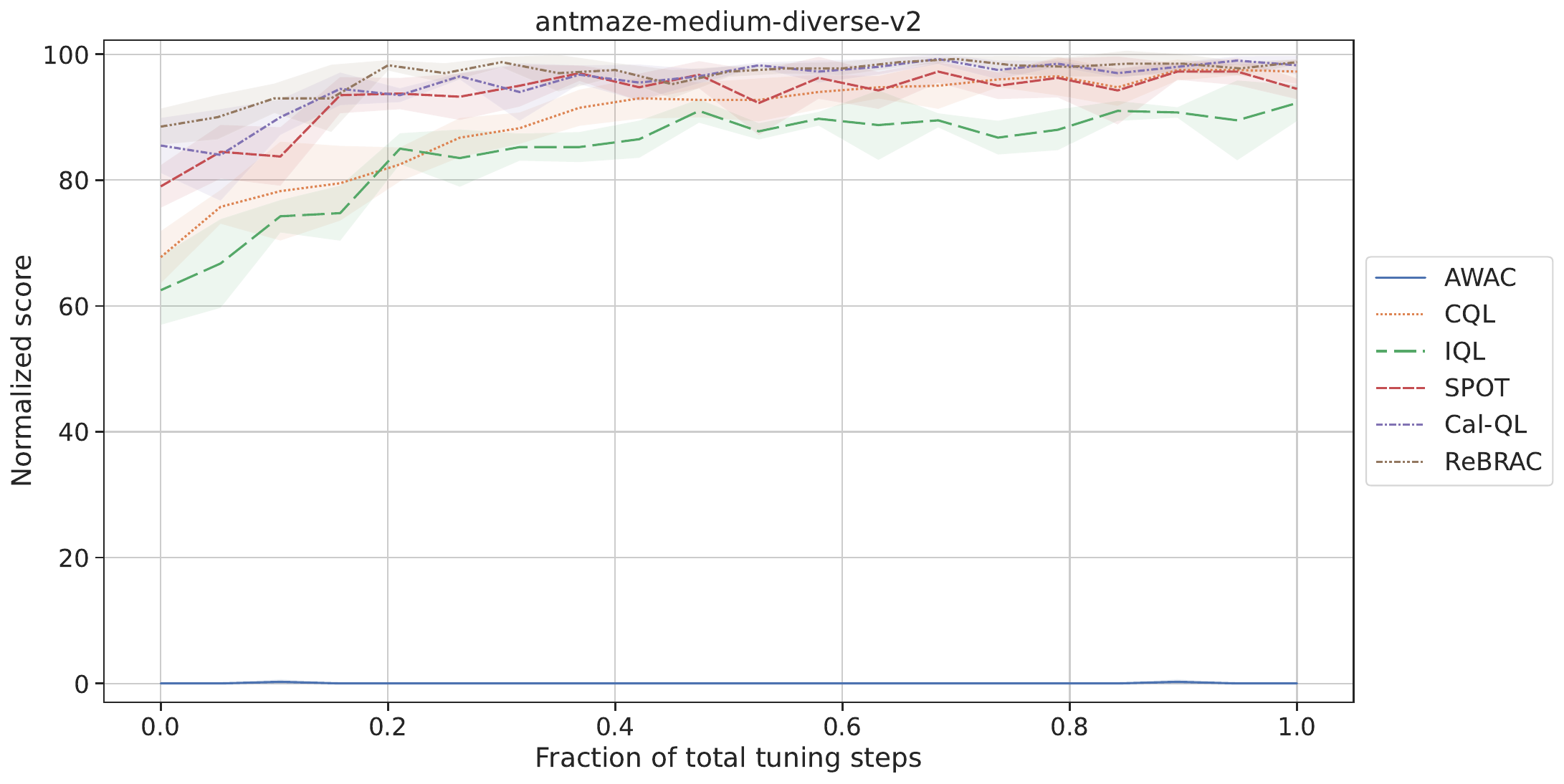}
         \caption{}
     \end{subfigure}
     \begin{subfigure}[b]{0.32\textwidth}
         \centering
         \includegraphics[width=\textwidth]{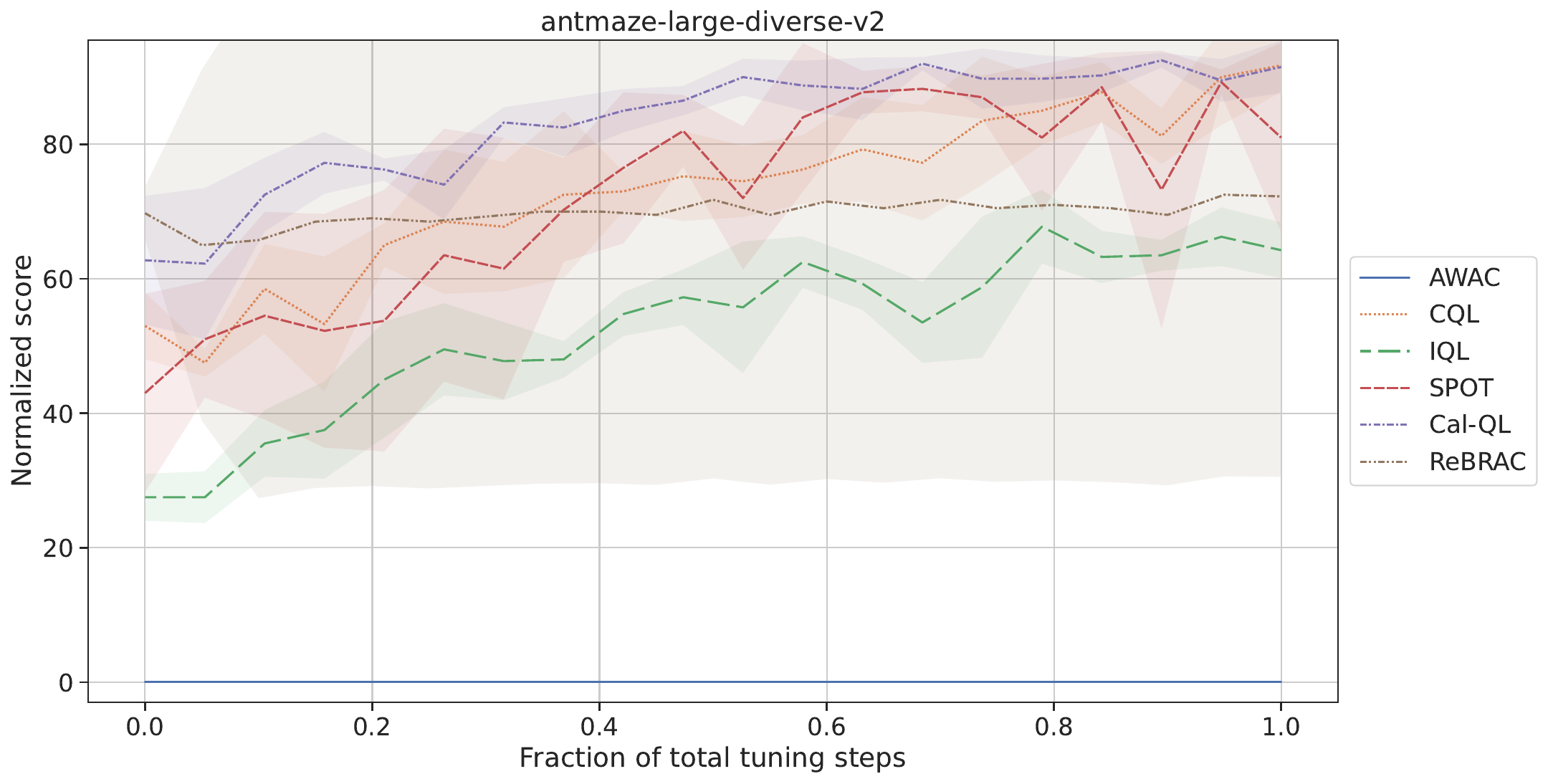}
         \caption{}
     \end{subfigure}
    \caption{Training curves for AntMaze task during online tuning.\\ (a) Umaze dataset, (b) Medium-play dataset, (c) Large-play dataset, (d) Umaze-diverse dataset, (e) Medium-diverse dataset, (f) Large-diverse dataset}
\end{figure}

\begin{figure}[ht]
\centering
\captionsetup{justification=centering}
     \centering
     \begin{subfigure}[b]{0.49\textwidth}
         \centering
         \includegraphics[width=\textwidth]{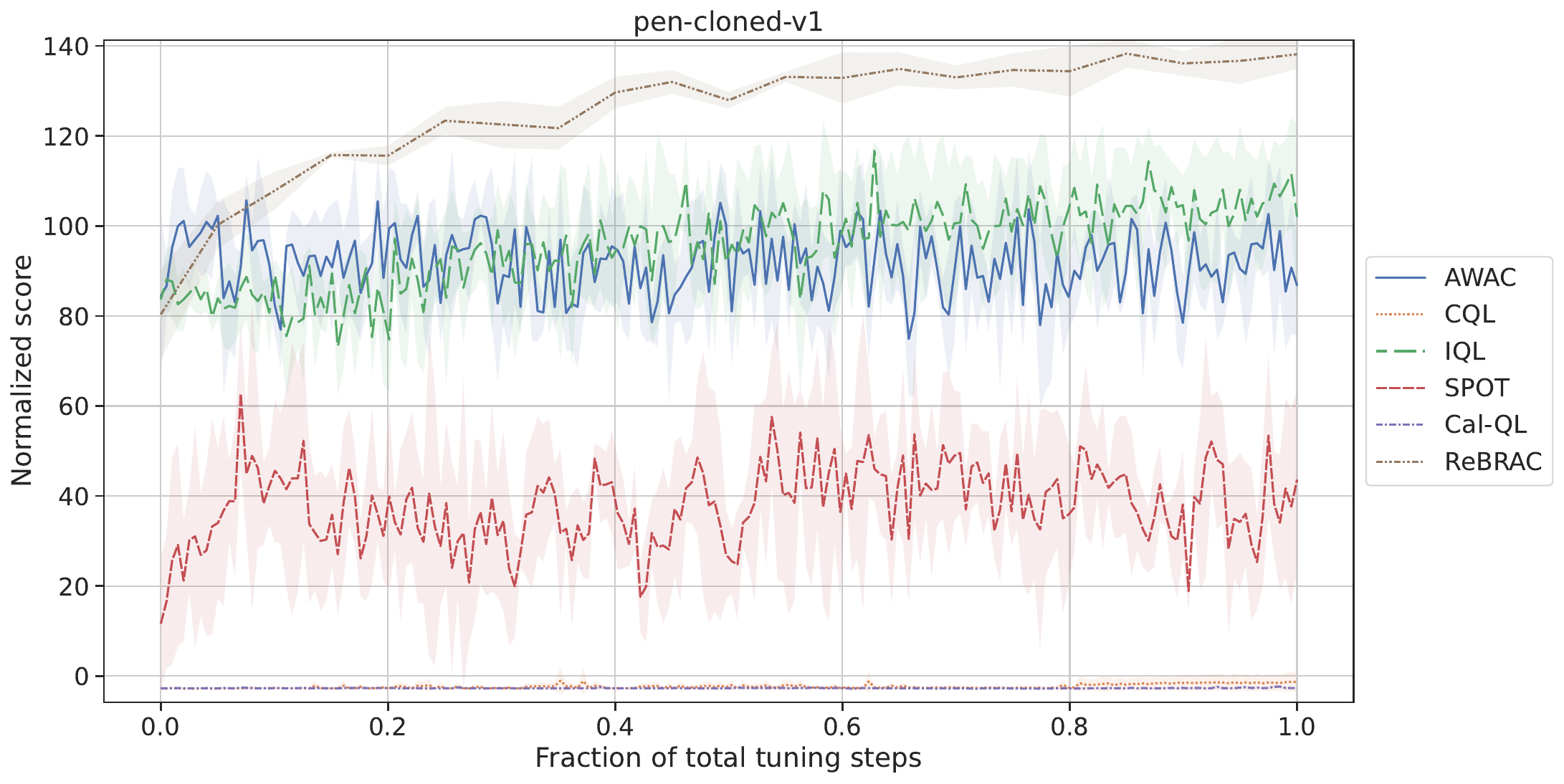}
         \caption{}
     \end{subfigure}
     \begin{subfigure}[b]{0.49\textwidth}
         \centering
         \includegraphics[width=\textwidth]{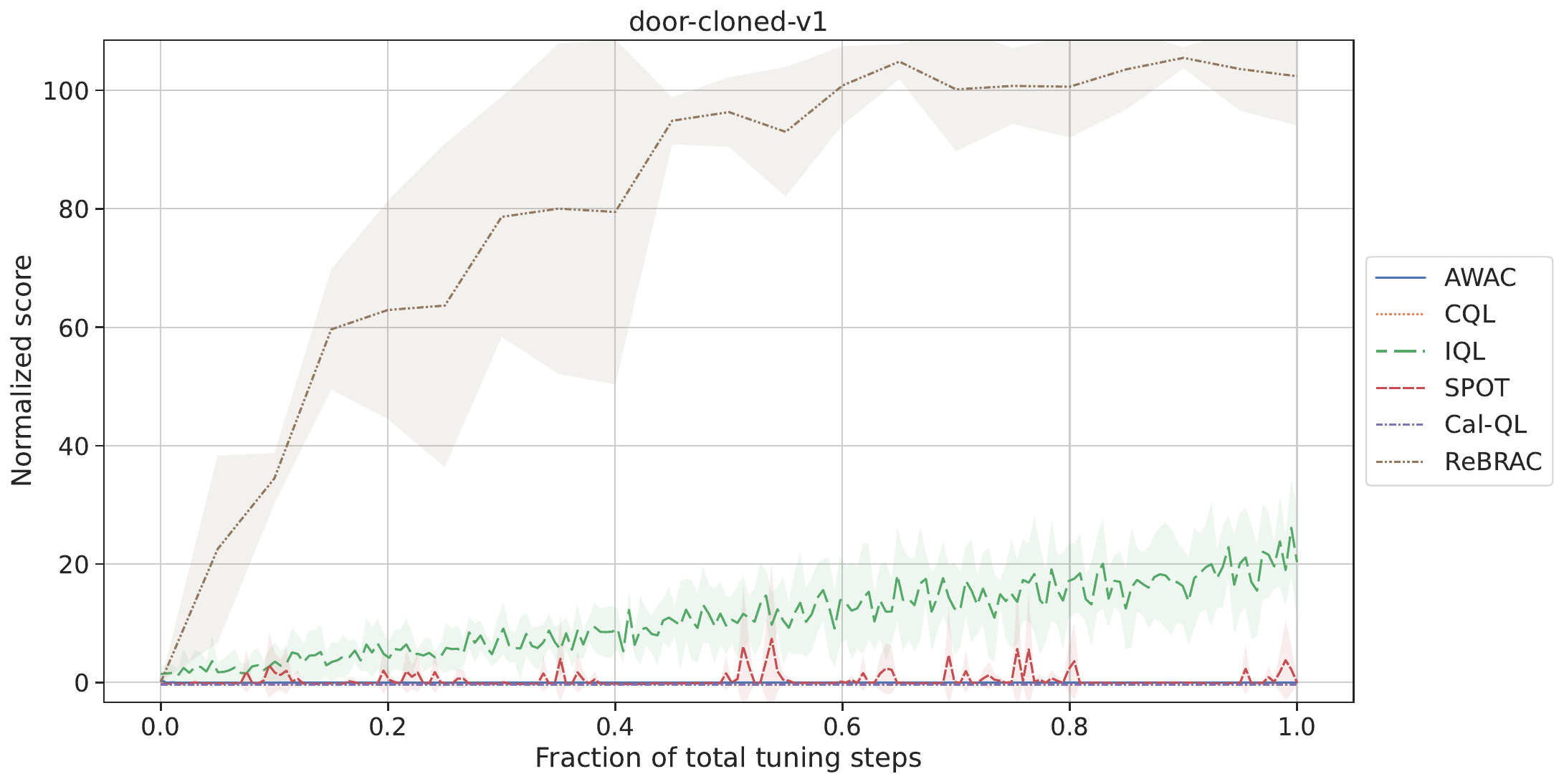}
         \caption{}
     \end{subfigure}
     \begin{subfigure}[b]{0.49\textwidth}
         \centering
         \includegraphics[width=\textwidth]{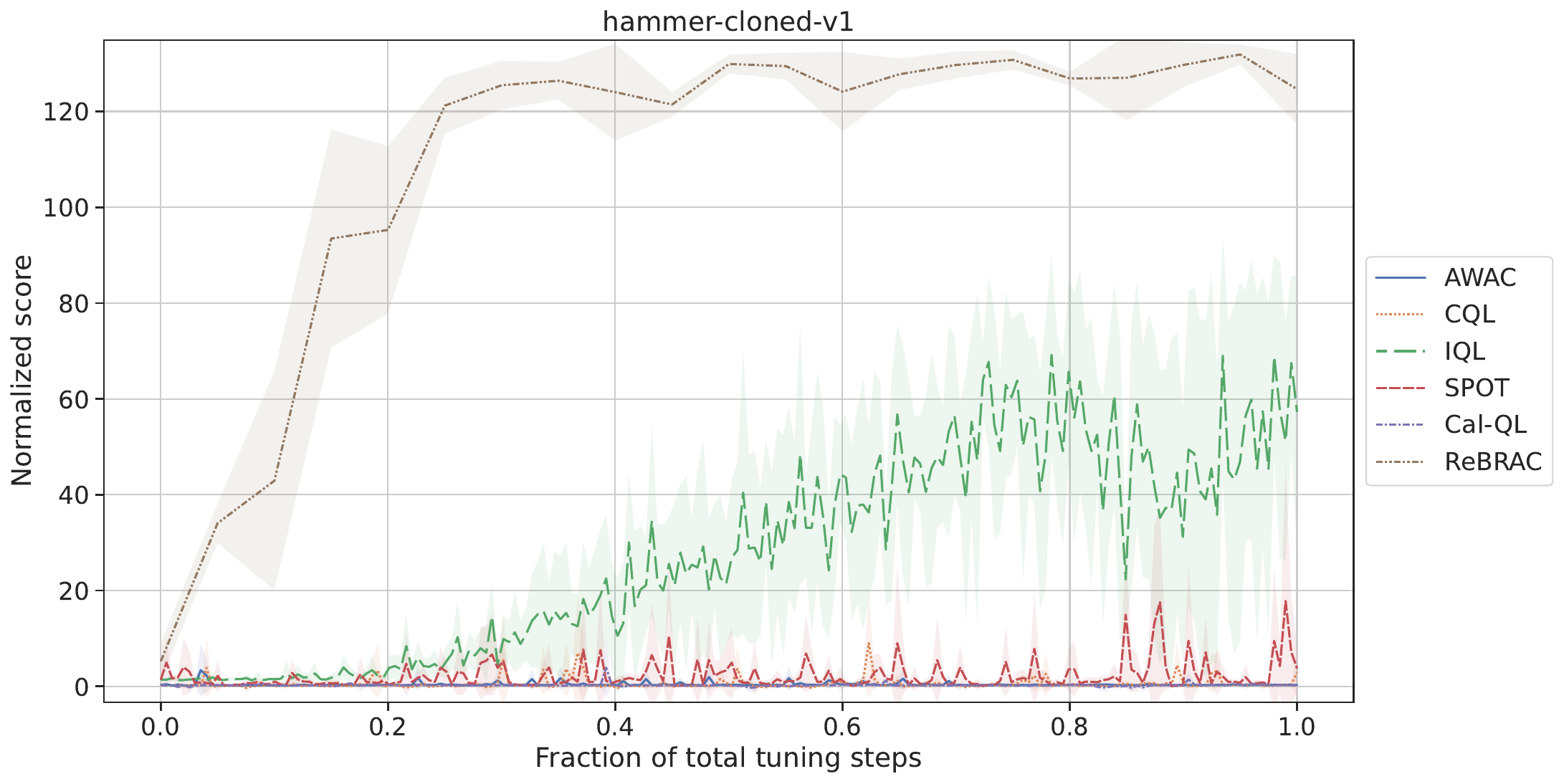}
         \caption{}
     \end{subfigure}
     \begin{subfigure}[b]{0.49\textwidth}
         \centering
         \includegraphics[width=\textwidth]{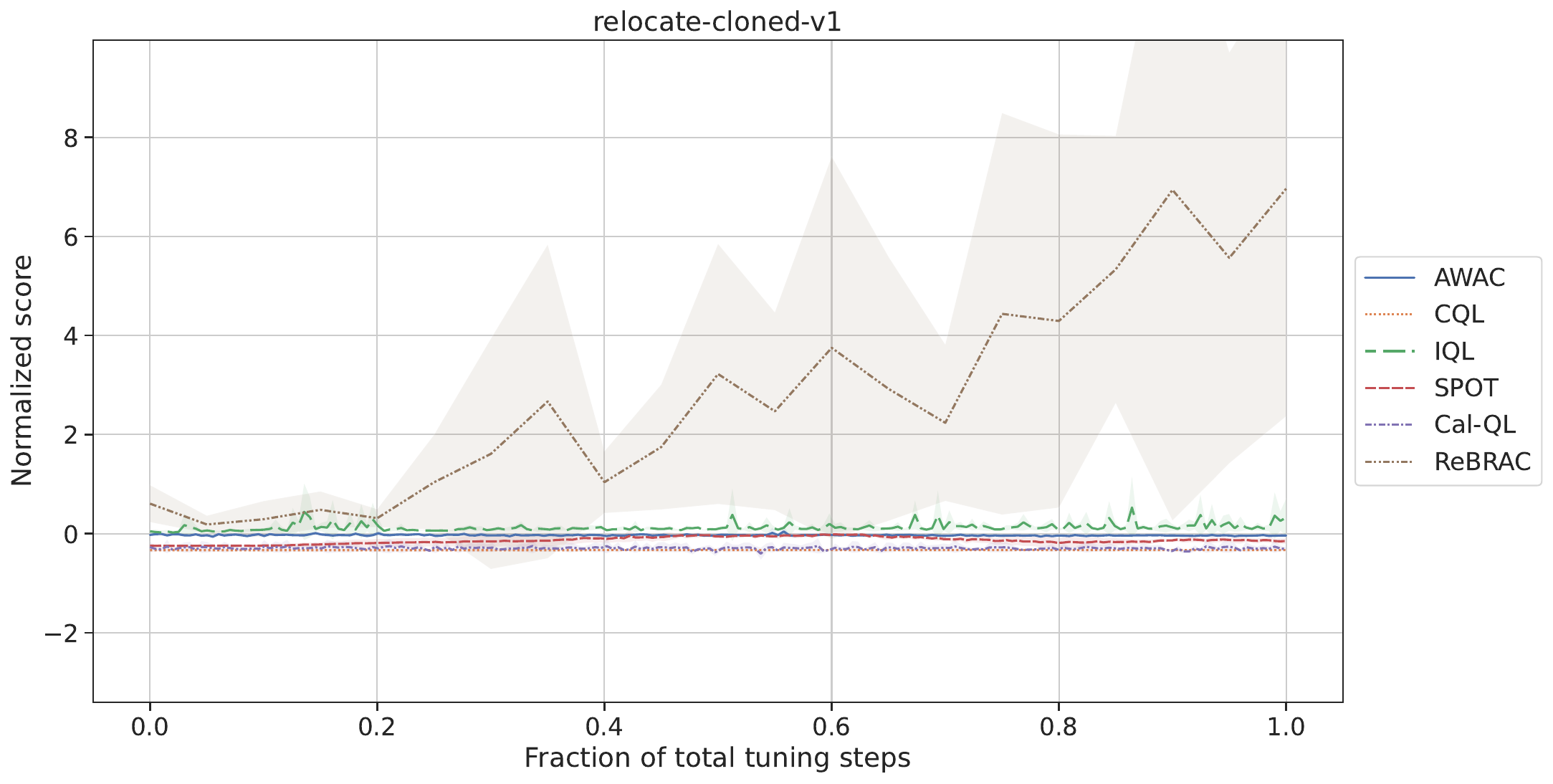}
         \caption{}
     \end{subfigure}
    \caption{Training curves for Adroit Cloned task during online tuning.\\ (a) Pen, (b) Door, (c) Hammer, (d) Relocate}
\end{figure}

\clearpage
\section{Weights\&Biases Tracking}
% ANY TEXT HERE?
\label{app:tracking}
\begin{figure}[ht]
\begin{center}
\centerline{\includegraphics[width=0.99\textwidth]{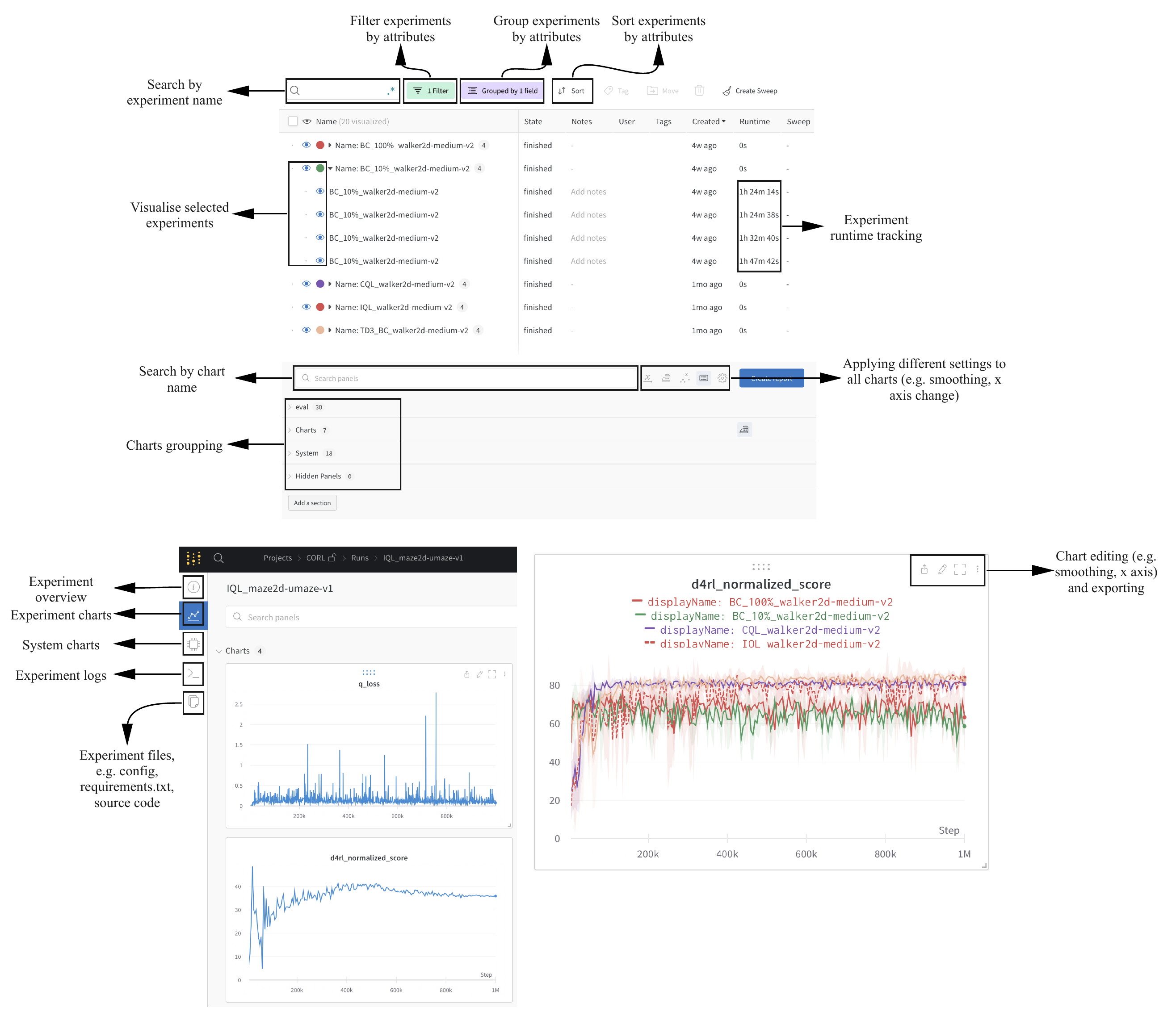}}
\caption{Screenshots of Weights\&Biases experiment tracking interface.}
\label{fig:wandb_ui}
\end{center}
\vskip -0.2in
\end{figure}

\section{License}
\label{appendix:license}
Our codebase is released under Apache License 2.0. The D4RL datasets \citep{d4rl} are released under Apache License 2.0.

\newpage

\section{Experimental Details}
\label{appendix:hyperparameters}
%footnote{\url{}}
We modify reward on AntMaze task by subtracting $1$ from reward as it is done in previous works except CQL and Cal-QL, where (0, 1) are mapped into (-5, 5).

We used original implementation of TD3 + BC\footnote{\url{https://github.com/sfujim/TD3_BC}}, SAC-$N$/EDAC\footnote{\label{sacn}\url{https://github.com/snu-mllab/EDAC}}, SPOT\footnote{\url{https://github.com/thuml/SPOT}}, ReBRAC\footnote{\url{https://github.com/tinkoff-ai/ReBRAC}}  and custom implementations of IQL\footnote{\url{https://github.com/gwthomas/IQL-PyTorch}} and CQL/Cal-QL\footnote{\url{https://github.com/young-geng/CQL}} as the basis for ours.

For most of the algorithms and datasets, we use default hyperparameters if available. Configuration files for every algorithm and environment are presented in our GitHub repository. Hyperparameters are also provided in \autoref{appendix:hyperparameters_list}.

All the experiments ran using V100 and A100 GPUs, which took approximately 5000 hours of compute in total.

\subsection{Number of update steps and evaluation rate}

Following original work, SAC-$N$ and EDAC are trained for 3 million steps (except AntMaze, which is trained for 1 million steps) in order to obtain state-of-the-art performance and tested every 10000 steps. Decision Transformer (DT) training is splitted into datasets pass epochs. We train DT for 50 epochs on each dataset and evaluate every 5 epochs. All other algorithms are trained for 1 million steps and evaluated every 5000 steps (50000 for AntMaze). We evaluate every policy for 10 episodes on Gym-MuJoCo and Adroit tasks and for 100 for Maze2d and AntMaze tasks.

\subsection{Hyperparameters}
\label{appendix:hyperparameters_list}
\begin{table}[ht]
\centering
\caption{BC and BC-$N\%$ hyperparameters. $\dagger$ used for the best trajectories choice.}\label{table:bc_hyp}
\begin{tabular}{cll}
\toprule
& Hyperparameter & Value \\
\midrule
\multirow{3}{*}{BC hyperparameters} & Optimizer & Adam~\citep{adam}    \\
                    & Learning Rate & 3e-4       \\
                    & Mini-batch size      & 256        \\
\midrule
\multirow{3}{*}{Architecture}         & Policy hidden dim    & 256        \\
                                      & Policy hidden layers & 2          \\
                                      & Policy activation function & ReLU \\
\midrule
\multirow{3}{*}{BC-$N\%$ hyperparameters} & Ratio of best trajectories used & 0.1    \\
                         & Discount factor$^\dagger$      & 1.0       \\
                         & Max trajectory length$^\dagger$      & 1000       \\
\bottomrule
\end{tabular}
\vspace{6pt}
\end{table}

\begin{table}
\centering
\caption{TD3+BC hyperparameters.}
\label{table:td3_hyp}
\begin{tabular}{cll}
\toprule
& Hyperparameter & Value \\
\midrule
\multirow{9}{*}{TD3 hyperparameters} & Optimizer & Adam~\citep{adam} \\
                                      & Critic learning rate & 3e-4 \\
                                      & Actor learning rate  & 3e-4 \\
                                      & Mini-batch size      & 256 \\
                                      & Discount factor      & 0.99 \\
                                      & Target update rate   & 5e-3 \\
                                      & Policy noise         & 0.2 \\
                                      & Policy noise clipping & (-0.5, 0.5) \\
                                      & Policy update frequency & 2 \\
\midrule
\multirow{6}{*}{Architecture}         & Critic hidden dim    & 256        \\
                                      & Critic hidden layers & 2          \\
                                      & Critic activation function & ReLU \\
                                      & Actor hidden dim     & 256        \\
                                      & Actor hidden layers  & 2          \\
                                      & Actor activation function & ReLU \\
\midrule
\multirow{1}{*}{TD3+BC hyperparameters}  & $\alpha$             & 2.5 \\
\bottomrule
\end{tabular}
\vspace{6pt}
\end{table} %

\begin{table}
\centering
\caption{\centering CQL and Cal-QL hyperparameters. Note: used hyperparameters are suboptimal on Adroit for the implementation we provide.}\begin{tabular}{cll}
\toprule
& Hyperparameter & Value \\
\midrule
\multirow{7}{*}{SAC hyperparameters} & Optimizer & Adam~\citep{adam}    \\
                                      & Critic learning rate & 3e-4       \\
                                      & Actor learning rate  & 1e-4       \\
                                      & Mini-batch size      & 256        \\
                                      & Discount factor      & 0.99       \\
                                      & Target update rate   & 5e-3   \\
                                      & Target entropy       & -1 $\cdot$ Action Dim \\
                                      & Entropy in Q target  & False  \\ 
\midrule
\multirow{7}{*}{Architecture}         & Critic hidden dim    & 256        \\
                                      & Critic hidden layers & 5, AntMaze          \\
                                      & & 3, otherwise          \\
                                      & Critic activation function & ReLU \\
                                      & Actor hidden dim     & 256        \\
                                      & Actor hidden layers  & 3          \\
                                      & Actor activation function & ReLU \\
\midrule
\multirow{10}{*}{CQL hyperparameters}  & Lagrange             & True, Maze2d and AntMaze      \\
                                      &                      & False, otherwise     \\
                                      & Offline $\alpha$                & 1.0, Adroit         \\
                                      &                & 5.0, AntMaze         \\
                                      &                & 10.0, otherwise         \\
                                      & Lagrange gap                & 5, Maze2d         \\
                                      &                 & 0.8, AntMaze         \\
                                      & Pre-training steps   & 0       \\
                                      & Num sampled actions (during eval) & 10 \\
                                      & Num sampled actions (logsumexp) & 10 \\
\midrule
\multirow{3}{*}{Cal-QL hyperparameters}  & Mixing ratio & 0.5\\
 & Online $\alpha$ & 1.0, Adroit         \\
 & & 5.0, AntMaze         \\
\bottomrule
\end{tabular}
\vspace{6pt}
\label{table:cql_hyp}
\end{table}

\begin{table}
\centering
\caption{IQL hyperparameters.}\label{table:iql_hyp}
\begin{tabular}{cll}
\toprule
& Hyperparameter & Value \\
\midrule
\multirow{16}{*}{IQL hyperparameters}  & Optimizer & Adam~\citep{adam} \\
                                      & Critic learning rate & 3e-4 \\
                                      & Actor learning rate  & 3e-4 \\
                                      & Value learning rate  & 3e-4 \\
                                      & Mini-batch size      & 256 \\
                                      & Discount factor      & 0.99 \\
                                      & Target update rate   & 5e-3 \\
                                      & Learning rate decay & Cosine \\
                                      & Deterministic policy & True, Hopper Medium and Medium-replay\\
                                      && False, otherwise\\
                                      & $\beta$         & 6.0, Hopper Medium-expert\\
                                      && 10.0, AntMaze\\
                                      && 3.0, otherwise\\
                                      & $\tau$         & 0.9, AntMaze \\
                                      &          & 0.5, Hopper Medium-expert\\
                                      &          & 0.7, otherwise\\
                                      
\midrule
\multirow{9}{*}{Architecture}         & Critic hidden dim    & 256        \\
                                      & Critic hidden layers & 2          \\
                                      & Critic activation function & ReLU \\
                                      & Actor hidden dim     & 256        \\
                                      & Actor hidden layers  & 2          \\
                                      & Actor activation function & ReLU \\
                                      & Value hidden dim     & 256        \\
                                      & Value hidden layers  & 2          \\
                                      & Value activation function & ReLU \\
\bottomrule
\end{tabular}
\vspace{6pt}
\end{table}

\begin{table}
\centering
\caption{AWAC hyperparameters.}\label{table:awac_hyp}
\begin{tabular}{cll}
\toprule
& Hyperparameter & Value \\
\midrule
\multirow{8}{*}{AWAC hyperparameters} & Optimizer & Adam~\citep{adam} \\
                                      & Critic learning rate & 3e-4 \\
                                      & Actor learning rate  & 3e-4 \\
                                      & Mini-batch size      & 256 \\
                                      & Discount factor      & 0.99 \\
                                      & Target update rate   & 5e-3 \\
                                      & $\lambda$            & 0.1, Maze2d, AntMaze \\
                                      && 0.3333, otherwise\\
\midrule
\multirow{6}{*}{Architecture}         & Critic hidden dim    & 256        \\
                                      & Critic hidden layers & 2          \\
                                      & Critic activation function & ReLU \\
                                      & Actor hidden dim     & 256        \\
                                      & Actor hidden layers  & 2          \\
                                      & Actor activation function & ReLU \\
\bottomrule
\end{tabular}
\vspace{6pt}
\end{table} %

\begin{table}
\centering
\caption{SAC-$N$ and EDAC hyperparameters.}\label{table:sacn_hyp}
\begin{tabular}{cll}
\toprule
& Hyperparameter & Value \\
\midrule
\multirow{8}{*}{SAC hyperparameters} & Optimizer & Adam~\citep{adam}    \\
                                      & Critic learning rate & 3e-4       \\
                                      & Actor learning rate  & 3e-4       \\
                                      & $\alpha$ learning rate  & 3e-4     \\
                                      & Mini-batch size      & 256        \\
                                      & Discount factor      & 0.99       \\
                                      & Target update rate   & 5e-3   \\
                                      & Target entropy       & -1 $\cdot$ Action Dim \\
\midrule
\multirow{6}{*}{Architecture}         & Critic hidden dim    & 256        \\
                                      & Critic hidden layers & 3          \\
                                      & Critic activation function & ReLU \\
                                      & Actor hidden dim     & 256        \\
                                      & Actor hidden layers  & 3          \\
                                      & Actor activation function & ReLU \\
\midrule
\multirow{5}{*}{SAC-N hyperparameters}  & Number of critics            & 10, HalfCheetah\\
                                        &                & 20, Walker2d         \\
                                        &                & 25, AntMaze         \\
                                        &                & 200, Hopper Medium-expert, Medium-replay         \\
                                        &                & 500, Hopper Medium         \\
\midrule
\multirow{5}{*}{EDAC hyperparameters}  & Number of critics            & 10, HalfCheetah\\
                                        &                & 10, Walker2d, AntMaze         \\
                                        &                & 50, Hopper\\
                                        & $\mu$          & 5.0, HalfCheetah Medium-expert, Walker2d Medium-expert\\
                                        &                & 1.0, otherwise         \\
\bottomrule
\end{tabular}
\vspace{6pt}
\end{table}

\begin{table}[ht]
\centering
\caption{DT hyperparameters.}\label{table:dt_hyp}
\begin{tabular}{cll}
\toprule
& Hyperparameter & Value \\
\midrule
\multirow{20}{*}{DT hyperparameters}
& Optimizer & AdamW~\citep{adamw} \\
&Batch size   & $256$, AntMaze \\
&& $4096$, otherwise\\
&Return-to-go conditioning & (12000, 6000), HalfCheetah\\
&& (3600, 1800), Hopper \\
&& (5000, 2500), Walker2d \\
&& (160, 80), Maze2d umaze \\
&& (280, 140), Maze2d medium and large \\
&& (1, 0.5), AntMaze \\
&& (3100, 1550), Pen\\
&& (2900, 1450), Door\\
&& (12800, 6400), Hammer\\
&& (4300, 2150), Relocate\\
& Reward scale & 1.0, AntMaze \\
&& 0.001, otherwise \\
&Dropout & 0.1 \\
&Learning rate & 0.0008 \\
&Adam betas & (0.9, 0.999) \\
&Clip grad norm & 0.25 \\
&Weight decay & 0.0003 \\
& Total gradient steps & 100000 \\
&Linear warmup steps & 10000 \\
\midrule
\multirow{4}{*}{Architecture}         & Number of layers & 3  \\ 
&Number of attention heads    & 1  \\
&Embedding dimension    & 128  \\ 
&Activation function & GELU\\
%&Encoder channels & $32, 64, 64$ \\
% &Encoder filter sizes & $8 \times 8, 4 \times 4, 3 \times 3$ \\
% &Encoder strides & $4, 2, 1$ \\
\bottomrule
\end{tabular}
\vspace{6pt}
\end{table}

\begin{table}
\centering
\caption{SPOT hyperparameters.}
\label{table:spot_hyp}
\begin{tabular}{cll}
\toprule
& Hyperparameter & Value \\
\midrule
\multirow{5}{*}{VAE hyperparameters} & Optimizer & Adam~\citep{adam} \\
& Learning rate & 1e-3 \\
& Mini-batch size & 256 \\
& Number of iterations & $10^5$ \\
& KL term weight & 0.5 \\
\multirow{5}{*}{VAE architecture} & Encoder hidden dim & 750 \\
& Encoder layers & 3 \\
& Latent dim & 2 $\times$ action dim \\
& Decoder hidden dim & 750 \\
& Decoder layers & 3 \\
\multirow{10}{*}{TD3 hyperparameters} & Optimizer & Adam~\citep{adam} \\
                                      & Critic learning rate & 3e-4 \\
                                      & Actor learning rate  & 1e-4 \\
                                      & Mini-batch size      & 256 \\
                                      & Discount factor      & 0.99 \\
                                      & Target update rate   & 5e-3 \\
                                      & Policy noise         & 0.2 \\
                                      & Policy noise clipping & (-0.5, 0.5) \\
                                      & Policy update frequency & 2 \\
\midrule
\multirow{6}{*}{Architecture}         & Critic hidden dim    & 256        \\
                                      & Critic hidden layers & 2          \\
                                      & Critic activation function & ReLU \\
                                      & Actor hidden dim     & 256        \\
                                      & Actor hidden layers  & 2          \\
                                      & Actor activation function & ReLU \\
\midrule
\multirow{2}{*}{SPOT hyperparameters}  & $\lambda$             & {0.05, 0.1, 0.2, 0.5, 1.0, 2.0}, AntMaze \\
&& 1.0, Adroit\\
\bottomrule
\end{tabular}
\vspace{6pt}
\end{table} %

\end{document}